\documentclass[11pt]{article}
\usepackage[margin=25mm]{geometry}
\usepackage[utf8]{inputenc}
\usepackage{amsmath}
\usepackage{amsfonts}
\usepackage{amssymb}
\usepackage{graphicx}
\usepackage{verbatim}

\usepackage{microtype}
\usepackage{graphicx}
\usepackage{booktabs} 
\usepackage{amsmath}
\usepackage{amsthm}
\usepackage{amssymb}
\usepackage{amsfonts}
\usepackage{fixfoot}
\usepackage{algorithm}
\usepackage{algorithmic}
\usepackage{threeparttable}
\usepackage{graphicx}
\usepackage[dvipsnames]{xcolor}
\usepackage{parskip}
\usepackage{diagbox}
\usepackage{adjustbox}
\usepackage{multirow}
\usepackage[inline]{enumitem}
\usepackage{wrapfig}
\allowdisplaybreaks

\usepackage[caption=false]{subfig}

\usepackage{amsfonts}
\usepackage[toc,page,header]{appendix}

\allowdisplaybreaks

\usepackage{hyperref}
\usepackage{url}

\usepackage[english]{babel}
\usepackage[toc,page]{appendix}

\allowdisplaybreaks

\usepackage{amsmath,amsfonts,bm}
\usepackage{amsthm}
\usepackage{mathtools}

\newtheorem{lemma}{Lemma}[section]
\newtheorem{theorem}[lemma]{Theorem}
\newtheorem{assumption}[lemma]{Assumption}
\newtheorem{definition}[lemma]{Definition}
\newtheorem{corollary}[lemma]{Corollary}
\newtheorem{proposition}[lemma]{Proposition}

\newtheorem{remark}[lemma]{Remark}

        {\hspace*{\fill}$\Box$\par}

\newtheorem{innercustomgeneric}{\customgenericname}
\providecommand{\customgenericname}{}
\newcommand{\newcustomtheorem}[2]{%
  \newenvironment{#1}[1]
  {%
  \renewcommand\customgenericname{#2}%
  \renewcommand\theinnercustomgeneric{##1}%
  \innercustomgeneric
  }
  {\endinnercustomgeneric}
}

\newcustomtheorem{customthm}{Theorem}
\newcustomtheorem{customlemma}{Lemma}
\newcustomtheorem{customproblem}{Problem}

\DeclarePairedDelimiterX{\infdivx}[2]{(}{)}{%
  #1\;\delimsize\|\;#2%
}

\newcommand{\infdivalpha}{D_{\alpha}\infdivx}

\newcommand{\noisePrivPub}{\sigma_{\text{priv-pub}}^2}
\newcommand{\noisePubPriv}{\sigma_{\text{pub-priv}}^2}
\newcommand{\noiseInter}{\sigma_{\text{int}}^2}

\newcommand{\g}{\nabla}
\newcommand{\nn}{\nonumber}

\def\eqref#1{equation~\ref{#1}}

\def\1{\bm{1}}

\def\eps{{\epsilon}}

\newcommand{\Lsm}{L} 

\newcommand{\G}{G} 

\def\rva{{\mathbf{a}}}

\def\rvd{{\mathbf{d}}}

\def\rvg{{\mathbf{g}}}

\def\rvp{{\mathbf{p}}}
\def\rvq{{\mathbf{q}}}

\def\rvw{{\mathbf{w}}}
\def\rvx{{\mathbf{x}}}
\def\rvy{{\mathbf{y}}}
\def\rvz{{\mathbf{z}}}

\def\mA{{\bm{A}}}

\def\mR{{\bm{R}}}

\def\mX{{\bm{X}}}

\DeclareMathAlphabet{\mathsfit}{\encodingdefault}{\sfdefault}{m}{sl}
\SetMathAlphabet{\mathsfit}{bold}{\encodingdefault}{\sfdefault}{bx}{n}

\def\gB{{\mathsf{B}}}

\def\gD{{\mathsf{D}}}

\def\gI{{\mathcal{I}}}

\def\gM{{\mathcal{M}}}
\def\gN{{\mathcal{N}}}
\def\gO{{\mathcal{O}}}
\def\gP{{\mathsf{P}}}

\def\gR{{\mathcal{R}}}

\def\gW{{\mathcal{W}}}

\def\gZ{{\mathcal{Z}}}

\def\sI{{\mathbb{I}}}

\newcommand{\E}{\mathbb{E}}

\newcommand{\R}{\mathbb{R}}

\DeclareMathOperator*{\argmin}{arg\,min}

\newcommand{\SG}{\textit{ShuffleG}}
\newcommand{\DPSG}{\textit{DP-ShuffleG}}
\newcommand{\interleaved}{\textit{Interleaved-ShuffleG}}
\newcommand{\privpub}{\textit{Priv-Pub-ShuffleG}}
\newcommand{\pubpriv}{\textit{Pub-Priv-ShuffleG}}

\title{Improving the Convergence of Private Shuffled Gradient Methods with Public Data}
\author{
    Shuli Jiang$^{1}$\footnote{Email: \href{shulij@andrew.cmu.edu}{shulij@andrew.cmu.edu}}, Pranay Sharma$^{1}$, Zhiwei Steven Wu$^{1}$, Gauri Joshi$^{1}$\\
    \quad\\
    1 Carnegie Mellon University, Pittsburgh, PA
}
\date{February 2026} 

\begin{document}
\maketitle

\begin{abstract}
    We consider the problem of differentially private (DP) convex empirical risk minimization (ERM). While the standard DP-SGD algorithm is theoretically well-established, practical implementations often rely on shuffled gradient methods that traverse the training data sequentially rather than sampling with replacement in each iteration. Despite their widespread use, the theoretical privacy-accuracy trade-offs of private shuffled gradient methods (\textit{DP-ShuffleG}) remain poorly understood, leading to a gap between theory and practice. In this work, we leverage privacy amplification by iteration (PABI) and a novel application of Stein's lemma to provide the first empirical excess risk bound of \textit{DP-ShuffleG}. Our result shows that data shuffling results in worse empirical excess risk for \textit{DP-ShuffleG} compared to DP-SGD.
    To address this limitation, we propose \textit{Interleaved-ShuffleG}, a hybrid approach that integrates public data samples in private optimization. By alternating optimization steps that use private and public samples, \textit{Interleaved-ShuffleG} effectively reduces empirical excess risk. 
    Our analysis introduces a new optimization framework with surrogate objectives, varying levels of noise injection, and a dissimilarity metric, which can be of independent interest.
    Our experiments on diverse datasets and tasks demonstrate the superiority of \textit{Interleaved-ShuffleG} over several baselines.
\end{abstract}

\newpage
\tableofcontents
\newpage

\section{Introduction}
\label{sec:introduction}

Differential privacy (DP)~\cite{dwork2014algorithmic} has become a cornerstone of privacy-preserving machine learning, providing robust guarantees against the leakage of sensitive information in training datasets. 
In this work, we revisit the classical problem of $(\eps, \delta)$-differentially private convex empirical risk minimization (ERM)~\cite{bassily2014private_erm}, a framework that underpins many privacy-preserving machine learning tasks. Given the training dataset $\gD = \{\rvd_1, \dots, \rvd_n\}$, the private ERM problem can be formulated as:
\begin{align}
\label{eq:main_problem}
    &\min_{\rvx \in \R^d} \Big\{G(\rvx; \gD) = F(\rvx; \gD) + \psi(\rvx)\Big\},\\
    \nonumber
    &\text{ where } F(\rvx; \gD) = \frac{1}{n}\sum_{i=1}^{n} \Big\{ f_i(\rvx) := f(\rvx;\rvd_i) \Big\},
\end{align}
while ensuring 
$(\eps, \delta)$-differential privacy.
Here, $\rvx$ represents the model parameters, $\psi$ is a convex regularization and
$f_i$'s, for all $i$, are assumed to be convex, smooth, and Lipschitz-continuous\footnote{Convexity and smoothness are standard assumptions in the optimization literature. Lipschitzness is used only for privacy analysis \cite{Feldman2018privacy_amp_iter, ye2022singapore_paper}, and is not required for convergence analysis. Indeed, the Lipschitzness assumption can be removed by using gradient clipping ~\cite{Abadi2016dpsgd} in practice. For simplicity, we retain the Lipschitz assumption. }. 
For clarity of presentation, we consider twice differentiable $\psi$ (e.g., $\psi(\rvx) = \|\rvx\|^2$) in the main paper\footnote{
In our experiments, we consider $\psi$ as $\ell_1$ regularizer, $\ell_2$ regularizer and the projection operator. 
Detailed proofs for $\psi$ as the $\ell_1$ regularizer and as the projection operator onto a convex set are provided in Appendix~\ref{sec:appendix_other_reg}.
}.

A well-known approach to address the above problem is DP-SGD~\cite{bassily2014private_erm, Abadi2016dpsgd}, the private variant of stochastic gradient descent. In DP-SGD, at each iteration, a gradient is computed using a randomly picked sample from the training data,
followed by the addition of Gaussian noise to ensure differential privacy. 
However, the reliance on i.i.d. sampling introduces practical challenges. 
Consequently, DP-SGD in its original form is seldom implemented in practice. Instead, as noted in~\cite{Ponomareva2023dpfyml, chua2024how_private_dp_sgd, chua2024scalable_dp}, 
\textit
{shuffled gradient methods}, which traverse samples from the training dataset sequentially,
are often used in private optimization codebases and libraries, such as \textit{Tensorflow Privacy}~\cite{tensorflow_privacy} and \textit{PyTorch Opacus}~\cite{pytorch_opacus}. However, their privacy parameters are often incorrectly set based on the analysis of DP-SGD.

In the non-private setting, shuffled gradient methods (a class of methods, which we abbreviate with $\SG$) converge provably faster than SGD~\cite{liu2024last_iterate_shuffled_gradient}.
However, the convergence of \textit{private} shuffled gradient methods (which we denote by $\DPSG$), and how it compares to DP-SGD is poorly understood. 
This gap motivates our first key question: 

\fbox{%
\parbox{0.98\textwidth}{%
\textit{What is the privacy-convergence trade-off of private shuffled gradient methods?}
}%
}

To evaluate the privacy-convergence trade-off for a private optimization algorithm, we fix the privacy loss and measure the empirical excess risk, a standard metric in ERM. This metric captures the trade-off by accounting for both the error from noise injection for privacy preservation and the optimization error.

The privacy analysis of private $\SG$ presents unique challenges. 
For DP-SGD, the optimal privacy-convergence trade-off is achieved using a technique called privacy amplification by subsampling (PABS)~\cite{bassily2014private_erm}. However, PABS requires independent sampling of data points at every iteration, hence is not applicable to shuffled gradient methods.
Instead, privacy amplification by iteration (PABI), where privacy is amplified by hiding the intermediate parameters, emerges as a viable alternative in the convex setting. 
While prior work~\cite{Feldman2018privacy_amp_iter, altschuler2022apple_paper, ye2022singapore_paper} has studied the privacy guarantees of PABI and its use in analyzing the convergence of DP-SGD~\cite{feldman2020private_stochastic_convex_opt}, its application in the context of private $\SG$ remains unexplored.
Moreover, key challenge in analyzing private $\SG$ stems from the interaction between privacy noise and the bias in gradient estimates. Unlike DP-SGD, which uses unbiased gradients, $\SG$ accumulates biased gradients over each epoch. This bias, when combined with injected noise, creates nontrivial error terms that couple noise and model parameters. To handle this, we introduce a novel analysis using Stein’s lemma. Our approach addresses a challenge unique to private shuffled methods and goes beyond standard techniques.

\textbf{Excess Risk for Private Shuffled Gradient Methods.}
Addressing these challenges, we establish for the first time that the empirical excess risk of private shuffled gradient methods ($\DPSG$)
is $\widetilde{\gO} \Big( \frac{1}{n^{2/3}} \big( \frac{\sqrt{d}}{\eps} \big)^{4/3} \Big)$, given that the algorithm satisfies $(\eps, \delta)$-differential privacy.
This rate is worse than the empirical excess risk of DP-SGD, $\widetilde{\gO} \big( \frac{\sqrt{d}}{n \eps} \big)$, with matching lower bound~\cite{bassily2014private_erm}. The worse excess risk for $\DPSG$ matches similar empirical observations in \cite{chua2024scalable_dp}.
This disparity can perhaps be intuitively understood: shuffled gradient methods outperform SGD in non-private settings by trading in the bias of the gradient estimator for reduced variance. However, this also implies reduced inherent randomness, resulting in a worse privacy guarantee.

A promising direction to improve the empirical excess risk of private shuffled gradient methods is to leverage public data. The use of public samples, which can be accessed cheaply in many real-world scenarios, has been shown to improve utility in private learning problems, both theoretically~\cite{bassily20priv_query_release_pub_data, ullah2024public_data_priv_sco, block2024oracleefficient_dp_learning} and empirically~\cite{yu2022dp_fine_tune_lm, bu2023dp_fine_tune_fm}.
However, no prior work has explored the use of public samples in the context of private shuffled gradient methods. We consider the practical setting where the public and private datasets may come from different distributions.
While using public samples enhances the privacy guarantee, leading to less noise being added, it also risks greater divergence from the target objective. 
This trade-off motivates our second key question: 

\fbox{%
\parbox{0.98\textwidth}{%
\textit{
\small
Can public samples help improve the privacy-convergence trade-offs of private shuffled gradient methods?
}
}%
}

To answer this question, we propose the novel \textbf{generalized shuffled gradient framework} (see Algorithm~\ref{alg:generalized_shuffled_gradient_fm}), which introduces flexibility along several dimensions. First, instead of using a fixed objective function across all epochs, this framework allows optimizing a potentially different surrogate objective $G(\rvx; \gD^{(s)} \cup \gP^{(s)})$ in each epoch $s$,
where the dataset $\gD^{(s)} \cup \gP^{(s)}$ contains both private ($\gD^{(s)}$) and public ($\gP^{(s)}$) samples. 
Second, it supports varying levels of noise injection across epochs
to ensure the desired privacy guarantee.
For example, when optimizing with only public samples, no noise needs to be added, while noise is necessary when using private samples.
Further, to analyze this setup, we introduce a novel metric to measure the dissimilarity between the true and the surrogate objective
(see Assumption~\ref{ass:dissim_partial_lipschitzness}), specifically designed for shuffled gradient methods.
Unlike prior metrics, it explicitly accounts for the sample order used in gradient computations, enabling tighter convergence bounds.

Using the generalized shuffled gradient framework, we study three algorithms (see Section~\ref{subsec:pub_data_algos}): 1) $\privpub$, where the initial few epochs involve optimizing only using private samples, followed by some epochs on public samples; 2) in $\pubpriv$, the order of using public and private samples is reversed compared to $\privpub$; and 3) $\interleaved$, which involves using both private and public samples within each epoch. 
We show that $\interleaved$ achieves lower empirical excess risk than both $\privpub$ and $\pubpriv$ (\ref{tab:emp_risk_comp_main}), as well as $\DPSG$.

\textbf{Our Contributions:} 
\begin{enumerate}[itemsep=0mm, leftmargin=4mm, topsep=0mm]
    \item \textbf{Generalized Shuffled Gradient Framework.}
    In Section~\ref{sec:generalized_shufflg}, we present a generalized shuffled gradient framework (Algorithm~\ref{alg:generalized_shuffled_gradient_fm}) that allows surrogate objectives (based on public samples) and varying noise addition across epochs. We state the general convergence result, based on a novel dissimilarity metric, in Theorem~\ref{thm:convergence_generalized_shufflg_fm}.
    \item \textbf{Empirical Excess Risk of $\DPSG$.} In Section~\ref{sec:private_shuffled_g}, we show the empirical excess risk of $\DPSG$, a special case of Algorithm~\ref{alg:generalized_shuffled_gradient_fm}.
    \item \textbf{Effective Improvement of $\DPSG$ with Public Samples.}
    Based on the general framework, in Section~\ref{sec:public_data_dp_shuffleg}, we propose $\interleaved$ (see Algorithm~\ref{alg:interleaved}), an algorithm that uses both private and public samples within each epoch.
    $\interleaved$ achieves lower empirical excess risk compared to $\DPSG$ and other approaches that use public samples.  A summary of the empirical excess risk comparisons across algorithms is presented in Table~\ref{tab:emp_risk_comp_main}.
    \item \textbf{Experiments.}
    In Section~\ref{sec:exp}, we empirically demonstrate the superior performance of $\interleaved$ compared to the baselines
    in three tasks on diverse datasets.
\end{enumerate}

\subsection{Related Work}
\label{subsec:related_work}

\textbf{$\SG$.} Unlike SGD, theoretical convergence bounds for shuffled gradient methods in the non-private setting have only been established recently~\cite{mishchenko2021rr, mishchenko2021prox_fed_rr, liu2024last_iterate_shuffled_gradient}. Their performance compared to DP-SGD in the private setting remains unclear. 

\textbf{PABI.} 
The use of only the last-iterate model parameter at inference time has motivated privacy amplification by iteration (PABI)~\cite{Feldman2018privacy_amp_iter, altschuler2022apple_paper, ye2022singapore_paper}, which focuses on the privacy amplification by hiding intermediate model parameters. 
Most works on PABI focus exclusively on privacy analysis, with limited attention to its impact on convergence.
A few exceptions~\cite{Feldman2018privacy_amp_iter, feldman2020private_stochastic_convex_opt} study PABI in the context of DP-SGD for stochastic convex optimization, but the algorithms analyzed — such as \textit{Skip/Stop-PNSGD}~\cite{Feldman2018privacy_amp_iter}, or those requiring varying or extremely large batch sizes proportional to the dataset size~\cite{feldman2020private_stochastic_convex_opt} — are not practical for real-world applications (see Appendix~\ref{sec:appendix_related_work} for further discussion).
Moreover, PABI has not yet been explored in the context of private shuffled gradient methods. In this work, we address this gap by analyzing PABI within private shuffled gradient methods, with a focus on practical, widely used algorithms and their associated privacy–convergence trade-offs.

\textbf{Using Public Data or Surrogate Objectives.} 
While a substantial body of work has explored using public samples to improve model performance in private learning tasks, e.g.,~\cite{bassily20priv_query_release_pub_data, ullah2024public_data_priv_sco} —including the seminal work on PABI~\cite{feldman2020private_stochastic_convex_opt}, which demonstrated how PABI can improve privacy–convergence trade-offs in DP-SGD with access to public data 
— only a few~\cite{bie2022private_est_pub_data_shift, Bassily2023priv_adp_from_pub_source} consider distribution shifts between public and private datasets. 
No work, to our knowledge, addresses their usage in the context of shuffled gradient methods.
In addition, optimization of surrogate objectives has been studied in non-private SGD~\cite{Woodworth2023two_losses}, but remains unexplored in shuffled gradient methods.
For a more detailed discussion and a full survey, see Appendix~\ref{sec:appendix_related_work}.

\section{Problem Formulation}
\label{sec:prelim}

\textbf{Notation.} 
Given a positive integer $m$, $[m] \triangleq \{1,2,\dots, m\}$.
$\pi$ denotes a permutation, $\pi_j$ denotes the $j$-th element in permutation $\pi$, while $\Pi_n$ denotes the set of all permutations of $[n]$. $\|\cdot \|$ denotes the $\ell_2$ norm. 
$\sI_d$ denotes the identity matrix of dimension $d$.
$\rvx^* = \argmin_{\rvx \in \R^{d}} G(\rvx; \gD)$ denotes the minimizer of the true objective in \eqref{eq:main_problem}.

We solve the optimization problem given by (\ref{eq:main_problem}) under differential privacy, formally defined as follows:
\begin{definition}[Differential Privacy (DP)~\cite{dwork2014algorithmic}]
\label{def:DP}
    A randomized mechanism $\gM: \gW \rightarrow \gR$ 
    satisfies $(\eps, \delta)$-differential privacy, for $\eps \geq 0, \delta \in (0, 1) $, if for any two {\bf adjacent datasets} $\gD, \gD'$ 
    and for any subset of outputs $S \subseteq \gR$, it holds that $ \Pr[\gM(\gD) \in S] \leq e^{\eps} \Pr[\gM(\gD') \in S] + \delta$.
    $\eps$ and $\delta$ are called the privacy loss of the algorithm $\gM$. 
\end{definition}

In this work, we study private shuffled gradient methods (\DPSG), which optimize the objective in (\ref{eq:main_problem}) over $K$ epochs. During epoch $k \in [K]$, 
the $i$-th update is given by 
$\rvx_{i+1}^{(k)} = \rvx_{i}^{(k)} - \eta (\nabla f_{j}(\rvx_{i}^{(k)}) + \rho_i^{(k)} )$, $\forall i\in [n]$, where $\eta$ is the learning rate, $\rho_i^{(k)} \sim \gN(0, \sigma^2\sI_d)$ is the Gaussian noise vector, and $j\in [n]$ is the index of the sample selected for gradient computation. Each sample is used exactly once per epoch, with regularization $\psi$ being applied only at the end of every epoch, to ensure convergence~\cite{mishchenko2021prox_fed_rr}. 

Next, we discuss the three most commonly studied variants of shuffled gradient methods (see Algorithm~\ref{alg:vanilla_shuffled_g}), which differ in how samples are selected in each epoch. \textbf{Incremental Gradient (IG)} method processes samples in the same \textit{pre-determined} order across epochs. \textbf{Shuffle Once (SO)} also follows the same order across epochs, but the order is a random permutation $\pi$ of $[n]$. Finally, \textbf{Random Reshuffling (RR)} picks a new random permutation $\pi^{(k)}$ at the beginning of each epoch $s$, which determines the order for that epoch.

To understand the privacy-convergence trade-offs of \DPSG, we define the \textit{empirical excess risk}
\begin{align}
\label{eq:empirical_excess_risk}
    \E\left[G(\rvx; \gD) - G(\rvx^*; \gD)\right]
\end{align}
where  
$\gD = \{\rvd_1,\dots,\rvd_n\}$ is a 
private training dataset, and $G(\rvx;\gD)$ is the \textit{target objective}.
The empirical excess risk captures the trade-off between privacy and convergence, reflecting the optimization error incurred to ensure some fixed privacy guarantee.

Our second goal is to effectively use public samples to improve the empirical excess risk of $\DPSG$. 
Alongside $\gD$, we have access to some public dataset $\gP$, with a potentially different distribution.
To allow the flexibility of using varying proportions of samples from both datasets, we formulate the optimization objective as a sequence of surrogate objectives that capture the proportion of public and private data used in each epoch.
In each epoch $k$, we use $n_d^{(k)} (\leq n)$ private samples from $\gD$ and $n-n_d^{(k)}$ public samples from $\gP$.
The private dataset used in epoch $k$, denoted $\gD^{(k)}$
is formed by generating a random permutation $\pi^{(k)}$ of $[n]$ and selecting the first $n_d^{(k)}$ samples in $\pi^{(k)} (\gD)$, namely, $\gD^{(k)} := \{ \rvd_{\pi_{i}^{(k)}} \}_{i=1}^{n_d^{(k)}}$. 
The public data used in epoch $s$ is $\gP^{(k)} := \{ \rvp_j^{(k)} \}_{j=1}^{n-n_d^{(k)}} \subseteq \gP$ with $|\gP^{(k)}| = n-n_d^{(k)}$.
The \textit{surrogate objective} function used in epoch $k \in [K]$ is

{\small
\begin{align}
\label{eq:surrogate_objective}
    &G(\rvx; \gD^{(k)} \cup \gP^{(k)}) = F(\rvx; \gD^{(k)} \cup \gP^{(k)}) + \psi(\rvx),\\
    \nonumber
    &F(\rvx; \gD^{(k)} \cup \gP^{(k)}) 
    = \frac{1}{n} \Big(\sum_{\rvd \in \gD^{(k)}} f(\rvx; \rvd)
    + \sum_{\rvp \in \gP^{(k)}} f(\rvx; \rvp) \Big).
\end{align}
}

The above objective generalizes the target objective $ G(\rvx; \gD)$ in (\ref{eq:main_problem}) and recovers the objective of $\DPSG$,
when $n_d^{(k)} = n$ 
and $\gP^{(k)} = \emptyset, \forall k \in [K]$. 
It also allows flexible use of private and public samples, supporting schemes like public pre-training followed by private fine-tuning or mixed usage of private and public samples within an epoch. See Section~\ref{subsec:pub_data_algos} for the discussion on some such approaches.
We also define the objective difference between the target objective (\ref{eq:main_problem}) and the surrogate objective used in epoch $k$ (\ref{eq:surrogate_objective}),
\begin{align}
\label{eq:def_H}
    H^{(k)}(\rvx) 
    &= G(\rvx; \gD) - G(\rvx; \gD^{(k)} \cup \gP^{(k)}).
\end{align}

\section{Generalized Shuffled Gradient Framework}
\label{sec:generalized_shufflg}

\begin{algorithm}[h]
\caption{Generalized Shuffled Gradient Framework}
\label{alg:generalized_shuffled_gradient_fm} 
    \begin{algorithmic}[1]
    \STATE Input: 
    Initial point $\rvx_1^{(1)}$, learning rate $\eta$, number of epochs $K$.
    Private dataset $\gD$. Number of private samples to use $\{n_d^{(k)}\}_{k=1}^{K}$, $n_d^{(k)} \in \{0\} \bigcup [n]$. Public datasets $\{\gP^{(k)}\}_{k=1}^{K}$ with $|\gP^{(k)}| = n - n_d^{(k)}$.
    Noise standard deviation $\{\sigma^{(k)}\}_{k=1}^{K}$.
    \STATE \textit{IG}: Fix an order $\pi$, and set $\pi^{(k)} = \pi, \forall k \in [K]$
    \STATE \textit{SO}: Generate permutation $\pi$ of $[n]$, and set $\pi^{(k)} = \pi, \forall k\in [K]$
    \FOR{$k = 1,2,\dots,K$}
        \STATE \textit{RR}: Generate permutation $\pi^{(k)}$ of $[n]$
        \FOR{$i = 1,2,\dots, n_d^{(k)}$}
            \STATE $\rvx_{i+1}^{(k)} \leftarrow \rvx_i^{(k)} - \eta \Big(\nabla f(\rvx_i^{(k)};  \rvd_{\pi_i^{(k)}} ) + \rho_i^{(k)} \Big)$, where noise $\rho_i^{(k)} \sim \gN(0, (\sigma^{(k)})^2 \sI_d)$
        \ENDFOR
        \FOR{$i = n_d^{(k)}+1, n_d^{(k)} + 2, \dots, n$ 
        }
            \STATE $\rvx_{i+1}^{(k)} \leftarrow \rvx_i^{(k)} - \eta \Big( \nabla f(\rvx_i^{(k)}; \rvp_{i-n_d}^{(k)}) + \rho_i^{(k)}\Big)$, where noise $\rho_i^{(k)} \sim \gN(0, (\sigma^{(k)})^2\sI_d)$ 
        \ENDFOR
        \STATE $\rvx_1^{(k+1)} \leftarrow \argmin_{\rvx\in \R^{d}} n  \psi(\rvx) + \frac{\|\rvx - \rvx_{n+1}^{(k)}\|^2}{2\eta}$ 
    \ENDFOR
    \STATE \textbf{return} $\rvx_1^{(K+1)}$
    \end{algorithmic}
\end{algorithm}

In this section, we introduce the generalized shuffled gradient framework (Algorithm~\ref{alg:generalized_shuffled_gradient_fm}), which incorporates surrogate objectives, and noise injection for privacy preservation. This framework unifies 
the shuffled gradient methods ($\SG$) and their private variant $\DPSG$.
Specifically, $\DPSG$ corresponds to using only the true private dataset across all epochs ($n_d^{(k)}=n$ and $\gP^{(k)} = \emptyset$, $\forall k\in [K]$), while in the non-private version, no noise is added to the gradients ($\sigma^{(k)} = 0$). We provide the convergence analysis of the general framework here, with the convergence of $\DPSG$ as a corollary and its empirical excess risk derived in Section~\ref{sec:private_shuffled_g}.
We first introduce the assumptions and notation in Section~\ref{subsec:ass}.
To analyze the impact of surrogate objectives on convergence, we introduce a novel dissimilarity measure in Section~\ref{subsec:dissim_measure}
to measure the difference between the target objective, i.e., $G(\rvx;\gD)$, and surrogate objectives, i.e., $G(\rvx; \gD^{(k)} \cup \gP^{(k)})$.
Using this measure, we present the convergence result in Section~\ref{subsec:convergence}.

\subsection{Assumptions and Notation}
\label{subsec:ass}
We emphasize that only Assumptions~\ref{ass:convexity},~\ref{ass:smoothness}, and~\ref{ass:reg} are required for convergence analysis. Assumption~\ref{ass:lipschitzness} is standard for privacy analysis and can be removed by using gradient clipping in practice. 
Recall that $\gD$ is the training dataset in the target objective (\ref{eq:main_problem}), and $\gP$ is the public dataset. 

\begin{assumption}[Convexity]
\label{ass:convexity}
    $f(\rvx; \rvd)$ is convex, for all $\rvd \in \gD \cup \gP$.
\end{assumption}

\begin{assumption}[Smoothness]
\label{ass:smoothness}
    A function $f: \R^{d} \rightarrow \R$ is $L$-smooth if $\|\nabla f(\rvx) - \nabla f(\rvy)\| \leq \Lsm \|\rvx - \rvy\|$, for some $\Lsm \geq 0$, for all $\rvx, \rvy$. 
    $f(\rvx; \rvd)$ is $L$-smooth, $\forall \rvd \in \gD$;
    $f(\rvx; \rvp)$ is $\widetilde{L}$-smooth, $\forall \rvp \in \gP$.
\end{assumption}

\begin{assumption}[Lipschitz Continuity]
    \label{ass:lipschitzness}
    A convex function $f: \R^d \rightarrow \R$ is $\G$-Lipschitz if $\|\nabla f(\rvx)\| \leq \G$.
    $f(\rvx, \rvd)$ is $\G$-Lipschitz, $\forall \rvd\in\gD$;
    $f(\rvx; \rvp)$ is $\widetilde{G}$-Lipschitz, for all $\rvp \in \gP$.
\end{assumption}

\begin{assumption}
\label{ass:reg}
    The regularization function $\psi$ is twice differentiable and $\mu_{\psi} (\geq 0)$-strongly convex.
\end{assumption}
$L^* = \max\{L, \widetilde{L}\}$ and $G^{*}=\max\{G, \widetilde{G}\}$
denote the maximum smoothness and Lipschitz constants.

\subsection{Dissimilarity Measure}
\label{subsec:dissim_measure}

Next, we measure the dissimilarity between the target objective function $G(\rvx;\gD)$ (\ref{eq:main_problem}) and the surrogate objective function $G(\rvx; \gD^{(k)} \cup \gP^{(k)})$ in epoch $k \in [K]$ (\ref{eq:surrogate_objective}), based on the smoothness and the Lipschitzness of the objective difference $H^{(k)}(\rvx)$ defined in (\ref{eq:def_H}). 
It follows from Assumptions \ref{ass:smoothness} and \ref{ass:lipschitzness} that $H^{(k)}$ is $(L + L^*)$-smooth,
and $(G + G^*)$-Lipschitz continuous.
However, these constants can be too large, leading to loose convergence bounds. For example, when the dataset used in every epoch is exactly the same as the training dataset $\gD$, i.e., $n_d^{(k)} = n, \gP^{(k)} = \emptyset$, $\forall k\in [K]$,
then $H^{(k)} \equiv 0$. In this case, the smoothness and the Lipschitzness parameters of $H^{(k)}$ are both $0$. Therefore, for sharper analysis, we explicitly model the smoothness and Lipschitzness of $H^{(k)}$.

\begin{assumption}
\label{ass:H_smoothness}
    $H^{(k)}$ is $L_H^{(k)}$-smooth, for all $k \in[K]$.
\end{assumption}

\begin{assumption}
\label{ass:dissim_partial_lipschitzness}
    For epoch $k\in [K]$, there exists constants $\{C_i^{(k)}\}_{i=n_d^{(k)}+1}^{n}$ such that 
    \begin{align}
    \nonumber
        &\max_{\pi \in \Pi_n}\E\Big[ \Big\|
        \sum\textstyle_{j=n_d^{(k)} + 1}^{i} \Big( \nabla f(\rvx; \rvd_{\pi_j}) - \nabla f(\rvx; \rvp_{j-n_d^{(k)}}^{(k)}) \Big) \Big\|
        \Big]
         \leq C_i^{(k)}
    \end{align}
\end{assumption}

This measure of dissimilarity 
is inspired by prior work on distributed SGD-based optimization in non-private settings ~\cite{wang2021fednova},
and optimization with surrogate objectives \cite{Woodworth2023two_losses}.
These works define dissimilarity by directly comparing the gradients 
evaluated at individual samples. 
For example, in our case, this would be
$\|\nabla f(\rvx; \rvd) - \nabla f(\rvx; \rvp)\| \leq C$, for two private and public samples, $\rvd, \rvp$, respectively.
However, this crude notion of dissimilarity is less suitable for analyzing shuffled gradient methods and can lead to overly loose bounds.

To illustrate this, consider an extreme case where the public dataset $\gP$  is simply a permuted version of the private dataset $\gD$ under some fixed permutation $\widehat{\pi}\in \Pi_n$, i.e., $\rvp_i = \rvd_{\widehat{\pi}_i}, \forall i\in [n]$,
and $n_d^{(k)} = 0$, for all $k\in [K]$. 
In other words, the public dataset is identical to the private dataset.
The typical dissimilarity measure based on the gradients of individual samples would imply $\| \nabla H^{(k)} (\rvx)\| \leq C$, suggesting nonzero dissimilarity.
However, the two datasets are essentially identical.
Our proposed dissimilarity measure in Assumption~\ref{ass:dissim_partial_lipschitzness} using
$C_n^{(k)}=0$ correctly captures this, which implies $\| \nabla H^{(k)} (\rvx) \| \equiv 0, \forall k \in[K]$, more accurately reflecting the underlying intuition.

\subsection{Convergence}
\label{subsec:convergence}

Based on the above dissimilarity measure, we present the convergence results of \textit{generalized shuffled gradient framework} in Algorithm~\ref{alg:generalized_shuffled_gradient_fm}.
In the convergence bound, we use $\sigma_{any}^2 = \frac{1}{n} \sum_{i=1}^{n} \|\nabla f_i(\rvx^{*})\|^2$ to
measure the optimization uncertainty,
following~\cite{liu2024last_iterate_shuffled_gradient}.
See Appendix~\ref{sec:appendix_proof_main_thm} for the full proof.

\begin{theorem}[Convergence of Generalized Shuffled Gradient Framework]
\label{thm:convergence_generalized_shufflg_fm}
    Under Assumptions \ref{ass:convexity}, \ref{ass:smoothness}, \ref{ass:reg}, \ref{ass:H_smoothness}, \ref{ass:dissim_partial_lipschitzness}, for $\beta > 0$, 
    if $\mu_{\psi} \geq L_H^{(k)} + \beta$, $\forall k\in [K]$, and 
     $\eta \lesssim \frac{1}{n \Lsm^* \sqrt{1+\log K}}$,   
    Algorithm~\ref{alg:generalized_shuffled_gradient_fm} guarantees
    \begin{align}
    \label{eq:convergence_generalized_shufflg_fm}
        &\E\left[ G(\rvx_1^{(K+1)};\gD) \right] - G(\rvx^*;\gD)
        \lesssim \underbrace{\eta^2 n^2 \sigma_{any}^2 (1+\log K) \Lsm^*}_{\text{Optimization Uncertainty}} + \underbrace{ \frac{\|\rvx_1^{(1)} - \rvx^{*}\|^2}{\eta n K}}_{\text{Due to Initialization}} \\
        \nonumber
        &\quad \quad + \max_{s \in [K]} \Bigg(
        \underbrace{ \frac{1}{n^2 \beta}\sum_{k=1}^{s} \frac{ (C_n^{(k)})^2 }{s+1-k} }_{\text{Non-vanishing Dissimilarity}} 
        + \underbrace{ \eta^2 \Lsm^* \sum_{k=1}^{s} \frac{\frac{1}{n}\sum_{i=n_d^{(k)}+1}^{n-1}(C_i^{(k)})^2} {s+1-k} }_{\text{Vanishing Dissimilarity}}
        + \underbrace{ \eta^2 \Lsm^* nd \sum_{k=1}^{s} \frac{ (\sigma^{(k)})^2}{s+1-k} }_{\text{Due to Noise Injection}} \Bigg). \nn
    \end{align}
    The expectation is taken w.r.t. the noise $\{\rho_i^{(k)}\}$ 
    and the permutations $\pi^{(k)}$, $\forall i\in [n], k\in [K]$.
\end{theorem}

\textbf{Convergence Bound for Non-Private Shuffled Gradient Methods.} In the special case where we only use private data, i.e., $n_d^{(k)}= n$, and no noise is injected, $\sigma^{(k)}=0$, and the dissimilarity is $C_i^{(k)} = 0$, $\forall i\in \{n_d^{(k)}+1,\dots,n\}$, $\forall k\in [K]$.
Consequently, the bound in (\ref{eq:convergence_generalized_shufflg_fm}) recovers the convergence rate of non-private shuffled gradient methods in~\cite{liu2024last_iterate_shuffled_gradient}.
See Corollary~\ref{corollary:convergence_non_private_shuffled_g} in Appendix~\ref{subsec:appendix_non_private_shuffled_g} for details.

\textbf{Impact of Dissimilarity.}
In the convergence bound (\ref{eq:convergence_generalized_shufflg_fm}), we identify two dissimilarity terms: one \textit{vanishing} and the other \textit{non-vanishing}.
When the dataset used for optimization, $\gD^{(k)}\cup \gP^{(k)}$, differs from the original dataset $\gD$, Algorithm~\ref{alg:generalized_shuffled_gradient_fm} cannot be expected to converge exactly to the true solution $\rvx^{*}$.
This discrepancy is captured by the \textit{Non-vanishing Dissimilarity} term, which remains nonzero when $C_n^{(k)} \neq 0$ (Assumption~\ref{ass:dissim_partial_lipschitzness}).
Conversely, if the datasets $\gD^{(k)}\cup \gP^{(k)}$ used across all epochs are permutations of the original dataset $\gD$ (e.g., see the discussion following Assumption~\ref{ass:dissim_partial_lipschitzness} in Section~\ref{subsec:dissim_measure}), then $C_n^{(k)} = 0, \forall k$ and the \textit{Non-vanishing Dissimilarity} term disappears.
Nevertheless, even with identical data, different sample orderings can still lead to varying optimization trajectories in \SG. 
This effect is reflected through $\{C_i^{(k)} > 0\}$ in the \textit{Vanishing Dissimilarity} term. 
However, the $\eta^2$ scaling in this term ensures it does not dominate the overall convergence bound.

\section{Private Shuffled Gradient Methods}
\label{sec:private_shuffled_g}

In this section, we derive the convergence rate and the empirical excess risk of $\DPSG$. As discussed earlier, $\DPSG$ is a special case of the \textit{generalized shuffled gradient framework} (Algorithm~\ref{alg:generalized_shuffled_gradient_fm}) where the surrogate objectives
are identical to the target objective,
i.e., $n_d^{(k)} = n, \gP^{(k)} = \emptyset$, and $\sigma^{(k)} = \sigma$, $\forall k \in [K]$. We first present the convergence bound of $\DPSG$ as a corollary of Theorem~\ref{thm:convergence_generalized_shufflg_fm} in Corollary~\ref{corollary:conv_dpsg}. In Lemma~\ref{lemma:privacy_private_shuffled_g}, we state the differential privacy guarantee of $\DPSG$, in terms of the noise variance $\sigma$.
Finally, we discuss the choice of the learning rate $\eta$ and the number of epochs $K$ to 
achieve the minimal empirical excess risk while ensuring $(\eps, \delta)$-DP.

\begin{corollary}[Convergence of $\DPSG$\footnote{One can set $\beta = 0$ when $L_H^{(k)} = 0$, which is the case here since no surrogate datasets is used. This implies $\mu_{\psi} \geq 0$, as indicated in Assumption~\ref{ass:reg}, suffices to ensure convergence.}]
\label{corollary:conv_dpsg}
    Under the conditions in Theorem~\ref{thm:convergence_generalized_shufflg_fm}, with $n_d^{(k)} = n$, $\gP^{(k)} = \emptyset$, and 
    $\sigma^{(k)} = \sigma$ for all $k \in [K]$, Algorithm~\ref{alg:generalized_shuffled_gradient_fm} ($\DPSG$) guarantees
    \begin{align*}
        & \E [G(\rvx_1^{(K+1)};\gD)] - G(\rvx^*;\gD)
        \lesssim \eta^2 n^2(1+\log K) L^* + \frac{\|\rvx_1^{(1)} - \rvx^{*}\|^2}{\eta n K} + \eta^2 n d \sigma^2 L^{*} (1 + \log K)
    \end{align*}
\end{corollary}

\textbf{Privacy of $\DPSG$.}
We use privacy amplification by iteration (PABI \cite{Feldman2018privacy_amp_iter}) to bound the privacy loss within an epoch.
The privacy amplification arises because the intermediate iterates within an epoch $\{\rvx_i^{(k)} \}_{i=1}^n$ are hidden. 
However, PABI requires the update steps to be ``contractive" (Definition~\ref{def:contraction}). While each gradient step within an epoch (line 8 of Algorithm~\ref{alg:generalized_shuffled_gradient_fm}) satisfies this property, the regularization step at the end of each epoch (line 14 of Algorithm~\ref{alg:generalized_shuffled_gradient_fm}) need not be contractive. This prevents us from using PABI \textit{across} epochs.
Hence, we use composition (Proposition~\ref{prop:rdp_composition}) to bound the total privacy loss across the $K$ epochs.
See Appendix~\ref{subsec:appendix_private_shuffled_g_privacy} for the proof.

\begin{lemma}[Privacy of $\DPSG$]
\label{lemma:privacy_private_shuffled_g}
    Under Assumptions~\ref{ass:convexity},~\ref{ass:smoothness} and~\ref{ass:lipschitzness}, given the learning rate $\eta \leq 1/L$, $\DPSG$ is $(\frac{2\alpha G^2 K}{\sigma^2} + \frac{\log 1/\delta}{\alpha - 1}, \delta)$-DP, for $\alpha > 1, \delta \in (0, 1)$.
\end{lemma}

\textbf{Empirical Excess Risk.}  
To ensure $\DPSG$ satisfies $(\eps, \delta)$-DP, we set
$\sigma = \widetilde{\gO}(\frac{G\sqrt{K}}{\eps})$, $\eta = \widetilde{\gO}(\frac{1}{nL^{*} K^{1/3}})$ and $K = \gO(\frac{n\eps^2}{d})$ to minimize the bound in Corollary~\ref{corollary:conv_dpsg}.
These choices yield the empirical excess risk of $\DPSG$
in of $\widetilde{O} ( ( \frac{1}{\eps} \sqrt{\frac{d}{n}} )^{4/3} )$.
See Appendix~\ref{subsec:appendix_private_shuffle_g_empirical_excess_risk} for a full derivation.

\textbf{Comparison with DP-(S)GD.} The lower bound for empirical excess risk when minimizing convex, smooth objectives is $\Omega(\sqrt{d}/(n\eps))$~\cite{bassily2014private_erm}. DP-GD and DP-SGD
achieve matching upper bounds. However, the above bound suggests a worse empirical excess risk for $\DPSG$.
This aligns partially with the empirical findings of~\cite{chua2024scalable_dp}
that shuffled gradient methods (SO and RR) underperform DP-SGD in private binary classification tasks with the same privacy guarantees. Their setting, however, allows intermediate model parameter releases and does not require convex objectives.
The worse privacy guarantee of $\DPSG$ 
can be intuitively explained: 
the gradient estimators in shuffled gradient methods~\cite{liu2024last_iterate_shuffled_gradient} have a smaller variance compared to SGD. Consequently, to ensure the same privacy guarantee, these methods need a larger noise variance. 

\section{Leveraging Public Data}
\label{sec:public_data_dp_shuffleg}

Given the pessimistic empirical excess risk of $\DPSG$ discussed above, how can it be improved? In this section, we explore leveraging public data samples $\gP$ in the context of private shuffled gradient methods. We propose a novel approach that interleaves the usage of public and private samples during training, demonstrating its effectiveness in reducing empirical excess risk.

\subsection{Algorithms}
\label{subsec:pub_data_algos}

\begin{figure}[h]
    \centering
    \includegraphics[width=0.45\linewidth]{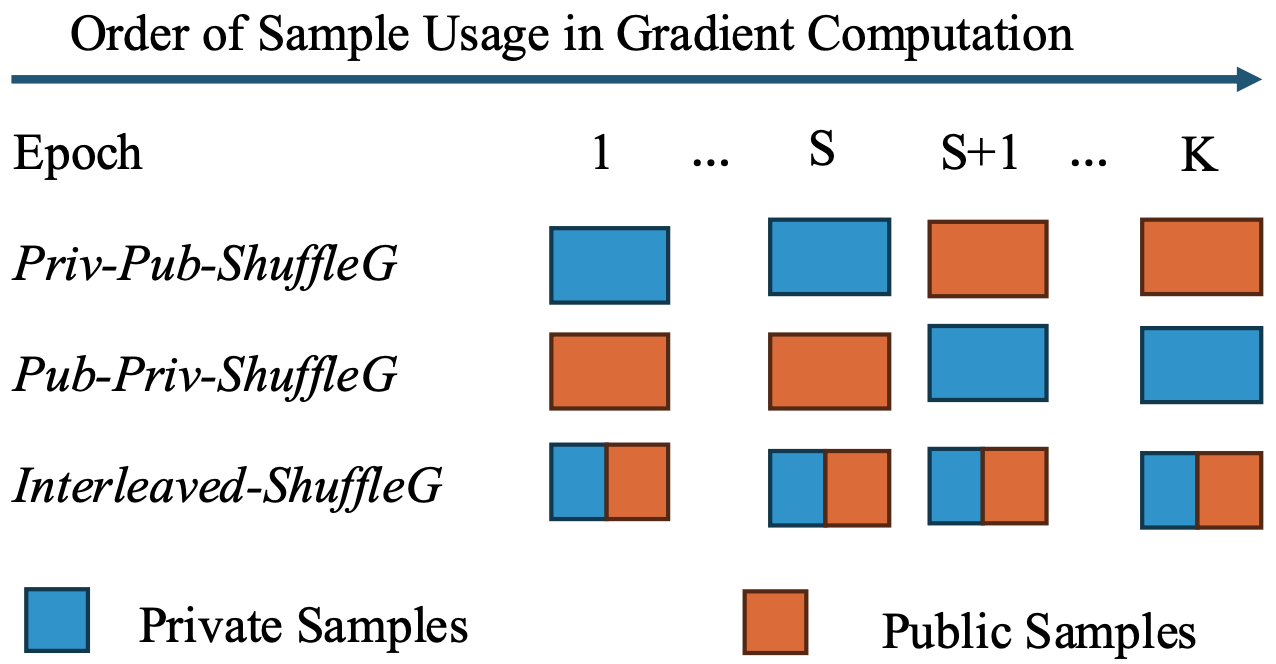}
    \caption{Illustration of algorithms. }
    \label{fig:alg_pub_data}
\end{figure}

The following algorithms, which leverage public samples, are specific instantiations of Algorithm~\ref{alg:generalized_shuffled_gradient_fm}. An illustration of these algorithms is provided in Figure~\ref{fig:alg_pub_data}.
We begin with two common baselines:

\textbf{1) $\privpub$} (Algorithm~\ref{alg:privpub}): Train only using the private dataset $\gD$ for the first $S$ epochs, where $S \in [K-1]$. For the remaining $K-S$ epochs, train only using the public dataset $\gP$. Specifically, in Algorithm~\ref{alg:generalized_shuffled_gradient_fm}:
\begin{enumerate*}
    \item[\textcircled{1}] For the first $S$ epochs ($k \leq S$): $n_d^{(k)} = n$ and $\gP^{(k)} = \emptyset$,
    \item[\textcircled{2}] For the remaining $K-S$ epochs ($k \geq S+1$): $n_d^{(k)} = 0$ and $\gP^{(k)} = \gP$.
\end{enumerate*}
Consequently, during the first $S$ epochs, there is no non-vanishing dissimilarity. 
In addition, to preserve privacy, noise with variance $\noisePrivPub$ is added during these epochs.  During the last $K-S$ epochs, exclusively using the public data $\gP$ 
results in the non-vanishing dissimilarity $C_n^{(k)} = C_n^{\text{full}}$, $\forall k \in \{S+1,\dots,K\}$.
No noise is needed during these epochs.

\textbf{2) $\pubpriv$} (Algorithm~\ref{alg:pubpriv}): Train only on the public dataset $\gP$ for the first $S$ epochs, then switch to the private dataset $\gD$ for the remaining $K - S$ epochs. In Algorithm~\ref{alg:generalized_shuffled_gradient_fm}:
\begin{enumerate*}
    \item[\textcircled{1}] For the first $S$ epochs ($k \leq S$), $n_d^{(k)} = 0$ and $\gP^{(k)} = \gP$,
    \item[\textcircled{2}] For the remaining $K-S$ epochs ($k \geq S+1$), $n_d^{(k)} = n$ and $\gP^{(k)} = \emptyset$.
\end{enumerate*}
Consequently, during the first $S$ epochs, the non-vanishing dissimilarity is $C_n^{(k)} = C_n^{\text{full}}$, $\forall k\in [S]$, and no additive noise is added. During the last $K-S$ epochs, there is no non-vanishing dissimilarity (i.e., $C_n^{(k)} = 0$, for $k  \geq S+1$), and noise with variance $\noisePubPriv$ is added.

In addition, we propose
\textbf{3) $\interleaved$}: In each epoch $k\in [K]$, we fix $n_d^{(k)} = n_d$, where the first $n_d \in [n]$ steps use samples from the private dataset $\gD$ for gradient computation, followed by $n - n_d$ steps using samples from the public dataset $\gP$. 
As a result, during every epoch, 
the first $n_d$ steps involve no non-vanishing dissimilarity, while the remaining $n-n_d$ steps introduce dissimilarity arising from the use of samples from the public dataset.
Specifically, we denote the non-vanishing dissimilarity as $C_n^{(k)} = C_n^{\text{part}}$, $\forall k\in [K]$.
Noise with variance $\noiseInter$ is added at \textit{every} gradient step, even when using public samples, to enable privacy amplification by a factor of $1/(n-n_d+1)$~\cite{Feldman2018privacy_amp_iter}. The idea is to use noise from public gradients to further obscure the influence of private data.

\textbf{Convergence and Privacy.} We summarize the key parameters and convergence bounds for each algorithm as corollaries of Theorem~\ref{thm:convergence_generalized_shufflg_fm} in Appendix~\ref{subsec:appendix_algo_pub_data_param_convergence}.
Similar to the privacy analysis of $\DPSG$, we present the privacy guarantees for these algorithms that use public samples in Appendix~\ref{subsec:appendix_algo_pub_data_privacy}.

\subsection{Empirical Excess Risk Comparison}
\label{subsec:comparison}

We fix the number of gradient steps in all three algorithms: $K$ epochs, each with $n$ gradient steps.
Let $p$ denote the fraction of gradient steps computed using private samples. 
Therefore, in $\privpub$, $S = pK$; in $\pubpriv$, $K-S = pK$; and in \interleaved, $n_d^{(k)} = p n, \forall k$.
For simplicity, we assume both $pK$ and $pn$ are integers, and restrict $p$ to $[\frac{1}{K}, 1]$.

We ensure all the algorithms satisfy the same $(\eps, \delta)$-DP guarantee, and compare their empirical excess risk bounds in Table~\ref{tab:emp_risk_comp_main}.
Detailed derivations along with the choice of the noise variance, learning rate $\eta$, and number of epochs $K$, are provided in Appendix~\ref{subsec:appendix_algo_pub_data_risk}. The bounds in Table~\ref{tab:emp_risk_comp_main} illustrate a trade-off when using public samples. The first term, which reflects the cost of privacy, is reduced compared to $\DPSG$ (since $p \leq 1$). However, due to the dissimilarity between the public and private datasets, we get an additional non-vanishing term.

\begin{table}[]
    \centering
\begin{adjustbox}{width=0.65\textwidth}
    \begin{tabular}{|c|c|}
    \hline
        Algorithm & Empirical Excess Risk \\
    \hline
        {\small \privpub} & $\widetilde{\gO}\left( 
         \left(\frac{p}{n}\right)^{2/3} \left(\frac{\sqrt{d}}{\eps} \right)^{4/3}
         + \frac{(C_n^{\text{full}})^2}{n^2 \beta}
         \right)$ \\
    \hline
        {\small \pubpriv} & 
        $\widetilde{\gO}\left( 
         \left( \frac{p}{n} \right)^{2/3} \left(\frac{\sqrt{d}}{\eps} \right)^{4/3}
         + \frac{(C_n^{\text{full}})^2}{n^2 \beta}
         \right)$ \\
    \hline
        {\small \interleaved} & $\widetilde{\gO}\left( 
        \left(\frac{1}{n[(1-p)n + 1]}\right)^{2/3}
        \left( \frac{\sqrt{d}}{\eps} \right)^{4/3}
        + \frac{(C_n^{\text{part}})^2}{n^2\beta}
        \right)$\\
    \hline
    \end{tabular}
\end{adjustbox}
    \caption{
    Empirical excess risk
    in terms of dataset size $n$, model dimension $d$, privacy parameter $\eps$, the fraction of gradient steps that use private samples $p\in [\frac{1}{K},1]$, and the 
    dissimilarity measures $C_n^{\text{full}}$ and $C_n^{\text{part}}$, defined in \ref{subsec:pub_data_algos}.
    The notation $\widetilde{\gO}$ suppresses logarithmic factors.
    }
    \label{tab:emp_risk_comp_main}
\end{table}

First, $\interleaved$ reduces the privacy-related (first) term more aggressively than the other two schemes when 
$(\frac{1}{ n[(1-p)n + 1]})^{2/3} \leq (\frac{p}{n})^{2/3}$,
which holds for $p \geq 1/n$.
This improvement, shown in Appendix~\ref{subsec:appendix_algo_pub_data_privacy}, results from the more effective use of PABI within each epoch, which causes a reduction in privacy loss by a factor of $\frac{1}{n+1-n_d}$. On the other hand, the privacy loss bounds for $\privpub$ and $\pubpriv$ remain independent of $n$.

To compare the second terms in \ref{tab:emp_risk_comp_main}, recall that $C_n^{\text{full}}$ and $C_n^{\text{part}}$ measure the dissimilarity when, respectively, all or a part of the $n$ gradient steps in an epoch are computed using samples from the public dataset $\gP$.
Clearly $C_n^{\text{part}} \leq C_n^{\text{full}}$, hence, the dissimilarity term for $\interleaved$ is lower compared to the other two schemes.

To summarize, $\interleaved$ achieves a lower empirical excess risk than the other two baselines, $\privpub$ and $\pubpriv$. 
It also reduces the privacy-related term compared to $\DPSG$, at the cost of an additional error term due to dissimilarity.

\section{Experiments}
\label{sec:exp}

\textbf{Tasks.}
We consider three tasks, each associated with a distinct objective function. For every task, we describe the component function $f(\rvx; \rvq)$ on a given sample $\rvq \in \gD \cup \gP$, and the regularization function $\psi(\rvx)$. The true and the surrogate objective functions are constructed based on $f$ and $\psi$ accordingly. 
\begin{enumerate}[itemsep=0mm, leftmargin=4mm, topsep=0mm]
    \item \textbf{Mean Estimation}: $f(\rvx; \rvq) = \frac{1}{2}\| \rvx - \rvq\|^2$. $\psi(\rvx) = \gI\{\rvx \in \gB_{C}\}$, where $\gB_{C}$ is a ball of radius $C$ at the origin.
    \item \textbf{Ridge Regression}: 
    Let $\rvq = (\rva, y)$, where $\rva$ and $y$ represent the feature vector and the response, respectively.
    $f(\rvx; \rvq) = (\langle \rvx, \rva\rangle - y)^2$, $\psi(\rvx) = \frac{\lambda_{r}}{2}\|\rvx\|^2$ for $\lambda_{r} > 0$.
    \item \textbf{Lasso Logistic Regression}: 
    Let $\rvq = (\rva, y)$, for $y \in \{\pm 1\}$, represent the feature vector and label, respectively.
    $f(\rvx; \rvq) = -y \log (h(\rvx; \rva)) - (1-y) \log (h(\rvx; \rva))$, where $h(\rvx; \rva) = \frac{1}{1+\exp(-\langle \rvx, \rva \rangle)}$. $\psi(\rvx) = \lambda_{l} \|\rvx\|_1$ for $\lambda_{l} > 0$.
\end{enumerate}

\textbf{Datasets.} 
We use \texttt{MNIST-69} for mean estimation, \texttt{CIFAR-10} and \texttt{Crime} for ridge regression, and \texttt{COMPAS} and \texttt{CreditCard} for logistic regression.
We construct a private and a public set of samples from each dataset. 
The construction simulates real-worlds scenarios such as data corruption and demographic biases.  
See Appendix~\ref{subsec:appendix_datasets} for details about each dataset and the public-private samples partition.

\begin{figure}[h]
    \includegraphics[width=0.24\linewidth]{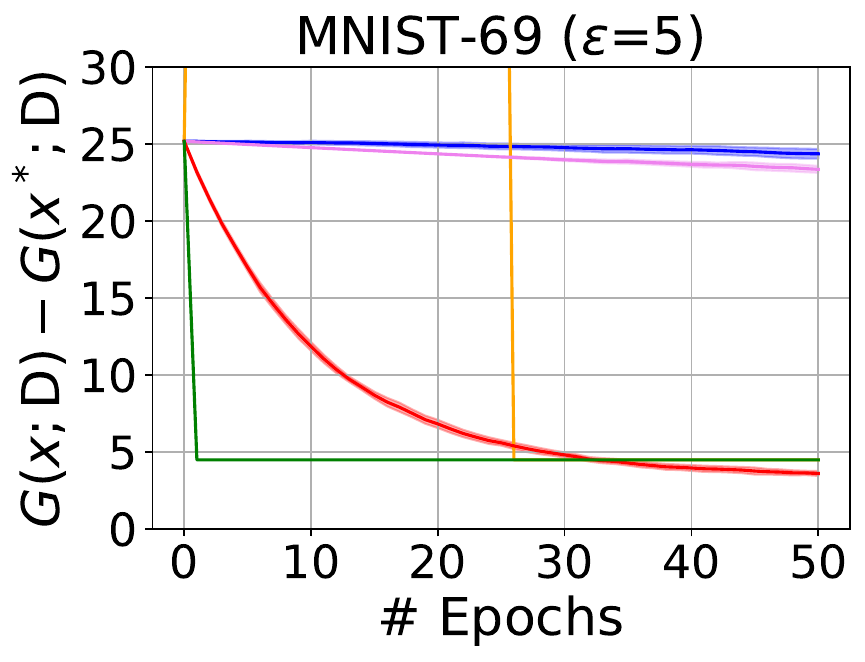}
    \includegraphics[width=0.24\linewidth]{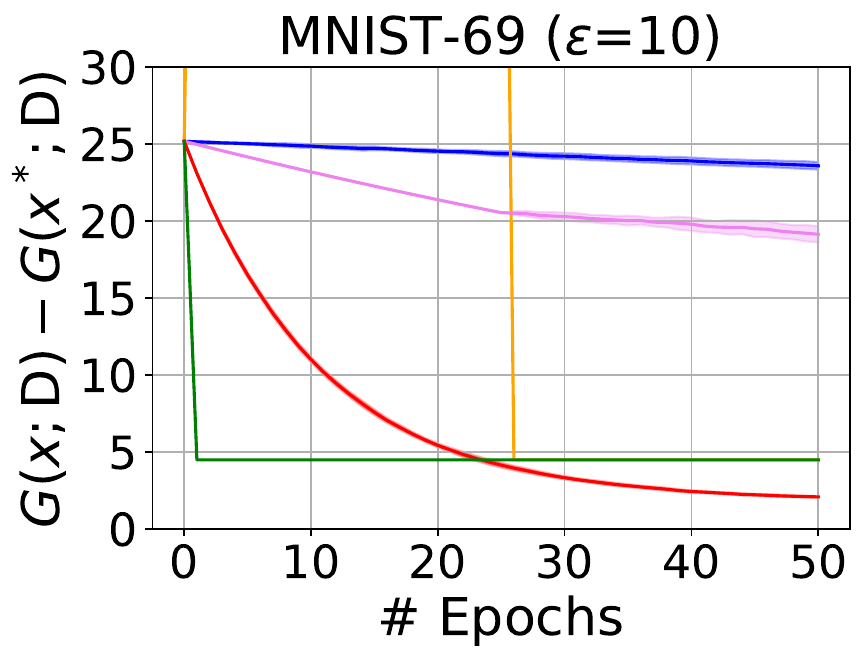}
    \includegraphics[width=0.24\linewidth]{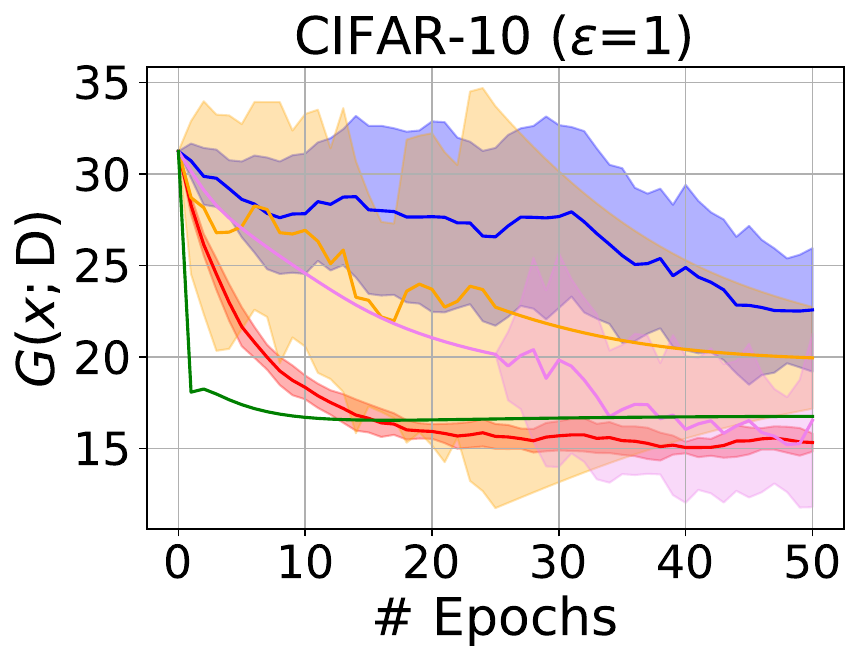}
    \includegraphics[width=0.24\linewidth]{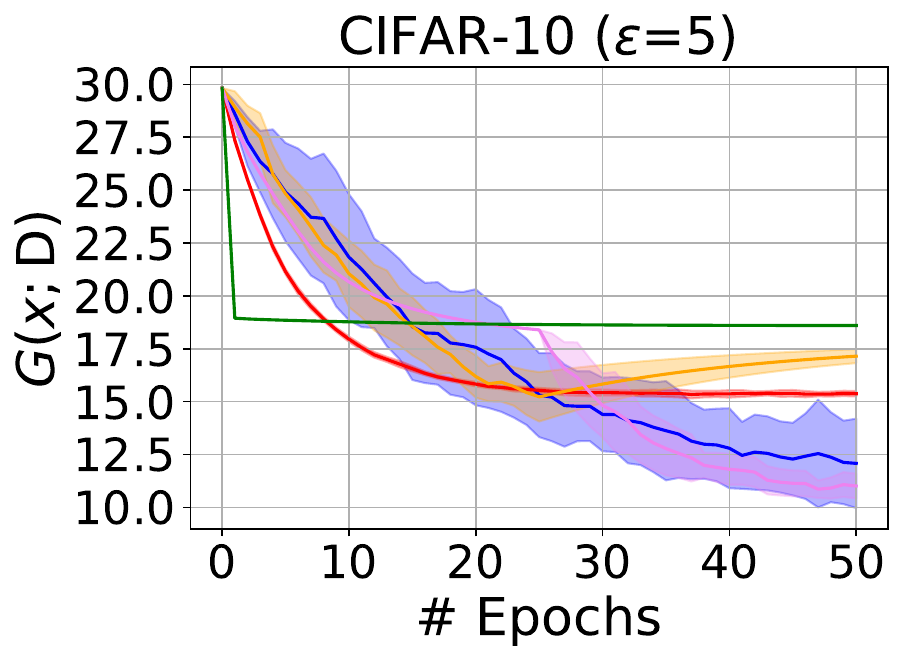}\\
    \includegraphics[width=0.24\linewidth]{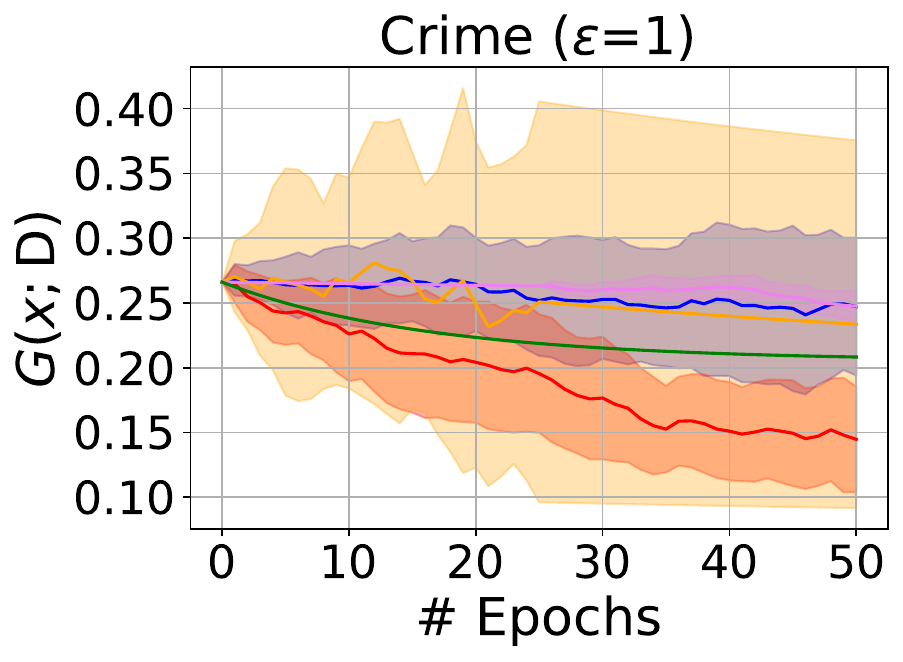}
    \includegraphics[width=0.24\linewidth]{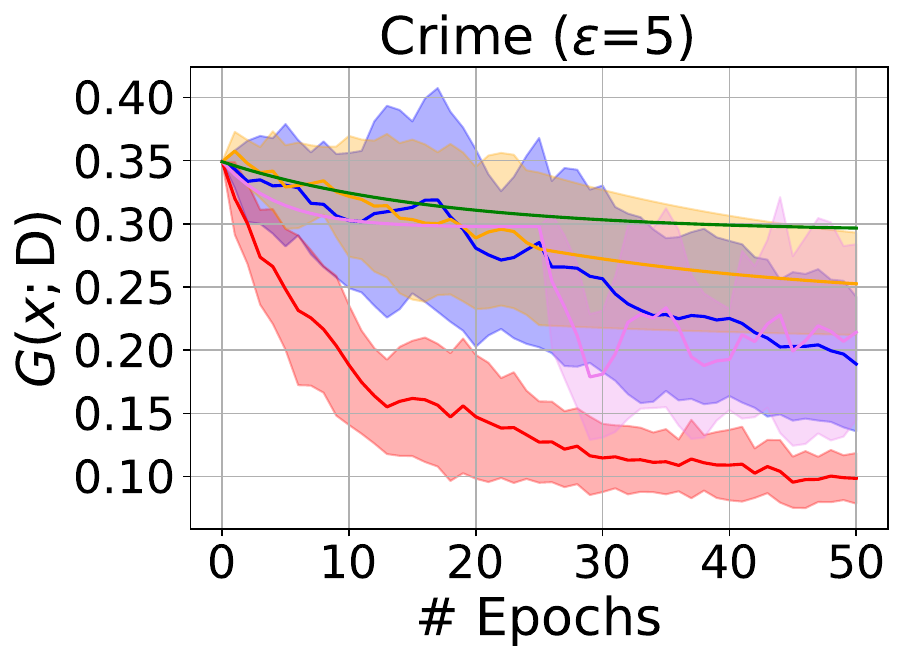}
    \includegraphics[width=0.24\linewidth]{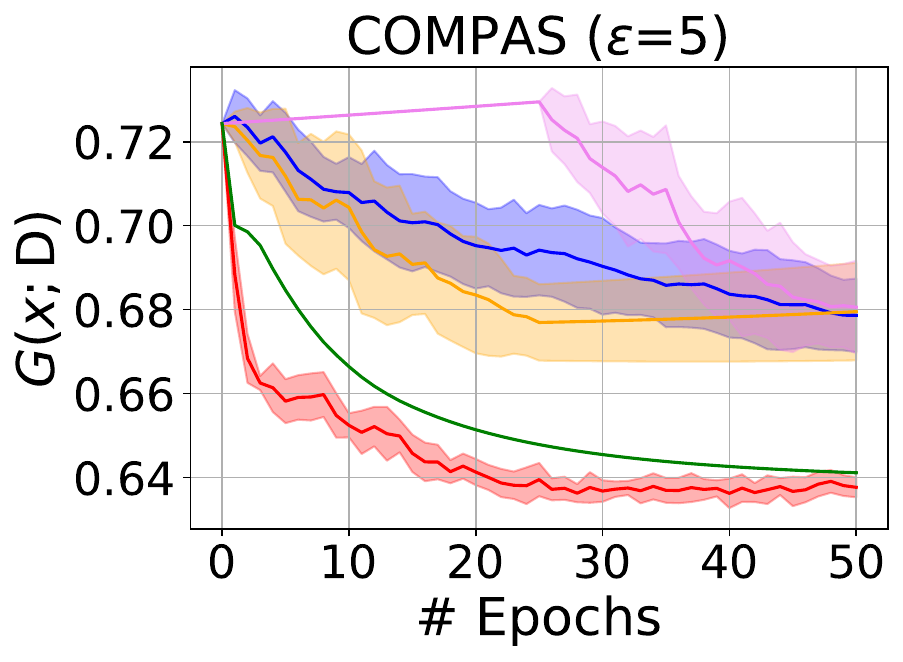}
    \includegraphics[width=0.24\linewidth]{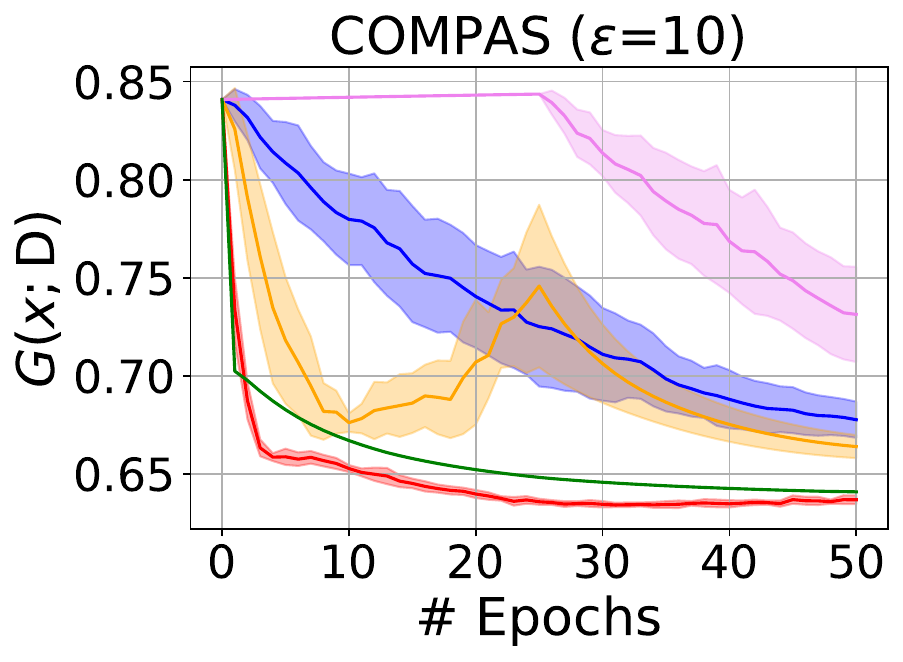}\\
    \centering
    \includegraphics[width=0.24\linewidth]{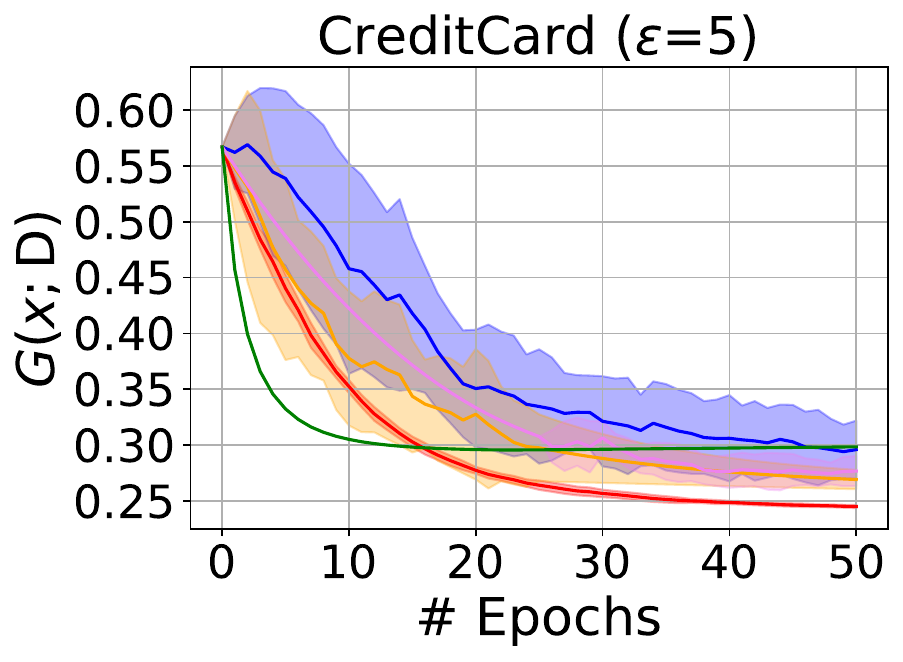}
    \includegraphics[width=0.24\linewidth]{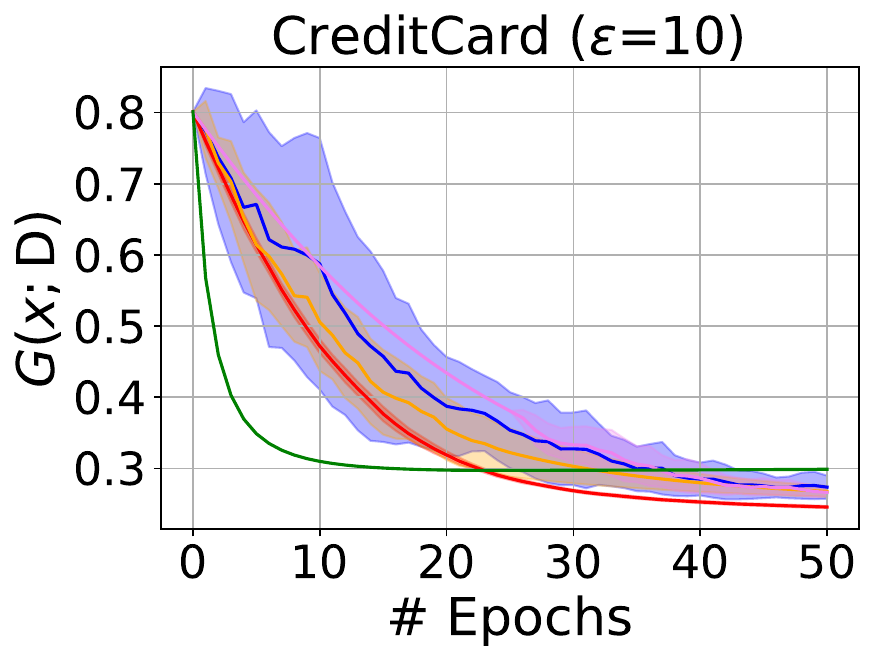}\\
    \centering
    \includegraphics[width=0.48\linewidth]{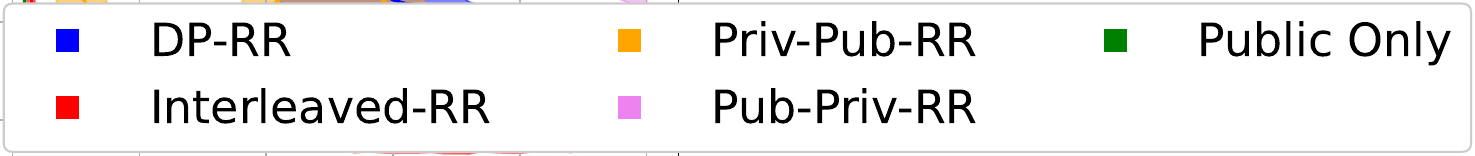}
    \caption{Results on each dataset across different tasks. Each algorithm runs for $K=50$ epochs, with privacy loss $\eps \in \{1, 5, 10\}$ and $\delta=10^{-6}$. 
   The solid lines represent the mean performance, while the shaded regions denote one std. across 10 random runs.
   }
    \label{fig:exp_res_main}
\end{figure}

\textbf{Baselines.} 
In our experiments, all optimization algorithms apply Random Reshuffling (RR) to private samples.
Thus, we replace ``$\SG$'' in their names with RR, resulting in \textit{Interleaved-RR}, \textit{Priv-Pub-RR}, \textit{Pub-Priv-RR} and \textit{DP-RR}.
And we include one additional baseline: \textit{Public Only}, which uses only the public dataset without noise injection.

\textbf{Hyperparameters.} 
In algorithms that use public samples, we set percentage of private sample usage as $p=0.5$. 
We set regularization parameters as $C=10, \lambda_{r}=0.1$, $\lambda_{l}=0.1$.
The number of epochs is $K = 50$.
To ensure the Lipschitz continuity of the objectives, we apply gradient clipping with a norm of 10. The privacy parameters are $\delta = 10^{-6}$, with $\eps \in \{5, 10\}$ in mean estimation and lasso logistic regression, and $\eps \in \{1, 5\}$ in ridge regression. 
We perform a grid search on the learning rate $\eta \in \{0.5, 0.1, 0.05, 0.01, \dots, 5\times 10^{-9}, 10^{-9}\}$.
Each experiment is repeated for 10 runs.

\textbf{Results.} 
All results are shown in Figure~\ref{fig:exp_res_main}, with each color corresponding to a specific optimization algorithm. Solid lines represent the mean performance over 10 runs, and shaded areas indicate one standard deviation. Results are reported for the best hyperparameter setting, chosen based on the lowest last-iterate loss.
Note that, due to this selection, results from the first half of epochs in $\pubpriv$ or $\privpub$ may not fully align with those of \textit{Public Only} or \DPSG.
Additional results of using other variants of $\SG$ to private samples and varying $p$ can be found in Appendix~\ref{subsec:appendix_more_results}.

\textbf{Discussion.}
Optimizing solely on the public dataset often leads to suboptimal solutions when the private and public datasets have slight distributional differences, as shown by the green curves. Conversely, relying only on the private dataset (i.e., $\DPSG$) is also suboptimal in high-privacy regimes, as shown by the blue curve, where excessive noise addition slows convergence. 
This is evident across all plots, except for \texttt{CIFAR-10} at $\epsilon = 5$, where the exception arises because larger $\eps$ values require less noise and hence, and the benefits of incorporating public data are reduced.
Moreover, in regimes with a smaller $\eps$, $\interleaved$
consistently outperforms the baselines, as shown by the red curves. 
This is consistent with our theoretical findings.

\section{Conclusion}
\label{sec:conclusion}

We study private convex ERM problems solved via shuffled gradient methods ($\DPSG$) and provide the first empirical excess risk bound, which is larger than the lower bound.
To reduce this risk, we incorporate public samples, and propose $\interleaved$, which interleaves the usage of private and public samples during training.
We demonstrate its superior performance compared to $\DPSG$ and other baselines, theoretically and empirically.
Impacts, limitations, and practical considerations are discussed in Appendix~\ref{sec:appendix_broader_impacts_limitation}.

\bibliography{mybib}
\bibliographystyle{unsrt}

\clearpage
\appendix
\section{Impacts, Limitations and Practicality}
\label{sec:appendix_broader_impacts_limitation}

\textbf{Impacts.}
Our work is among the first to analyze the privacy-utility trade-offs of private shuffled gradient methods (\DPSG), which are widely used in modern deep learning frameworks but differ subtly from the theoretically studied DP-SGD. By focusing on \DPSG—implemented in libraries like TensorFlow Privacy—we highlight a critical yet often overlooked gap between theory and practice. Our analysis shows that \DPSG\ can suffer from worse empirical excess risk, underscoring the importance of understanding how implementation choices, such as using shuffling instead of i.i.d. sampling for gradient computation, impact privacy and utility. This has practical implications for developers and researchers who may unknowingly rely on mismatched assumptions when deploying private optimization methods.
Moreover, we propose a novel training strategy that interleaves private and public samples to reduce empirical excess risk and improve the privacy-utility trade-off in shuffled gradient methods.

\textbf{Limitations.}
Our analysis provides only an upper bound on the empirical excess risk of $\DPSG$, without a matching lower bound. While we have made a concerted effort to tighten this upper bound using all techniques available to us, to the best of our knowledge, the gap leaves open the possibility that the observed worse performance of $\DPSG$ is an artifact of the analysis rather than an inherent limitation. Deriving a matching lower bound is challenging, as it requires fundamentally different tools and is non-standard—unlike typical lower bounds for problems, we seek one tailored to a specific algorithm. This makes the task highly non-trivial, and we leave it as an open problem.

Nevertheless, we conjecture that $\DPSG$ inherently incurs higher empirical excess risk than DP-SGD. 
As mentioned in the paragraph on \textbf{Comparison with DP-(S)GD} in Section~\ref{sec:private_shuffled_g}, our result partially aligns with prior empirical findings~\cite{chua2024scalable_dp}, which show that even with minimal noise, $\DPSG$ does not outperform DP-SGD on private binary classification tasks.
See this section also for the intuition on why \DPSG\ achieves a higher empirical excess risk compared to DP-SGD.

\textbf{A Note on Tensorflow's\footnote{https://www.tensorflow.org/} Shuffle Buffer in Optimization.} 
In TensorFlow, when the dataset is small, the implementation coincides with Random Reshuffling (RR), as discussed in this work. However, for large datasets, fully shuffling at every epoch is computationally expensive, and TensorFlow instead adopts the shuffle buffer mechanism.

The shuffle buffer introduces partial randomness controlled by the buffer size: when the buffer is as large as the dataset, it reduces to RR; when the buffer size is one, it effectively matches Incremental Gradient (IG), also discussed in this wrok, where the dataset is not permuted at all. Thus, the shuffle buffer can be viewed as an intermediate method lying between IG and RR.

We note that our work offers additional insight in the shuffled buffer mechanism:
\begin{enumerate}
    \item \textit{Shuffle buffer in the non-private setting.} 
    Our generalized shuffled gradient framework subsumes all variants of shuffled gradient methods, including IG and RR. In the non-private setting with constrained objectives, IG and RR achieve the same order-wise convergence rate. The analysis of IG ensures convergence with the worst-possible dataset ordering, hence also providing a bound on the convergence with the shuffle buffer.

    \item \textit{Shuffle buffer under privacy requirement.} 
    From a privacy standpoint, as discussed in Appendix E.2, we build upon the Privacy Amplification by Iteration (PABI) framework and leverage the joint convexity of scaled exponentiation of Rényi divergence (Lemma E.3) to analyze the privacy benefits of shuffling. Similar techniques were used in the seminal PABI paper~\cite{Feldman2018privacy_amp_iter} and its follow-up~\cite{ye2022singapore_paper}. Our results show that for moderate dataset sizes, there is no order-wise difference in the privacy loss between RR and IG; therefore, the shuffled buffer would also achieve a similar privacy loss. 
    We acknowledge that there might be room to further improve our results with RR in certain $\epsilon$ regimes. 
\end{enumerate}

To summarize, because TensorFlow’s shuffle buffer interpolates between IG and RR, our framework also provides theoretical insights into this widely used mechanism. In practice, TensorFlow has limited support for Poisson sampling, which underpins the original privacy analysis of DP-SGD~\cite{Abadi2016dpsgd}. As a result, practitioners often apply the privacy bounds of DP-SGD incorrectly when using RR or the shuffle buffer in place of Poisson sampling.
We believe our work takes a significant step towards understanding how shuffling affects the privacy and optimization trade-offs in noisy gradient-based algorithms and lays the groundwork for future improvements in this area.

\section{More about Related Work}
\label{sec:appendix_related_work}

\textbf{Private Optimization. }
The privacy loss and convergence of DP-SGD is well understood~\cite{Abadi2016dpsgd}, including tight upper and lower bounds for solving private empirical risk minimization problems in convex settings \cite{bassily2014private_erm}.
However, recent work has observed the gap between theory and practice: shuffled gradient methods are widely implemented in codebases, while the amount of noise applied to the gradients to ensure privacy guarantees is computed based on the analysis of DP-SGD \cite{chua2024how_private_dp_sgd, chua2024scalable_dp}.
This line of work, however, focuses on 
comparisons of the privacy loss between \DPSG\ and DP-SGD.
There is no unified analyses that consider both optimization and privacy.

\textbf{Shuffled Gradient Methods in the Non-Private Setting.}
While the convergence rate of SGD in non-private settings is well understood~\cite{Shamir2013sgd}, understanding the convergence of shuffled gradient methods, particularly Random Reshuffling (RR), has been a more recent development. Significant advances include characterizing the convergence rate of RR~\cite{mishchenko2021rr, mishchenko2021prox_fed_rr} and establishing last-iterate convergence results for shuffled gradient methods in general~\cite{liu2024last_iterate_shuffled_gradient}.
It is known that the best convergence rate by SGD in the non-private setting is $\gO\left(\frac{1}{\sqrt{T}}\right)$ for $T$ gradient steps, while~\cite{liu2024last_iterate_shuffled_gradient} shows that the convergence of shuffled gradient methods is $\gO\left(\frac{1}{K^{2/3}}\right)$, where $K = T / n$ is the number of epochs, each consisting of $n$ gradient steps based on $n$ samples. The results suggest that shuffled gradient methods converge faster than SGD in the non-private setting. However, it is unclear how their performances compare in the private setting, and we address this gap in this work.

\textbf{Privacy Amplification by Iteration (PABI). }
In many applications, only the last iterate model parameter is used during inference, while intermediate model parameters generated during training are discarded. However, the common privacy analyses based on composition of privacy loss per gradient step implicitly assumes that all intermediate model parameters are released. This discrepancy has motivated a line of research investigating the privacy loss of releasing only the last iterate model parameter~\cite{Feldman2018privacy_amp_iter, altschuler2022apple_paper, ye2022singapore_paper} and the privacy amplification that arises by hiding intermediate model parameters is referred to as privacy amplification by iteration (PABI). 

Most prior work on PABI focuses on privacy guarantees, with limited attention to its implication on the convergence of private optimizers in solving stochastic convex optimization (SCO)~\cite{Feldman2018privacy_amp_iter, feldman2020private_stochastic_convex_opt}, where the algorithms studied are often impractical. 
\cite{Feldman2018privacy_amp_iter} relies on convergence bounds for average iterates, contradicting the PABI setting where only the last iterate is released. To align with PABI, they analyze unrealistic algorithm variants like Skip-PNSGD and Stop-PNSGD, which randomly skip or stop gradient steps. 
While public data is discussed as a means for further privacy amplification, they do not address distributional shifts between public and private datasets.
In a related work, \cite{feldman2020private_stochastic_convex_opt} propose a DP-SGD based algorithm that utilizes PABI and achieves a tight upper bound on both the excess risk and the number of gradient computation. Its implementation, however, will be non-standard due to the use of exponentially decaying learning rates and varying batch sizes. In practice, fixed batch sizes are preferred for their simplicity, and efficiency in hardware acceleration.
Moreover,
the first few batches in the algorithm require sizes of $\frac{n}{2}, \frac{n}{4}, \dots$, where $n$ is the number of samples, which may be too large to fit into memory in practice.
\textit{The goal of our work is to analyze and to understand a practical variant of private optimization algorithm, i.e., \DPSG.}

\cite{altschuler2022apple_paper} shows that DP-SGD applied to convex, smooth, and Lipschitz objectives with a bounded domain $\gW$ incurs a finite privacy loss, rather an infinite privacy loss as privacy composition indicates. However, their analysis critically depends on privacy amplification by subsampling, specific to DP-SGD, and and the assumption that all model parameters remain within $\gW$ at every gradient step. 
Specifically, the update sequence analyzed is $\rvx_{t+1} = \text{Proj}_{\gW}\left( \rvx_{t} - \eta (\nabla f_i(\rvx_t) + \rho) \right)$, $\forall t\in [T]$, where $\rvx_t$ is the model parameter at $t$-th gradient step, $i \in [n]$ is the index of the sample used for gradient computation, $\rho$ is the Gaussian noise vector and $\text{Proj}$ is the projection operator, or a special case of regularization. 
This update ensures $\rvx_t \in \gW$, $\forall t\in [T]$, a key condition for applying their privacy bound.
However, in shuffled gradient methods, 
if regularization (e.g., projection) is applied after each gradient step, 
one would no longer approximate the full gradient after an epoch to ensure convergence to the target objective and hence, it is crucial to apply the regularization only at the end of every epoch~\cite{mishchenko2021prox_fed_rr}.
This implies that in shuffled gradient methods, we cannot guarantee $\rvx_t\in \gW$ for every gradient step.
These differences make the results of~\cite{altschuler2022apple_paper} inapplicable to private shuffled gradient methods.
Another work~\cite{ye2022singapore_paper} shows that in a more restricted setting where the objective function is strongly convex, even without a bounded domain, hiding intermediate model parameters leads to a finite privacy loss.

\textbf{Public Data Assisted Private Learning. }
There is a long line of work on using public samples to improve statistical learning tasks, e.g.,~\cite{bassily20priv_query_release_pub_data, block2024oracleefficient_dp_learning, ullah2024public_data_priv_sco}.
In machine learning, public data is commonly used to improve model performance by either identifying gradient subspaces~\cite{zhou2021bypassing, kairouz2021priv_erm_pub_subspace} or through public pre-training~\cite{yu2022dp_fine_tune_lm, bu2023dp_fine_tune_fm}.
Empirical studies have also explored the use of public samples in DP-SGD to solve ERM problems~\cite{Wang2020dp_small_pub_data}.
Limited attention has been given to addressing distribution shifts between private and public datasets in statistical learning tasks~\cite{bie2022private_est_pub_data_shift, Bassily2023priv_adp_from_pub_source}.
None of these works investigate the use of public samples in the context of private shuffled gradient methods for solving ERM or tackle distributional differences between public and private datasets in this specific setting.

\textbf{Optimization on a Surrogate Objective.}
The use of surrogate objectives in optimization is studied in the non-private setting in \cite{Woodworth2023two_losses},
which analyzes SGD using average-iterate methods.
However, no prior work has investigated the use of surrogate objectives in the context of shuffled gradient methods.

\section{Preliminaries}
\label{sec:appendix_prelim}

\subsection{Differential Privacy}
\label{subsec:appendix_dp_related}

We begin by defining standard $(\eps,\delta)$-differential privacy (DP) and Rényi Differential Privacy (RDP), the conversion between these two definitions and the composition theorem. 

\begin{definition}[Differential Privacy (DP)~\cite{dwork2014algorithmic}]
\label{def:appendix_DP}
    A randomized mechanism $\gM: \gW \rightarrow \gR$ with a domain $\gW$ and range $\gR$ satisfies $(\eps, \delta)$-differential privacy for $\eps \geq 0, \delta \in (0, 1) $, if for any two {\bf adjacent datasets} $\gD, \gD'$ 
    and for any subset of outputs $S \subseteq \gR$ it holds that 
    \begin{align*}
        \Pr[\gM(\gD) \in S] \leq e^{\eps} \Pr[\gM(\gD') \in S] + \delta
    \end{align*}
\end{definition}
Here, $\eps$ and $\delta$ are often referred to as the privacy loss of the algorithm $\gM$.

\begin{definition}[Renyi Divergence]
\label{def:renyi_divergence}
    For two probability distributions $P$ and $Q$ defined over $\gR$, the Renyi divergence of order $\alpha > 1$ is
    $\infdivalpha{P}{Q} := \frac{1}{\alpha-1}\log \E_{x \sim Q} \Big(\frac{P(x)}{Q(x)}\Big)^{\alpha}$.
\end{definition}

\begin{definition}[$(\alpha, \eps)$-Renyi Differential Privacy (RDP)~\cite{Mironov2017rdp}]
\label{def:RDP}
    A randomized mechanism $f: \gD\rightarrow \gR$ is said to have $\eps$-Renyi differential privacy of order $\alpha$, or $(\alpha, \eps)$-RDP for short, if for any adjacent $D, D' \in \gD$, it holds that $\infdivalpha{f(D)}{f(D')} \leq \eps$.
\end{definition}

\begin{proposition}[From RDP to DP (Proposition 3 of~\cite{Mironov2017rdp})]
\label{prop:rdp_to_dp}
    If $f$ is an $(\alpha, \eps)$-RDP mechanism, it also satisfies $(\eps + \frac{\log 1/\delta}{\alpha - 1}, \delta)$-DP for any $0 < \delta < 1$. 
\end{proposition}

\begin{proposition}[RDP Composition (Proposition 1 of~\cite{Mironov2017rdp})]
\label{prop:rdp_composition}
    Let $f: \gD \rightarrow \gR_1$ be $(\alpha, \eps_1)$-RDP and $g: \gR_1 \times \gD \rightarrow \gR_2$ be $(\alpha, \eps_2)$-RDP, then the mechanism defined as $(X, Y)$, where $X \sim f(D)$ and $Y \sim g(X, D)$, satisfies $(\alpha, \eps_1 + \eps_2)$-RDP. 
\end{proposition}

\subsection{Privacy Amplification By Iteration (PABI)}
\label{subsec:appendix_pabi_related}

We use PABI in the privacy analysis for improved privacy-convergence trade-offs.
At a high level, the privacy amplification arises due to hiding intermediate parameters and only release the last-iterate parameter in an optimization procedure.
Our analysis builds on the results of PABI in~\cite{Feldman2018privacy_amp_iter}. We begin by introducing the concept of contractive noisy iterations, the key setting where PABI applies, and how the optimization steps in private shuffled gradient methods fall under this setting.

\begin{definition}[Contraction (Definition 16 of~\cite{Feldman2018privacy_amp_iter})]
\label{def:contraction}
    For a Banach space $(\gZ, \|\cdot \|)$ A function $g: \gZ \rightarrow \gZ$ is said to be contractive if it is 1-Lipschitz, i.e., $\forall \rvx, \rvy \in \gZ$, $\|g(\rvx) - g(\rvy)\| \leq \| \rvx - \rvy \|$.
\end{definition}

\begin{remark}
\label{remark:contractive_operators}
    As shown in~\cite{Feldman2018privacy_amp_iter}, taking one gradient step of a convex and $L$-smooth objective $f$, i.e., $g(\rvx) = \rvx - \eta \nabla_{\rvx} f(\rvx)$, where the learning rate $\eta \leq 2/L$, is contractive. 
\end{remark}

\begin{definition}[Contractive Noisy Iteration (Definition 19 of~\cite{Feldman2018privacy_amp_iter})]
\label{def:cni}
    Given a random initial state $X_0 \in \gZ$, a sequence of contractive functions $g_t: \gZ \rightarrow \gZ$, and a sequence of noise distribution $\{\rho_t\}_{t=1}^{T}$, the contractive noisy iteration (CNI) is defined by the update rule: $X_{t+1} = g_{t+1}(X_t) + Z_{t+1}$,
    where $Z_{t+1}$, $\forall t\in [T]$, is drawn independently from $\rho_{t+1}$. The random variable output by this process after $T$ steps is denoted as $CNI(X_0, \{g_t\}_{t=1}^{T}, \{\rho_t\}_{t=1}^{T})$.  
\end{definition}

\begin{theorem}[Privacy Amplification by Iteration (Theorem 22 of~\cite{Feldman2018privacy_amp_iter} with Gaussian Noise)]
\label{thm:pabi}
    Let $X_T$ and $X_{T}'$ denote the output of $CNI_T(X_0, \{g_t\}_{t=1}^{T}, \{\rho_t\}_{t=1}^{T})$ and $CNI_T(X_0, \{g_t'\}_{t=1}^{T}, \{\rho_t\}_{t=1}^{T})$.
    Let $s_t := \sup_{\rvx} \|g_t(\rvx) - g_t'(\rvx)\|$, where $\rho_t \sim \gN(0, \sigma^2 \sI_d)$ for all $t$. Let $a_1, \dots, a_T$ be a sequence of reals and let $z_t := \sum_{i \leq t} s_i - \sum_{i\leq t}a_i$. If $z_t \geq 0$ for all $t$ and $z_T = 0$, then
    \begin{align*}
        \infdivalpha{X_T}{X_T'} \leq \sum_{t=1}^{T} \frac{\alpha a_t^2}{2 \sigma^2}
    \end{align*}
\end{theorem}

\subsection{Shuffled Gradient Methods}
\label{subsec:appendix_shuffled_gradient_methods}

We provide the pseudo-code of vanilla shuffled gradient methods (\SG) in Algorithm~\ref{alg:vanilla_shuffled_g}, including the three variants-Incremental Gradient (IG) methods, Shuffle Once (SO) and Random Resampling (RR). 

    \begin{algorithm}[H]
\caption{(Vanilla) Shuffled Gradient Methods}
\label{alg:vanilla_shuffled_g}
    \begin{algorithmic}
        \STATE Input: Initial point $\rvx_1^{(1)}$, learning rate $\eta$, number of epochs $K$. Dataset $\gD$ of size $n$. 
        \STATE \textit{IG}: Fix an order $\pi$, set $\pi^{(k)} = \pi, \forall k\in [K]$
        \STATE \textit{SO}: Generate permutation $\pi$ of $[n]$, set $\pi^{(k)} = \pi, \forall k \in [K]$
        \FOR{$k=1,2,\dots,K$}
            \STATE \textit{RR}: Generate permutation $\pi^{(k)}$ of $[n]$
            \FOR{$i=1,2,\dots,n$}
                \STATE $\rvx_{i+1}^{(k)} \leftarrow \rvx_i^{(k)} - \eta \nabla f(\rvx_i^{(k)}; \rvd_{\pi_i^{(k)}})$
            \ENDFOR
            \STATE $\rvx_{1}^{(k+1)} \leftarrow \rvx_{n+1}^{(k)}$
        \ENDFOR
        \RETURN $\rvx_1^{(K+1)}$
    \end{algorithmic}
\end{algorithm}

\section{Proof of Theorem~\ref{thm:convergence_generalized_shufflg_fm}}
\label{sec:appendix_proof_main_thm}

\textbf{Notation.}
In the proof, we denote the Bregman divergence induced by a real-valued convex function $g(\rvx): \R^d \rightarrow \R \cup \{+\infty\}$ as $B_g(\rvx, \rvy) = g(\rvx) - g(\rvy) - \langle \nabla g(\rvy), \rvx - \rvy\rangle, \forall \rvx, \rvy \in \R^d$, and $\text{dom}(g)$ denotes the domain of $g(\rvx)$.

We show convergence with more general assumptions on the smoothness and Lipschitzness constants.
We summarize important notations used in the proof and introduce the generalized assumptions as follows:
\begin{enumerate}[itemsep=0mm]
    \item Number of epochs: $K \geq 2$

    \item $\E_{A}\left[\cdot\right]$ denotes taking the expectation w.r.t. variable $A$. When the context is clear, $A$ is omitted.

    \item The target objective function: 
    \begin{align}
    \label{eq:appendix_true_objective_def}
        &G(\rvx) = G(\rvx; \gD) = F(\rvx; \gD) + \psi(\rvx)\\
        \nonumber
        &\text{where } \gD = \{\rvd_1,\dots,\rvd_n\}, \quad 
        F(\rvx) := F(\rvx; \gD) = \frac{1}{n}\sum_{i=1}^{n} f(\rvx; \rvd_i) = \frac{1}{n}\sum_{i=1}^{n} f_i(\rvx)
    \end{align}

    \item The optimum: $\rvx^* = \argmin_{\rvx\in\R^d} G(\rvx)$.

    Note that we always care about the convergence of the target objective function, i.e., $\E\left[G(\rvx; \gD)\right] - \E\left[G(\rvx^*; \gD)\right]$
    
    \item Optimization uncertainty: $\sigma_{any}^2 = \frac{1}{n}\sum_{i=1}^{n} \|\nabla f_i(\rvx^*)\|^2$.

    \item 
    Objective function used in the $k$-th epoch, under permutation $\pi^{(k)} \in \Pi_n$, for $k \in [K]$:
    \begin{align}
    \label{eq:appendix_surrogate_objective_def}
        &G^{(k)}(\rvx) = G(\rvx; \gD^{(k)} \cup \gP^{(k)}) = F(\rvx; \gD^{(k)} \cup \gP^{(k)}) + \psi(\rvx)
    \end{align}
    where
    \begin{itemize}[itemsep=0mm]
        \item  $\gD^{(k)} \in \{\emptyset\} \cup \{\{\rvd_{\pi_1^{(k}}^{(k)}, \dots, \rvd_{\pi_{n_d}^{(k)}}^{(k)}\}: 1\leq n_d^{(k)} \leq n\}$ is the private dataset used in epoch $k$, generated by first permuting $\gD$ and then taking the first $n_d^{(k)}$ samples
        \item $\gP^{(k)} \in  \{\emptyset\} \cup \{\{\rvp_1^{(k)}, \dots, \rvp_{n - n_d^{(k)}}^{(k)}\}\}$, $\gP^{(k)} \subseteq \gP$,
        is the public dataset used in epoch $k$
    \end{itemize}
    and
    \begin{align*}
        F^{(k)}(\rvx) = F(\rvx; \gD^{(k)} \cup \gP^{(k)}) 
        &
        = \frac{1}{n} \Big(
            \sum_{i=1}^{n_d^{(k)}} f(\rvx; \rvd_{\pi_i^{(k)}})
            + \sum_{i=n_d^{(k)} + 1}^{n} f(\rvx; \rvp_{i-n_d^{(k)}}^{(k)})
        \Big)\\
        &\quad = \frac{1}{n} \Big(
            \sum_{i=1}^{n_d^{(k)}} f_{\pi_i^{(k)}}(\rvx)
            + \sum_{i=n_d^{(k)} + 1}^{n} f_{i-n_d^{(k)}}^{(k, pub)}(\rvx)
        \Big)
    \end{align*}

    \item The objective difference for epoch $k\in [K]$: 
    \begin{align}
        H^{(k)}(\rvx) 
        &= G(\rvx;\gD) - G(\rvx;\gD^{(k)} \cup \gP^{(k)})
    \end{align}

    \item Smoothness:
    \begin{assumption}[Smoothness (Generalized Version of Assumption~\ref{ass:smoothness})]
    \label{ass:appendix_refined_smoothness}
    $f(\rvx; \rvd_i)$ is $L_i$-smooth, $\forall i\in [n]$ and $\rvd_i \in \gD$.
    $f(\rvx; \rvp_{j}^{(k)})$ is $\widetilde{L}_{j}^{(k)}$-smooth, $\forall j\in [n-n_d^{(k)}]$, $\rvp_j^{(k)}\in \gP$ and $\forall k\in [K]$.
    \end{assumption}
    
    \item The average smoothness constant 
    \begin{enumerate}[itemsep=0mm]
        \item of the target objective: $L = \frac{1}{n}\sum_{i=1}^{n} L_i$.
        \item of the objective used in the $k$-th epoch: 
        $\widehat{L}^{(k)} = \frac{1}{n}\Big( \sum_{i=1}^{n_d^{(k)}} L_{\pi_i^{(k)}} + \sum_{j=1}^{n-n_d^{(k)}} \widetilde{L}_{j}^{(k)} \Big)$.
    \end{enumerate}

    \item The maximum smoothness constant
    \begin{enumerate}[itemsep=0mm]
        \item of the target objective: $L^* = \max_{i\in[n]}\{L_i\}$.
        \item of the objective used in the $k$-th epoch: $\widehat{L}^{(k)*} = \max\{
        \{L_{\pi_i^{(k)}}\}_{i=1}^{n_d^{(k)}} \cup
        \{\widetilde{L}_{i}^{(k)}\}_{i=1}^{n-n_d^{(k)}} \}$
    \end{enumerate}

    \item The maximum average smoothness constant: $\bar{L}^{*} = \max\{L, \max_{k \in [K]}\widehat{L}^{(k)}\}$.

    \item Lipschitzness (only needed for privacy analysis):
    \begin{assumption}[Lipschitz Continuity (Re-statement of Assumption~\ref{ass:lipschitzness})]
        A convex function $f: \R^d \rightarrow \R$ is $\G$-Lipschitz if $\|\nabla f(\rvx)\| \leq \G$.
        $f(\rvx, \rvd)$ is $\G$-Lipschitz, $\forall \rvd\in\gD$;
        $f(\rvx; \rvp)$ is $\widetilde{G}$-Lipschitz, for all $\rvp \in \gP$.
    \end{assumption}

    \item The maximum Lipschitz parameter: $G^{*} = \max\{G, \widetilde{G}\}$

    \item Gaussian noise applied to the gradient at the $i$-th step in epoch $k$, for $i \in [n], k\in [K]$: $\rho_i^{(k)} \sim \gN(0, (\sigma^{(k)})^2\sI_d)$

\end{enumerate}

\textbf{Roadmap.} We begin by presenting useful lemmas used in the convergence proof in section~\ref{subsec:appendix_useful_lemmas}.
After that, we show the one epoch convergence section~\ref{subsec:appendix_one_epoch_convergence} and the expected one epoch convergence, taking into account the randomness due to data shuffling and noise injection, in section~\ref{subsec:appendix_expected_one_epoch_convergence}.
Finally, we give show the convergence bound across $K$ epochs in section~\ref{subsec:appendix_k_epoch_convergence}.

\subsection{Useful Lemmas}
\label{subsec:appendix_useful_lemmas}

\begin{lemma}[Stein's Lemma]
\label{lemma:steins_lemma}
    For a zero-mean isotropic Gaussian random variable $\rho \sim \gN(0, \sigma^2\sI_d)$, and a differentiable function $h: \R^{d} \rightarrow \R^d$, the following holds:
    \begin{align*}
        \E\left[ \langle \rho, h(\rho) \rangle \right]
        = \sigma^2 \E\left[\text{tr}(\nabla_{\rho} h(\rho))\right]
    \end{align*}
    where $\nabla h(\rho)$ is the Jacobian matrix of $h(\rho)$ and $\text{tr}(\cdot)$ denotes the trace operator.
\end{lemma}

\begin{lemma}[Lemma 3.6 of~\cite{liu2024last_iterate_shuffled_gradient}]
\label{lemma:breg_div_ub_lb}
    Given a convex and differentiable function $g(\rvx): \R^d \rightarrow \R$ satisfying 
    $\| \nabla g(\rvx) - \nabla g(\rvy) \| \leq L \|\rvx - \rvy\|$, $\forall \rvx, \rvy \in \R^d$ for some $L > 0$, then $\forall \rvx, \rvy \in \R^d$,
    \begin{align*}
        \frac{\| \nabla g(\rvx) - \nabla g(\rvy) \|^2}{2 L} \leq B_g(\rvx, \rvy) \leq \frac{L}{2}\|\rvx - \rvy\|^2
    \end{align*}
\end{lemma}

\begin{lemma}[Lemma E.1 of~\cite{liu2024last_iterate_shuffled_gradient}]
\label{lemma:opt_noise_bound}
    Under Assumption~\ref{ass:appendix_refined_smoothness}, for any permutation $\pi$ of $[n]$,
    \begin{align*}
        \frac{1}{n}\sum_{i=2}^{n} L_i \left\| \sum_{j=1}^{i-1} \nabla f_j(\rvx^*) \right\|^2 \leq n^2 L \sigma_{any}^2,
    \end{align*}
    where $L = \frac{1}{n}\sum_{i=1}^{n} L_i$. 
\end{lemma}

\begin{lemma}[Extension of Lemma 6.2 of~\cite{liu2024last_iterate_shuffled_gradient}]
\label{lemma:bregmandiv_relationship}
    Given two sequences of reals: $d^{(1)}, d^{(2)}, \dots, d^{(K)}, d^{(K+1)}$ and $e^{(1)}, e^{(2)}, \dots, e^{(K)}$, suppose there exist positive constants $a, b, c$ satisfying
    \begin{align}
    \label{eq:breg_div_ub}
        d^{(k+1)} \leq \frac{a}{k} + b(1+\log k) + c\sum_{l=2}^{k} \frac{d^{(l)}}{k-l+2} + \sum_{l=1}^{k} \frac{e^{(l)}}{k-l+1}, \quad \forall k\in [K]
    \end{align}
    then the following inequality holds
    \begin{align}
    \label{eq:breg_div_rel}
        d^{(k+1)} \leq \Big(\frac{a}{k} + b(1+\log k) + M \Big)\sum_{i=0}^{k-1} (2c(1+\log k))^{i}
    \end{align}
    where $M := \max_{k\in [K]}\sum_{l=1}^{k} \frac{e^{(l)}}{k-l+1}$.
\end{lemma}

\begin{proof}
    We use induction to show Eq.~\ref{eq:breg_div_rel}. 
    
    Base Case: for $k = 1$, 
    by Eq.~\ref{eq:breg_div_ub} and the definition of $M$,
    $d^{(2)}\leq a+b + e^{(1)} \leq a + b + M$, which also satisfies Eq.~\ref{eq:breg_div_rel}.

    Induction Hypothesis: suppose Eq.~\ref{eq:breg_div_rel} holds for 1 to $k-1$ (where $2 \leq k \leq K$), i.e., 
    \begin{align}
        d^{(l)} \leq \Big(\frac{a}{l-1} + b(1+\log (l-1)) + M \Big)\sum_{i=0}^{l-2} (2c(1+\log (l-1)))^{i}
    \end{align}
    which implies
    \begin{align}
        d^{(l)} \leq \Big(\frac{a}{l-1} + b(1+\log k) + M \Big)\sum_{i=0}^{l-2} (2c(1+\log k))^{i}
    \end{align}

    Now for $d^{(k+1)}$, by Eq.~\ref{eq:breg_div_ub},
    \begin{align}
        d^{(k+1)} &\leq \frac{a}{k} + b(1+\log k) + c\sum_{l=2}^{k} \frac{d^{(l)}}{k-l+2} + \sum_{l=1}^{k} \frac{e^{(l)}}{k-l+1}\\
        &\leq \frac{a}{k} + b(1+\log k) + c\sum_{l=2}^{k} \frac{d^{(l)}}{k-l+2} + M\\
    \label{eq:breg_div_rel_interm1}
        &\leq \frac{a}{k} + ac \sum_{l=2}^{k} \sum_{i=0}^{i-2} \frac{(2c(1+\log k))^i}{(k-l+2) (l-1)} \\
        \nonumber
        &\quad + \Big( b(1+\log k) + M \Big) \Big(1 + c \sum_{l=2}^{k} \sum_{i=0}^{l-2} \frac{(2c(1+\log k))^i}{k-l+2} \Big)
    \end{align}

    Note that
    \begin{align}
        c \sum_{l=2}^{k} \sum_{i=0}^{i-2} \frac{(2c(1+\log k))^i}{(k-l+2) (l-1)}
        &= c \sum_{i=0}^{k-2} (2c(1+\log k))^i \Big(\sum_{l=2+i}^{k} \frac{1}{(k-l+2)(l-1)} \Big)\\
        &= \frac{c}{k+1}\sum_{i=0}^{k-2} (2c(1+\log k))^{i} \Big(
        \sum_{l=2+i}^{k} \frac{1}{k-l+2} + \frac{1}{l-1}
        \Big)\\
        &\leq \frac{c}{k+1}\sum_{i=0}^{k-2} (2c(1+\log k))^{i}
        \sum_{l=1}^{k} \frac{2}{l}\\
        &\leq \frac{\sum_{i=0}^{k-2} (2c(1+\log k))^{i+1}}{k+1}\\
        &\leq \frac{\sum_{i=0}^{k-1} (2c(1+\log k))^i}{k+1}\\
    \label{eq:breg_div_rel_interm2}
        &\leq \frac{\sum_{i=1}^{k-1} (2c (1+\log k))^i}{k}
    \end{align}
    and
    \begin{align}
        c \sum_{l=2}^{k}\sum_{i=0}^{l-2} \frac{(2c(1+\log k))^i}{k-l+2}
        &= c \sum_{i=0}^{k-2} (2c(1+\log k))^i
        \sum_{l=2+i}^{k} \frac{1}{k-l+2}
        \leq c \sum_{i=0}^{k-2} (2c(1+\log k))^i
        \sum_{l=1}^{k} \frac{1}{l}\\
        &\leq c(1+\log k)\sum_{i=0}^{k-2} (2c(1+\log k))^{i} 
        \leq \sum_{i=0}^{k-2} (2c(1+\log k))^{i+1}\\
    \label{eq:breg_div_rel_interm3}
        &\leq \sum_{i=1}^{k-1} (2c(1+\log k))^{i}
    \end{align}

    Combining Eq.~\ref{eq:breg_div_rel_interm1}, Eq.~\ref{eq:breg_div_rel_interm2} and Eq.~\ref{eq:breg_div_rel_interm3}, 
    \begin{align}
        d^{(k+1)}
        &\leq \frac{a}{k} + \frac{a}{k} \sum_{i=1}^{k-1}(2c(1+\log k))^{i}
        + \Big(b (1+\log k) + M \Big)\Big( 1 + \sum_{i=1}^{k-1}(2c(1+\log k))^{i} \Big)\\
        &= \Big( \frac{a}{k} + b(1+\log k) + M \Big) \sum_{i=0}^{k-1} (2c(1+\log k))^{i}
    \end{align}
    which finishes the induction.
    
\end{proof}

\subsection{One Epoch Convergence}
\label{subsec:appendix_one_epoch_convergence}
The following lemma  is a generalization of Lemma D.1 of~\cite{liu2024last_iterate_shuffled_gradient} from two dimensions: allowing the usage of surrogate objectives and adding additional noise for privacy preservation. 

\begin{lemma}
\label{lemma:one_epoch_bg_3pid}
    Under Assumptions~\ref{ass:convexity} and~\ref{ass:reg},
    for any epoch $k \in [K]$, permutation $\pi^{(k)}$ and $\rvz\in \R^d$, Algorithm~\ref{alg:generalized_shuffled_gradient_fm} guarantees
    \begin{align*}
        &G(\rvx_1^{(k+1)}) - G(\rvz)\\
        \nonumber
        &\leq H^{(k)}(\rvx_1^{(k+1)}) - H^{(k)}(\rvz)
        + \frac{\|\rvz - \rvx_1^{(k)}\|^2}{2n \eta} - (\frac{1}{2n \eta} + \frac{\mu_\psi}{2}) \|\rvz - \rvx_1^{(k+1)}\|^2
        - \frac{1}{2n \eta}\|\rvx_1^{(k+1)} - \rvx_1^{(k)}\|^2\\
        \nonumber
        &\quad + \frac{1}{n} \Big(\sum_{i=1}^{n_d^{(k)}}\Big(
            B_{f_{\pi_i^{(k)}}}(\rvx_1^{(k+1)}, \rvx_i^{(k)}) - B_{f_{\pi_i^{(k)}}}(\rvz, \rvx_i^{(k)})
        \Big)\\
        \nonumber
        &\quad + \sum_{i=n_d^{(k)}+1}^{n}\Big(
            B_{f_{i-n_d^{(k)}}^{(k, pub)}}(\rvx_1^{(k+1)}, \rvx_i^{(k)})
            - B_{f_{i-n_d^{(k)}}^{(k, pub)}}(\rvz, \rvx_i^{(k)})
        \Big)\Big)
        + \frac{1}{n}\sum_{i=1}^{n}\langle -\rho_i^{(k)}, \rvx_1^{(k+1)} - \rvz\rangle. 
    \end{align*}
\end{lemma}

\begin{proof}[Proof of Lemma~\ref{lemma:one_epoch_bg_3pid}]
    It suffices to only consider $\rvx \in \text{dom}(\psi)$. 

    Let $\rvg^{(k)} = \sum_{i=1}^{n_d^{(k)}}\Big( \nabla f_{\pi_i^{(k)}} (\rvx_i^{(k)}) + \rho_i^{(k)}\Big)
    + \sum_{i=n_d^{(k)}+1}^{n} \Big( \nabla f_{i-n_d^{(k)}}^{(k, pub)}(\rvx_{i}^{(k)}) + \rho_i^{(k)}\Big)$.

    According to the update rule in \ref{alg:generalized_shuffled_gradient_fm}, $\rvx_{n+1}^{(k)} = \rvx_1^{(k)} - \eta \cdot \rvg^{(k)}$. Observe that
    \begin{align*}
        \rvx_1^{(k+1)} 
        &= \argmin_{\rvx \in \R^d} \Big\{ n \psi(\rvx) + \frac{\| \rvx - \rvx_{n+1}^{(k)}\|^2}{2\eta} \Big\}
        = \argmin_{\rvx \in \R^d} \Big\{ n\psi(\rvx) + \frac{\| \rvx - \rvx_1^{(k)} + \eta \cdot \rvg^{(k)}\|^2}{2 \eta} \Big\}\\
        &= \argmin_{\rvx\in\R^d}\Big\{ n \psi(\rvx) + \frac{\|\rvx - \rvx_1^{(k)}\|^2 + \eta^2 \|\rvg^{(k)}\|^2 + 2 \langle \rvx - \rvx_1^{(k)}, \eta \rvg^{(k)} \rangle}{2 \eta}
        \Big\}\\
        &= \argmin_{\rvx \in \R^d}\Big\{ n \psi(\rvx) + \frac{\|\rvx - \rvx_1^{(k)}\|^2}{2\eta} + \langle \rvx - \rvx_1^{(k)}, \rvg^{(k)}\rangle
        \Big\}
    \end{align*}

    By the first-order optimality condition, there exists some vector $\nabla \psi(\rvx_1^{(k+1)})$ 
    in the subgradient of $\psi(\rvx_1^{(k+1)})$ such that 
    \begin{align*}
        n \nabla \psi (\rvx_1^{(k+1)}) + \rvg^{(k)} + \frac{\rvx_1^{(k+1)} - \rvx_1^{(k)}}{\eta} = \mathbf{0}
        \iff \rvg^{(k)} = -n \nabla \psi(\rvx_1^{(k+1)}) + \frac{\rvx_1^{(k)} - \rvx_1^{(k+1)}}{\eta}
    \end{align*}

    Therefore, for $\rvz \in \text{dom}(\psi)$,
    \begin{align}
    \nonumber
        &\langle \rvg^{(k)}, \rvx_1^{(k+1)} - \rvz \rangle\\
        \nonumber
        &= n \langle \nabla \psi(\rvx_1^{(k+1)}), \rvz - \rvx_1^{(k+1)}\rangle
        + \frac{1}{\eta}\langle \rvx_1^{(k)} - \rvx_1^{(k+1)}, \rvx_1^{(k+1)} - \rvz\rangle\\
    \nonumber
        &\stackrel{\text{(a)}}{\leq} n\Big( \psi(\rvz) - \psi(\rvx_1^{(k+1)}) - \frac{\mu_\psi}{2}\|\rvz - \rvx_1^{(k+1)}\|^2\Big)
        + \frac{1}{\eta}\langle \rvx_1^{(k)} - \rvx_1^{(k+1)}, \rvx_1^{(k+1)} - \rvz\rangle\\
    \nonumber
        &= n\Big( \psi(\rvz) - \psi(\rvx_1^{(k+1)}) - \frac{\mu_\psi}{2}\|\rvz - \rvx_1^{(k+1)}\|^2\Big)
        + \frac{1}{2\eta}\Big(\|\rvz - \rvx_1^{(k)}\|^2 - \|\rvz - \rvx_1^{(k+1)}\|^2 - \|\rvx_1^{(k+1)} - \rvx_1^{(k)}\|^2 \Big)\\
    \label{eq:dot_form_1}
        &= n\Big( \psi(\rvz) - \psi(\rvx_1^{(k+1)})\Big)
        + \frac{\|\rvz - \rvx_1^{(k)}\|^2}{2\eta} - (\frac{1}{2\eta} + \frac{n\mu_\psi}{2}) \|\rvz - \rvx_1^{(k+1)}\|^2
        - \frac{1}{2\eta}\|\rvx_1^{(k+1)} - \rvx_1^{(k)}\|^2
    \end{align}
    where (a) is by Assumption~\ref{ass:reg} on the $\mu_\psi$-strong convexity of $\psi$.

     By the definition of $\rvg^{(k)}$,
    \begin{align}
    \nonumber
        &\langle \rvg^{(k)}, \rvx_1^{(k+1)} - \rvz\rangle
        = \langle \sum_{i=1}^{n_d^{(k)}}\Big( \nabla f_{\pi_i^{(k)}}(\rvx_i^{(k)}) 
        + \rho_i^{(k)}\Big)
        + \sum_{i=n_d^{(k)}+1}^{n} \Big(
            \nabla f_{i-n_d^{(k)}}^{(k, pub)}(\rvx_i^{(k)})
            + \rho_i^{(k)}
        \Big)
        , \rvx_1^{(k+1)} - \rvz \rangle\\
    \label{eq:dot_interm}
        &= \sum_{i=1}^{n_d^{(k)}} \langle \nabla f_{\pi_i^{(k)}}(\rvx_i^{(k)}), \rvx_1^{(k+1)} - \rvz \rangle 
        + \sum_{i=n_d^{(k)}+1}^{n} \langle \nabla f_{i-n_d^{(k)}}^{(k, pub)}(\rvx_i^{(k)}),
        \rvx_1^{(k+1)} - \rvz
        \rangle
        + \sum_{i=1}^{n}\langle \rho_i^{(k)}, \rvx_1^{(k+1)} - \rvz\rangle
    \end{align}

    Since for $i \leq n_d^{(k)}$,
    \begin{align*}
        &B_{f_{\pi_i^{(k)}}}(\rvx_1^{(k+1)}, \rvx_i^{(k)}) 
        = f_{\pi_i^{(k)}}(\rvx_1^{(k+1)}) - f_{\pi_i^{(k)}}(\rvx_i^{(k)}) - \langle \nabla f_{\pi_i^{(k)}}(\rvx_i^{(k)}), \rvx_1^{(k+1)} - \rvx_i^{(k)}\rangle\\
        &B_{f_{\pi_i^{(k)}}}(\rvz, \rvx_i^{(k)}) 
        = f_{\pi_i^{(k)}}(\rvz) - f_{\pi_i^{(k)}}(\rvx_i^{(k)}) - \langle \nabla f_{\pi_i^{(k)}}(\rvx_i^{(k)}), \rvz - \rvx_i^{(k)}\rangle
    \end{align*}
    and for $n_d^{(k)} < i \leq n$,
    \begin{align*}
        &B_{f_{i-n_d^{(k)}}^{(k, pub)}}(\rvx_1^{(k+1)}, \rvx_i^{(k)})
        = f_{i-n_d^{(k)}}^{(k, pub)}(\rvx_1^{(k+1)}) 
        - f_{i-n_d^{(k)}}^{(k, pub)}(\rvx_i^{(k)})
        - \langle \nabla f_{i-n_d^{(k)}}^{(k, pub)}(\rvx_i^{(k)}), \rvx_1^{(k+1)} - \rvx_i^{(k)}\rangle\\
        &B_{f_{i-n_d^{(k)}}^{(k, pub)}}(\rvz, \rvx_i^{(k)}) = f_{i-n_d^{(k)}}^{(k, pub)}(\rvz) - f_{i-n_d^{(k)}}^{(k, pub)}(\rvx_i^{(k)})
        - \langle \nabla f_{i-n_d^{(k)}}^{(k, pub)}(\rvx_i^{(k)}),
        \rvz - \rvx_i^{(k)} \rangle
    \end{align*}
    there is for $i \leq n_d^{(k)}$,
    \begin{align}
    \label{eq:dot_to_bregmandiv_priv}
        &\sum_{i=1}^{n_d^{(k)}}\langle \g f_{\pi_i^{(k)}}(\rvx_i^{(k)}), \rvx_1^{(k+1)} - \rvz\rangle \\
        \nonumber
        &= \sum_{i=1}^{n_d^{(k)}}\Big(f_{\pi_i^{(k)}}(\rvx_1^{(k+1)}) - f_{\pi_i^{(k)}}(\rvz) - B_{f_{\pi_i^{(k)}}}(\rvx_1^{(k+1)}, \rvx_i^{(k)}) + B_{f_{\pi_i^{(k)}}}(\rvz, \rvx_i^{(k)}) \Big)
    \end{align}
    and for $n_d^{(k)} < i \leq n$,
    \begin{align}
    \label{eq:dot_to_bregmandiv_pub}
        &\sum_{i=n_d^{(k)}+1}^{n}\langle \g f_{i-n_d^{(k)}}^{(k, pub)}(\rvx_i^{(k)}), 
        \rvx_1^{(k+1)} - \rvz\rangle\\
    \nonumber
        &= \sum_{i=n_d^{(k)}+1}^{n}\Big(
            f_{i-n_d^{(k)}}^{(k, pub)}(\rvx_1^{(k+1)})
            - f_{i-n_d^{(k)}}^{(k, pub)}(\rvz)
            - B_{f_{i-n_d^{(k)}}^{(k, pub)}}(\rvx_1^{(k+1)}, \rvx_i^{(k)})
            + B_{f_{i-n_d^{(k)}}^{(k, pub)}}(\rvz, \rvx_i^{(k)})
        \Big)
    \end{align}
    Therefore, summing up Eq.~\ref{eq:dot_to_bregmandiv_priv} and Eq.~\ref{eq:dot_to_bregmandiv_pub}, we have
    \begin{align}
    \nonumber
        &\sum_{i=1}^{n_d^{(k)}} \langle \g f_{\pi_i^{(k)}}(\rvx_i^{(k)}), \rvx_1^{(k+1)} - \rvz\rangle
        + \sum_{i=n_d^{(k)}+1}^{n} \langle \g
        f_{i-n_d^{(k)}}^{(k, pub)}(\rvx_i^{(k)}),
        \rvx_1^{(k+1)} - \rvz\rangle\\
        \nonumber
        &= \sum_{i=1}^{n_d^{(k)}}\Big(f_{\pi_i^{(k)}}(\rvx_1^{(k+1)}) - f_{\pi_i^{(k)}}(\rvz)\Big)
        + \sum_{i=n_d^{(k)}+1}^{n}\Big(
            f_{i-n_d^{(k)}}^{(k, pub)}(\rvx_1^{(k+1)})
            - f_{i-n_d^{(k)}}^{(k, pub)}(\rvz)
        \Big)\\
        \nonumber
        &\quad - \sum_{i=1}^{n_d^{(k)}}\Big(
            B_{f_{\pi_i^{(k)}}}(\rvx_1^{(k+1)}, \rvx_i^{(k)}) - B_{f_{\pi_i^{(k)}}}(\rvz, \rvx_i^{(k)})
        \Big)
        - \sum_{i=n_d^{(k)}+1}^{n}\Big(
            B_{f_{i-n_d^{(k)}}^{(k, pub)}}(\rvx_1^{(k+1)}, \rvx_i^{(k)})
            - B_{f_{i-n_d^{(k)}}^{(k, pub)}}(\rvz, \rvx_i^{(k)})
        \Big)\\
    \label{eq:dot_to_bregmandiv}
        &= n F^{(k)}(\rvx_1^{(k+1)}) - n F^{(k)}(\rvz)
        - \sum_{i=1}^{n_d^{(k)}}\Big(
            B_{f_{\pi_i^{(k)}}^{(k, priv)}}(\rvx_1^{(k+1)}, \rvx_i^{(k)}) - B_{f_{\pi_i^{(k)}}^{(k, priv)}}(\rvz, \rvx_i^{(k)})
        \Big)\\
        \nonumber
        &\quad - \sum_{i=n_d^{(k)}+1}^{n}\Big(
            B_{f_{i-n_d^{(k)}}^{(k, pub)}}(\rvx_1^{(k+1)}, \rvx_i^{(k)})
            - B_{f_{i-n_d^{(k)}}^{(k, pub)}}(\rvz, \rvx_i^{(k)})
        \Big)
    \end{align}

    Hence, plugging Eq.~\ref{eq:dot_to_bregmandiv} back to Eq.~\ref{eq:dot_interm}, there is
    \begin{align}
    \label{eq:dot_form_2}
        &\langle \rvg^{(k)}, \rvx_1^{(k+1)} - \rvz\rangle
        = n F^{(k)}(\rvx_1^{(k+1)}) - n F^{(k)}(\rvz)
        + \sum_{i=1}^{n}\langle \rho_i^{(k)}, \rvx_1^{(k+1)} - \rvz\rangle\\
        \nonumber
        &\quad - \sum_{i=1}^{n_d^{(k)}}\Big(
            B_{f_{\pi_i^{(k)}}}(\rvx_1^{(k+1)}, \rvx_i^{(k)}) - B_{f_{\pi_i^{(k)}}}(\rvz, \rvx_i^{(k)})
        \Big)
        - \sum_{i=n_d^{(k)}+1}^{n}\Big(
            B_{f_{i-n_d^{(k)}}^{(k, pub)}}(\rvx_1^{(k+1)}, \rvx_i^{(k)})
            - B_{f_{i-n_d^{(k)}}^{(k, pub)}}(\rvz, \rvx_i^{(k)})
        \Big)
    \end{align}

    Recall that $G(\rvx) = F(\rvx; \gD) + \psi(\rvx)$ is the target objective (see Eq.~\ref{eq:appendix_true_objective_def}) and $G^{(k)}(\rvx) = F^{(k)}(\rvx; \gD^{(k)} \cup \gP^{(k)}) + \psi(\rvx)$ (see Eq.~\ref{eq:appendix_surrogate_objective_def}) is the objective used in the $k$-th epoch during optimization for $k\in [K]$. 
    
    Now, by Eq.~\ref{eq:dot_form_1} and Eq.~\ref{eq:dot_form_2}, after rearranging
    \begin{align*}
        &G^{(k)}(\rvx_1^{(k+1)}) - G^{(k)}(\rvz)
        \leq \frac{\|\rvz - \rvx_1^{(k)}\|^2}{2n \eta} - (\frac{1}{2n \eta} + \frac{\mu_\psi}{2}) \|\rvz - \rvx_1^{(k+1)}\|^2
        - \frac{1}{2n \eta}\|\rvx_1^{(k+1)} - \rvx_1^{(k)}\|^2\\
        \nonumber
        &\quad + \frac{1}{n} \sum_{i=1}^{n_d^{(k)}}\Big(
            B_{f_{\pi_i^{(k)}}}(\rvx_1^{(k+1)}, \rvx_i^{(k)}) - B_{f_{\pi_i^{(k)}}}(\rvz, \rvx_i^{(k)})
        \Big)\\
        \nonumber
         &\quad + \frac{1}{n} \sum_{i=n_d^{(k)}+1}^{n}\Big(
            B_{f_{i-n_d^{(k)}}^{(k, pub)}}(\rvx_1^{(k+1)}, \rvx_i^{(k)})
            - B_{f_{i-n_d^{(k)}}^{(k, pub)}}(\rvz, \rvx_i^{(k)})
        \Big)
        + \frac{1}{n}\sum_{i=1}^{n}\langle -\rho_i^{(k)}, \rvx_1^{(k+1)} - \rvz\rangle
    \end{align*}

    And following the above, for any $\rvz\in \R^d$ and $s\in [K]$, 
    \begin{align*}
        &G(\rvx_1^{(k+1)}) - G(\rvz)
        =\left( G^{(k)}(\rvx_1^{(k+1)}) - G^{(k)}(\rvz) \right)
        + \left( G(\rvx_1^{(k+1)}) - G^{(k)}(\rvx_1^{(k+1)}) \right)
        -  \left( G(\rvz) - G^{(k)}(\rvz) \right)\\
        &\leq H^{(k)}(\rvx_1^{(k+1)}) - H^{(k)}(\rvz)
        + \frac{\|\rvz - \rvx_1^{(k)}\|^2}{2n \eta} - (\frac{1}{2n \eta} + \frac{\mu_\psi}{2}) \|\rvz - \rvx_1^{(k+1)}\|^2
        - \frac{1}{2n \eta}\|\rvx_1^{(k+1)} - \rvx_1^{(k)}\|^2\\
        \nonumber
        &\quad + \frac{1}{n} \sum_{i=1}^{n_d^{(k)}}\Big(
            B_{f_{\pi_i^{(k)}}}(\rvx_1^{(k+1)}, \rvx_i^{(k)}) - B_{f_{\pi_i^{(k)}}}(\rvz, \rvx_i^{(k)})
        \Big)\\
        \nonumber
        &\quad + \frac{1}{n} \sum_{i=n_d^{(k)}+1}^{n}\Big(
            B_{f_{i-n_d^{(k)}}^{(k, pub)}}(\rvx_1^{(k+1)}, \rvx_i^{(k)})
            - B_{f_{i-n_d^{(k)}}^{(k, pub)}}(\rvz, \rvx_i^{(k)})
        \Big) 
        + \frac{1}{n}\sum_{i=1}^{n}\langle -\rho_i^{(k)}, \rvx_1^{(k+1)} - \rvz\rangle
    \end{align*}

\end{proof}

The following lemma is a generalization of Lemma D.2 of~\cite{liu2024last_iterate_shuffled_gradient} from two dimensions: allowing the usage of surrogate objectives and adding additional noise for privacy preservation.

\begin{lemma}
\label{lemma:one_epoch_bound_bregmandiv}
    Under Assumptions~\ref{ass:convexity},~\ref{ass:dissim_partial_lipschitzness} and~\ref{ass:appendix_refined_smoothness}, for any epoch $k \in [K]$, permutation $\pi^{(k)}$ and $\rvz\in\R^d$, if the learning rate $\eta \leq \frac{1}{n \sqrt{10 \widehat{L}^{(k)} \widehat{L}^{(k)*}}}$, Algorithm~\ref{alg:generalized_shuffled_gradient_fm} guarantees
     \begin{align*}
        &\frac{1}{n} \Big(\sum_{i=1}^{n_d^{(k)}}
            \Big( B_{f_{\pi_i^{(k)}}}(\rvx_{1}^{(k+1)}, \rvx_i^{(k)})
            - B_{f_{\pi_i^{(k)}}}(\rvz, \rvx_i^{(k)}) \Big)
            + \sum_{i=n_d^{(k)}+1}^{n}\Big(
                B_{f_{i-n_d^{(k)}}^{(k,pub)}}(\rvx_1^{(k+1)}, \rvx_i^{(k)})
                - B_{f_{i-n_d^{(k)}}^{(k, pub)}}(\rvz, \rvx_i^{(k)})
            \Big)
        \Big) \\
        \nonumber
        &\leq \widehat{L}^{(k)} \|\rvx_1^{(k+1)} - \rvx_1^{(k)} \|^2
        + 10\eta^2 n^2 \widehat{L}^{(k)} L B_F(\rvz, \rvx^{*})\\
        \nonumber
        &\quad + 5 \eta^2 \frac{1}{n} \Big(\sum_{i=2}^{n_d^{(k)}} L_{\pi_i^{(k)}} \Big\| \sum_{j=1}^{i-1} \nabla f_{\pi_j^{(k)}}(\rvx^*)\Big\|^2
        + \sum_{i=n_d^{(k)}+1}^{n}\widetilde{L}_{i-n_d^{(k)}}^{(k)}\Big\| \sum_{j=1}^{i-1}\nabla f_{\pi_j^{(k)}}(\rvx^*)\Big\|^2 \Big)\\
        \nonumber
        &\quad + 5\eta^2 \frac{1}{n} \Big( \sum_{i=2}^{n_d^{(k)}} L_{\pi_i^{(k)}} \Big\| \sum_{j=1}^{i-1} \rho_j^{(k)}\Big\|^2
        + \sum_{i=n_d^{(k)}+1}^{n} \widetilde{L}_{i-n_d^{(k)}}^{(k)}\Big\| \sum_{j=1}^{i-1} \rho_j^{(k)}\Big\|^2\Big)\\
        &\quad + 5\eta^2 L^{(k)*} \frac{1}{n} \sum_{i=n_d^{(k)} + 1}^{n} \Big\|
        \sum_{j=n_d^{(k)}+1}^{i-1} \nabla f_{i-n_d^{(k)}}^{(k, pub)}(\rvz) -
        \sum_{j=n_d^{(k)}+1}^{i-1} \nabla f_{\pi_j^{(k)}}(\rvz)
        \Big\|^2
    \end{align*}
    
\end{lemma}

\begin{proof}[Proof of Lemma~\ref{lemma:one_epoch_bound_bregmandiv}]

    By Lemma~\ref{lemma:breg_div_ub_lb}, for $i \leq n_d^{(k)}$, and permutation $\pi_i^{(k)} \in \Pi_n$,
    \begin{align*}
        &B_{f_{\pi_i^{(k)}}} (\rvx_{1}^{(k+1)}, \rvx_i^{(k)}) \leq \frac{ L_{\pi_i^{(k)}}}{2} \|\rvx_{1}^{(k+1)} - \rvx_{i}^{(k)}\|^2
        \leq L_{\pi_i^{(k)}} \Big( \|\rvx_{1}^{(k+1)} - \rvx_1^{(k)}\|^2 + \|\rvx_i^{(k)} - \rvx_1^{(k)} \|^2 \Big)\\
        &B_{f_{\pi_i^{(k)}}} (\rvz, \rvx_i^{(k)})
        \geq \frac{\left\|\nabla f_{\pi_i^{(k)}}(\rvx_i^{(k)}) - \nabla f_{\pi_i^{(k)}}(\rvz) \right\|^2}{2 L_{\pi_i^{(k)}}}
    \end{align*}
    and for $n_d^{(k)} < i \leq n$, 
    \begin{align*}
        &B_{f_{i-n_d^{(k)}}^{(k, pub)}}(\rvx_1^{(k+1)}, \rvx_i^{(k)})
        \leq \frac{\widetilde{L}_{i-n_d^{(k)}}^{(k)}}{2}
        \|\rvx_1^{(k+1)} - \rvx_i^{(k)}\|^2
        \leq \widetilde{L}_{i-n_d^{(k)}}^{(k)}\Big(
            \|\rvx_1^{(k+1)} - \rvx_1^{(k)}\|^2
            + \|\rvx_i^{(k)} - \rvx_1^{(k)}\|^2
        \Big)\\
        &B_{f_{i-n_d^{(k)}}^{(k, pub)}}(\rvz, \rvx_i^{(k)})
        \geq \frac{\left\| \nabla f_{i-n_d^{(k)}}^{(k, pub)}(\rvx_i^{(k)}) - \nabla f_{i-n_d^{(k)}}^{(k, pub)}(\rvz) \right\|^2}{2\widetilde{L}_{i-n_d^{(k)}}^{(k)}}
    \end{align*}
    Therefore,
    \begin{align}
    \nonumber
        &\frac{1}{n} \Big(\sum_{i=1}^{n_d^{(k)}}
        \Big( B_{f_{\pi_i^{(k)}}}(\rvx_{1}^{(k+1)}, \rvx_i^{(k)})
        - B_{f_{\pi_i^{(k)}}}(\rvz, \rvx_i^{(k)}) \Big)
        + \sum_{i=n_d^{(k)}+1}^{n}\Big(
            B_{f_{i-n_d^{(k)}}^{(k,pub)}}(\rvx_1^{(k+1)}, \rvx_i^{(k)})
            - B_{f_{i-n_d^{(k)}}^{(k, pub)}}(\rvz, \rvx_i^{(k)})
        \Big)
        \Big) \\
        \nonumber
        &\leq \frac{1}{n}\sum_{i=1}^{n_d^{(k)}}
        \Big( L_{\pi_i^{(k)}} \Big(\|\rvx_{1}^{(k+1)} - \rvx_1^{(k)}\|^2 + \|\rvx_i^{(k)} - \rvx_1^{(k)}\|^2 \Big)
        - \frac{ \left\|\nabla f_{\pi_i^{(k)}}(\rvx_i^{(k)}) - \nabla f_{\pi_i^{(k)}}(\rvz) \right\|^2}{2 L_{\pi_i^{(k)}}}
        \Big)\\
        \nonumber
        &\quad + \frac{1}{n}\sum_{i=n_d^{(k)}+1}^{n}\Big(
            \widetilde{L}_{i-n_d^{(k)}}^{(k)} \Big( \|\rvx_1^{(k+1)} - \rvx_1^{(k)} \|^2 + \|\rvx_i^{(k)} - \rvx_1^{(k)}\|^2\Big)
        - \frac{\left\| \nabla f_{i-n_d^{(k)}}^{(k, pub)}(\rvx_i^{(k)}) - \nabla f_{i-n_d^{(k)}}^{(k, pub)}(\rvz) \right\|^2}{2\widetilde{L}_{i-n_d^{(k)}}^{(k)}}
        \Big)\\
    \label{eq:breg_diff_interm_2}
        &= \widehat{L}^{(k)} \|\rvx_{1}^{(k+1)} - \rvx_1^{(k)}\|^2 
        + \frac{1}{n} \Big( \underbrace{\sum_{i=1}^{n_d^{(k)}}
        L_{\pi_i^{(k)}} \|\rvx_i^{(k)} - \rvx_1^{(k)}\|^2}_{:= I_1}
        + \underbrace{\sum_{i=n_d^{(k)}+1}^{n} \widetilde{L}_{i-n_d^{(k)}}^{(k)}\|\rvx_i^{(k)} - \rvx_1^{(k)}\|^2}_{:= I_2}
        \Big)\\
        \nonumber
        &\quad - \frac{1}{n} \Big( 
        \sum_{i=1}^{n_d^{(k)}} \frac{ \left\|\nabla f_{\pi_i^{(k)}}(\rvx_i^{(k)}) - \nabla f_{\pi_i^{(k)}}(\rvz) \right\|^2}{2 L_{\pi_{i}^{(k)}}}
        + \sum_{i=n_d^{(k)}+1}^{n} \frac{\left\| \nabla f_{i-n_d^{(k)}}^{(k, pub)}(\rvx_i^{(k)}) - \nabla f_{i-n_d^{(k)}}^{(k, pub)}(\rvz) \right\|^2}{2\widetilde{L}_{i-n_d^{(k)}}^{(k)}}
        \Big)
    \end{align}
    where recall that $\widehat{L}^{(k)} = \frac{1}{n}\sum_{i=1}^{n} \left( \sum_{i=1}^{n_d^{(k)}} L_{\pi_i^{(k)}} + \sum_{j=1}^{n-n_d^{(k)}} \widetilde{L}_j^{(k)}  \right)$.
    Now we bound terms $I_1 \triangleq \sum_{i=1}^{n_d^{(k)}} L_{\pi_i^{(k)}} \|\rvx_i^{(k)} - \rvx_1^{(k)}\|^2$ (in Part I) and $I_2 \triangleq \sum_{i=n_d^{(k)}+1}^{n} \widetilde{L}_{i-n_d^{(k)}}^{(k)} \|\rvx_i^{(k)} - \rvx_1^{(k)}\|^2$ (in Part II) as follows:
    
    \textbf{Part I: } For $i \leq n_d^{(k)}$,
    \begin{align}
        I_1 &\triangleq \sum_{i=1}^{n_d^{(k)}} L_{\pi_i^{(k)}} \|\rvx_i^{(k)} - \rvx_1^{(k)}\|^2
        = \sum_{i=2}^{n_d^{(k)}} L_{\pi_i^{(k)}} \|\rvx_i^{(k)} - \rvx_1^{(k)}\|^2
        = \sum_{i=2}^{n_d^{(k)}} L_{\pi_i^{(k)}} \eta^2 \Big\| \sum_{j=1}^{i-1} (\nabla f_{\pi_j^{(k)}}(\rvx_j^{(k)}) + \rho_j^{(k)})  \Big\|^2 \tag{From the update in \ref{alg:generalized_shuffled_gradient_fm}}\\
    \nonumber
        &= \eta^2 \sum_{i=2}^{n_d^{(k)}} L_{\pi_i^{(k)}} \Big\| \sum_{j=1}^{i-1} \Big(\nabla f_{\pi_j^{(k)}}(\rvx_j^{(k)}) 
        - \nabla f_{\pi_j^{(k)}}(\rvz) 
        + \nabla f_{\pi_j^{(k)}}(\rvz) - \nabla f_{\pi_j^{(k)}}(\rvx^*)
        + \nabla f_{\pi_j^{(k)}}(\rvx^*) + \rho_j^{(k)}
        \Big)\Big\|^2\\
    \label{eq:breg_diff_interm_1}
        &\leq \eta^2 \sum_{i=2}^{n_d^{(k)}} L_{\pi_i^{(k)}}\Big(
            4\Big\| \sum_{j=1}^{i-1} \Big(\nabla f_{\pi_j^{(k)}}(\rvx_j^{(k)}) - \nabla f_{\pi_j^{(k)}}(\rvz) \Big) \Big\|^2
            \\
        \nonumber
            &\quad + 4 \Big\|\sum_{j=1}^{i-1} \Big( \nabla f_{\pi_j^{(k)}}(\rvz) - \nabla f_{\pi_j^{(k)}}(\rvx^*) \Big) \Big\|^2
            + 4\Big\| \sum_{j=1}^{i-1} \nabla f_{\pi_j^{(k)}}(\rvx^*) \Big\|^2
            + 4\Big\| \sum_{j=1}^{i-1}\rho_j^{(k)} \Big\|^2
        \Big)
    \end{align}
    We proceed by bounding the first two terms in Eq.~\ref{eq:breg_diff_interm_1} separately. 
    First, 
    \begin{align}
    \nonumber
        &\sum_{i=2}^{n_d^{(k)}} L_{\pi_i^{(k)}}\Big\|\sum_{j=1}^{i-1} \Big( \nabla f_{\pi_j^{(k)}}(\rvx_j^{(k)}) - \nabla f_{\pi_j^{(k)}}(\rvz) \Big)\Big\|^2\\
        \nonumber
        &\leq \sum_{i=2}^{n_d^{(k)}} L_{\pi_i^{(k)}} (i-1)\sum_{j=1}^{i-1} \Big\| \nabla f_{\pi_j^{(k)}}(\rvx_j^{(k)}) - \nabla f_{\pi_j^{(k)}}(\rvz) \Big\|^2\\
    \nonumber
        &= \sum_{j=1}^{n_d^{(k)}-1} \Big( \sum_{i=j+1}^{n_d^{(k)}} L_{\pi_i^{(k)}} (i-1) \Big)\Big\| \nabla f_{\pi_j^{(k)}}(\rvx_j^{(k)}) - \nabla f_{\pi_j^{(k)}}(\rvz)\Big\|^2
        \Big) \\
    \nonumber
        &\leq \sum_{j=1}^{n_d^{(k)}-1} n (\sum_{i=1}^{n_d^{(k)}} L_{\pi_i^{(k)}})
        \Big\| \nabla f_{\pi_j^{(k)}}(\rvx_j^{(k)}) - \nabla f_{\pi_j^{(k)}}(\rvz)\Big\|^2 \\
        & \label{eq:breg_diff_term_1}
        \leq n (\sum_{i=1}^{n_d^{(k)}} L_{\pi_i^{(k)}}) \sum_{j=1}^{n_d^{(k)}}
        \Big\| \nabla f_{\pi_j^{(k)}}(\rvx_j^{(k)}) - \nabla f_{\pi_j^{(k)}}(\rvz)\Big\|^2
    \end{align}

    Next,
    \begin{align}
    \nonumber
        &\sum_{i=2}^{n_d^{(k)}} L_{\pi_i^{(k)}}
        \left\| \sum_{j=1}^{i-1} \Big( \nabla f_{\pi_{j}^{(k)}}(\rvz) - \nabla f_{\pi_j^{(k)}}(\rvx^*) \Big) \right\|^2\\
        \nonumber
        &\stackrel{\text{(a)}}{\leq} \sum_{i=2}^{n_d^{(k)}} L_{\pi_i^{(k)}} 2 \Big(\sum_{j=1}^{i-1} L_{\pi_j^{(k)}} \Big) \Big( \sum_{l=1}^{i-1} B_{f_{\pi_l^{(k)}}}(\rvz, \rvx^*)
        \Big)
        \leq 2 n L \sum_{i=1}^{n_d^{(k)}} L_{\pi_i^{(k)}} \Big(\sum_{l=1}^{i-1} B_{f_{\pi_l^{(k)}}}(\rvz, \rvx^*) \Big)\\
        &\stackrel{\text{(b)}}{\leq} 2 n L \sum_{i=1}^{n_d^{(k)}} L_{\pi_i^{(k)}} \Big(\sum_{l=1}^{n} B_{f_{\pi_l}^{(k)}}(\rvz, \rvx^*) \Big)
    \label{eq:breg_diff_term_2}
        = 2 n^2 L B_F(\rvz, \rvx^*) \cdot \sum_{i=1}^{n_d^{(k)}} L_{\pi_i^{(k)}}
    \end{align}
    where $(a)$ is by Lemma~\ref{lemma:breg_div_ub_lb} and $(b)$ is due to $B_{f_i^{(k)}}(\rvz, \rvx^*) \geq 0, \forall \rvz\in\R^{d}, i\in [n]$.

    Plugging Eq.~\ref{eq:breg_diff_term_1} and Eq.~\ref{eq:breg_diff_term_2} back to Eq.~\ref{eq:breg_diff_interm_1}, there is
    
    \begin{align}
    \label{eq:breg_div_interm_ub}
        I_1 \triangleq & \sum_{i=1}^{n_d^{(k)}} L_{\pi_i^{(k)}} \|\rvx_i^{(k)} - \rvx_1^{(k)}\|^2 \\
        \nonumber
        &\leq 4 \eta^2 n (\sum_{i=1}^{n_d^{(k)}} L_{\pi_i^{(k)}}) \sum_{j=1}^{n_d^{(k)}}
        \Big\| \nabla f_{\pi_j^{(k)}}(\rvx_j^{(k)}) - \nabla f_{\pi_j^{(k)}}(\rvz)\Big\|^2
        \\
        \nonumber
        &\quad + 8 \eta^2 n^2 L B_F(\rvz, \rvx^*) \cdot \sum_{i=1}^{n_d^{(k)}} L_{\pi_i^{(k)}}
        + 4 \eta^2 \sum_{i=2}^{n_d^{(k)}} L_{\pi_i^{(k)}}
        \Big\| \sum_{j=1}^{i-1} \nabla f_{\pi_j^{(k)}} (\rvx^*) \Big\|^2
        + 4 \eta^2 \sum_{i=2}^{n_d^{(k)}} L_{\pi_i^{(k)}}
        \Big\| \sum_{j=1}^{i-1}\rho_j^{(k)}\Big\|^2
    \end{align}

    \textbf{Part II: } Similarly, for $n_d^{(k)} < i \leq n$,
    \begin{align}
    \nonumber
        I_2 \triangleq &\sum_{i=n_d^{(k)}+1}^{n} \widetilde{L}_{i-n_d^{(k)}}^{(k)} \|\rvx_i^{(k)} - \rvx_1^{(k)}\|^2\\
        &= \sum_{i=n_d^{(k)}+1}^{n} \widetilde{L}_{i-n_d^{(k)}}^{(k)} \eta^2 \Big\| \sum_{j=1}^{n_d^{(k)}}(\nabla f_{\pi_{j}^{(k)}}(\rvx_j^{(k)}) + \rho_j^{(k)}) + \sum_{j=n_d^{(k)}+1}^{i-1} (\nabla f_{j-n_d^{(k)}}^{(k, pub)}(\rvx_j^{(k)}) + \rho_j^{(k)})  \Big\|^2
        \tag{From the update of Algorithm~\ref{alg:generalized_shuffled_gradient_fm}}\\
    \nonumber
        &=\eta^2 \sum_{i=n_d^{(k)}+1}^{n} \widetilde{L}_{i-n_d^{(k)}}^{(k)} 
        \Big\| \sum_{j=1}^{n_d^{(k)}} \nabla f_{\pi_j^{(k)}}(\rvx_j^{(k)}) - \sum_{j=1}^{n_d^{(k)}} \nabla f_{\pi_j^{(k)}}(\rvz)
        + \sum_{j=1}^{n_d^{(k)}} \nabla f_{\pi_j^{(k)}} (\rvz)
        - \sum_{j=1}^{n_d^{(k)}} \nabla f_{\pi_j^{(k)}}(\rvx^{*})
        + \sum_{j=1}^{n_d^{(k)}} \nabla f_{\pi_j^{(k)}}(\rvx^{*})\\
        \nonumber
        &\quad + \sum_{j=n_d^{(k)}+1}^{i-1} \nabla f_{j-n_d^{(k)}}^{(k, pub)}(\rvx_j^{(k)})
        - \sum_{j=n_d^{(k)}+1}^{i-1} \nabla f_{j-n_d^{(k)}}^{(k, pub)}(\rvz)
        + \sum_{j=n_d^{(k)}+1}^{i-1} \nabla f_{j-n_d^{(k)}}^{(k, pub)}(\rvz)
        - \sum_{j=n_d^{(k)}+1}^{i-1} \nabla f_{\pi_j^{(k)}}(\rvz)\\
        \nonumber
        &\quad + \sum_{j=n_d^{(k)}+1}^{i-1} \nabla f_{\pi_j^{(k)}}(\rvz)
        - \sum_{j=n_d^{(k)}+1}^{i-1}\nabla f_{\pi_j^{(k)}}(\rvx^{*})
        + \sum_{j=n_d^{(k)}+1}^{i-1}\nabla f_{\pi_j^{(k)}}(\rvx^{*})
        + \sum_{j=1}^{i-1} \rho_j^{(k)}
        \Big\|^2
        \\
    \nonumber
        &= \eta^2 \sum_{i=n_d^{(k)}+1}^{n} \widetilde{L}_{i-n_d^{(k)}}^{(k)}
        \Big\|\Big( \sum_{j=1}^{n_d^{(k)}} \nabla f_{\pi_j^{(k)}}(\rvx_j^{(k)}) + \sum_{j=n_d^{(k)}+1}^{i-1} \nabla f_{j-n_d^{(k)}}^{(k, pub)}(\rvx_j^{(k)})\\
        \nonumber
        &\quad - \sum_{j=1}^{n_d^{(k)}} \nabla f_{\pi_j^{(k)}}(\rvz) - \sum_{j=n_d^{(k)}+1}^{n} \nabla f_{j-n_d^{(k)}}^{(k, pub)}(\rvz)
        \Big)
        + \Big( \sum_{j=n_d^{(k)}+1}^{i-1} \nabla f_{j-n_d^{(k)}}^{(k, pub)}(\rvz)
        - \sum_{j=n_d^{(k)}+1}^{i-1} \nabla f_{\pi_j^{(k)}}(\rvz)
        \Big) \\
        \nonumber
        &\quad + \Big( \sum_{j=1}^{i-1} \nabla f_{\pi_j^{(k)}}(\rvz) - \sum_{j=1}^{i-1} \nabla f_{\pi_j^{(k)}}(\rvx^{*})
        \Big)
        + \sum_{j=1}^{i-1} \nabla f_{\pi_j^{(k)}}(\rvx^{*}) + \sum_{j=1}^{i-1} \rho_j^{(k)}
        \Big\|^2\\
    \label{eq:breg_diff_interm_1_pub}
        &\leq \eta^2 \sum_{i=n_d^{(k)}+1}^{n}\widetilde{L}_{i-n_d^{(k)}}^{(k)}\Big(
            5 \Big\| \sum_{j=1}^{n_d^{(k)}} \nabla f_{\pi_j^{(k)}}(\rvx_j^{(k)}) + \sum_{j=n_d^{(k)}+1}^{i-1} \nabla f_{j-n_d^{(k)}}^{(k, pub)}(\rvx_j^{(k)}) \\
        \nonumber
            &\quad - \sum_{j=1}^{n_d^{(k)}} \nabla f_{\pi_j^{(k)}}(\rvz) - \sum_{j=n_d^{(k)}+1}^{i-1} \nabla f_{j-n_d^{(k)}}^{(k, pub)}(\rvz) \Big\|^2
            + 5\Big\|  \sum_{j=n_d^{(k)}+1}^{i-1} \nabla f_{j-n_d^{(k)}}^{(k, pub)}(\rvz)
            - \sum_{j=n_d^{(k)}+1}^{i-1} \nabla f_{\pi_j^{(k)}}(\rvz)
        \Big\|^2\\
        \nonumber
        &\quad + 5\Big\| \sum_{j=1}^{i-1} \nabla f_{\pi_j^{(k)}}(\rvz)
            - \sum_{j=1}^{i-1} \nabla f_{\pi_j^{(k)}}(\rvx^*) \Big\|^2
        + 5\Big\|\sum_{j=1}^{i-1} \nabla f_{\pi_j^{(k)}}(\rvx^*) \Big\|^2
            + 5\Big\|\sum_{j=1}^{i-1}\rho_j^{(k)} \Big\|^2
        \Big)
    \end{align}

    We proceed by bounding the first three terms in Eq.~\ref{eq:breg_diff_interm_1_pub} separately. First, 
    \begin{align}
    \nonumber
        &\sum_{i=n_d^{(k)}+1}^{n} \widetilde{L}_{i-n_d^{(k)}}^{(k)}
        \Big\| \sum_{j=1}^{n_d^{(k)}} \nabla f_{\pi_j^{(k)}}(\rvx_j^{(k)}) + \sum_{j=n_d^{(k)}+1}^{i-1} \nabla f_{j-n_d^{(k)}}^{(k, pub)}(\rvx_j^{(k)}) 
        - \sum_{j=1}^{n_d^{(k)}} \nabla f_{\pi_j^{(k)}}(\rvz) - \sum_{j=n_d^{(k)}+1}^{i-1} \nabla f_{j-n_d^{(k)}}^{(k, pub)}(\rvz) \Big\|^2\\
    \nonumber
        &\leq \sum_{i=n_d^{(k)}+1}^{n}
        \widetilde{L}_{i-n_d^{(k)}}^{(k)}
        (i-1) \Big(\sum_{j=1}^{n_d^{(k)}} \Big\| \nabla f_{\pi_j^{(k)}}(\rvx_j^{(k)}) - \nabla f_{\pi_j^{(k)}}(\rvz)\Big\|^2
        + \sum_{j=n_d^{(k)}+1}^{i-1}\Big\| \nabla f_{j-n_d^{(k)}}^{(k, pub)}(\rvx_j^{(k)}) - \nabla f_{j-n_d^{(k)}}^{(k, pub)}(\rvz)\Big\|^2
        \Big)\\\
    \nonumber
        &\leq \sum_{j=1}^{n_d^{(k)}} \Big(\sum_{i=n_d^{(k)}+1}^{n} \widetilde{L}_{i-n_d^{(k)}}^{(k)}(i-1)\Big)\Big\| \nabla f_{\pi_j^{(k)}}(\rvx_j^{(k)}) - \nabla f_{\pi_j^{(k)}}(\rvz) \Big\|^2\\
        \nonumber
        &\quad + \sum_{j=n_d^{(k)}+1}^{n-1} \Big(\sum_{i=n_d^{(k)}+1}^{n} \widetilde{L}_{i-n_d^{(k)}}^{(k)}(i-1)\Big)\Big\| \nabla f_{j-n_d^{(k)}}^{(k, pub)}(\rvx_j^{(k)})
        - \nabla f_{j-n_d^{(k)}}^{(k, pub)}(\rvz)\Big\|^2\\
    \label{eq:breg_diff_term_1_pub}
        &\leq \sum_{j=1}^{n_d^{(k)}} n \Big(\sum_{i=n_d^{(k)}+1}^{n}\widetilde{L}_{i-n_d^{(k)}}^{(k)}\Big) \Big\| \nabla f_{\pi_j^{(k)}}(\rvx_j^{(k)}) - \nabla f_{\pi_j^{(k)}}(\rvz)\Big\|^2 \\
        \nonumber
        &\quad 
        + \sum_{j=n_d^{(k)}+1}^{n} n \Big(\sum_{i=n_d^{(k)}+1}^{n} \widetilde{L}_{i-n_d^{(k)}}^{(k)}\Big)
        \Big\| \nabla f_{j-n_d^{(k)}}^{(k, pub)}(\rvx_j^{(k)}) - \nabla f_{j-n_d^{(k)}}^{(k,pub)}(\rvz)\Big\|^2 
    \end{align}
    Next, 
    \begin{align}
    \label{eq:breg_diff_term_dissim_pub}
        &\sum_{i=n_d^{(k)}+1}^{n} \widetilde{L}_{i-n_d^{(k)}}^{(k)} \Big\|  \sum_{j=n_d^{(k)}+1}^{i-1} \nabla f_{i-n_d^{(k)}}^{(k, pub)}(\rvz)
        - \sum_{j=n_d^{(k)}+1}^{i-1} \nabla f_{\pi_j^{(k)}}(\rvz)
        \Big\|^2\\
        \nonumber
        &\leq L^{(k)*} \sum_{i=n_d^{(k)} + 1}^{n} \Big\| \sum_{j=n_d^{(k)}+1}^{i-1} \nabla f_{i-n_d^{(k)}}^{(k, pub)}(\rvz)
        - \sum_{j=n_d^{(k)}+1}^{i-1} \nabla f_{\pi_j^{(k)}}(\rvz)
        \Big\|^2
    \end{align}
    Moreover,
    \begin{align}
        \nonumber
        &\sum_{i=n_d^{(k)}+1}^{n}\widetilde{L}_{i-n_d^{(k)}}^{(k)}
        \Big\| \sum_{j=1}^{i-1} \nabla f_{\pi_j^{(k)}}(\rvz)
            - \sum_{j=1}^{i-1} \nabla f_{\pi_j^{(k)}}(\rvx^*) \Big\|^2\\
    \nonumber
        &\stackrel{\text{(a)}}{\leq} \sum_{i= n_d^{(k)}+1}^{n} \widetilde{L}_{i-n_d^{(k)}}^{(k)} \cdot 2\Big( \sum_{j=1}^{i-1} L_{\pi_j^{(k)}} \Big) \Big( \sum_{j=1}^{i-1} B_{\pi_j^{(k)}}(\rvz, \rvx^*)
        \Big)\\
    \nonumber
        &\stackrel{\text{(b)}}{\leq} \sum_{i=n_d^{(k)}+1}^{n} \widetilde{L}_{i-n_d^{(k)}}^{(k)} \cdot 2\Big( \sum_{j=1}^{i-1} L_{\pi_j^{(k)}} \Big)
        \Big( \sum_{j=1}^{n} B_{\pi_j^{(k)}}(\rvz, \rvx^*) \Big)\\
        &\leq 2 \sum_{i=n_d^{(k)}+1}^{n} \widetilde{L}_{i-n_d^{(k)}}^{(k)} \cdot n L \cdot \Big( \sum_{j=1}^{n} B_{j}(\rvz, \rvx^*) \Big)
    \label{eq:breg_diff_term_2_pub}
        \leq 2 n^2 L B_F(\rvz, \rvx^*) \cdot \sum_{i=n_d^{(k)}+1}^{n} \widetilde{L}_{i-n_d^{(k)}}^{(k)}
    \end{align}
    where $(a)$ is by Lemma~\ref{lemma:breg_div_ub_lb} and $(b)$ is due to $B_{f_i}(\rvz, \rvx^*) \geq 0, \forall \rvz\in\R^{d}, i\in [n]$.
    Plugging Eq.~\ref{eq:breg_diff_term_1_pub}, Eq.~\ref{eq:breg_diff_term_dissim_pub} and Eq.~\ref{eq:breg_diff_term_2_pub} back to Eq.~\ref{eq:breg_diff_interm_1_pub}, there is
    \begin{align}
    \label{eq:breg_div_interm_ub_pub}
        &I_2 = \sum_{i=n_d^{(k)}+1}^{n}\widetilde{L}_{i-n_d^{(k)}}^{(k)}\|\rvx_i^{(k)} - \rvx_1^{(k)}\|^2\\
        \nonumber
        &\leq 5 \eta^2 n (\sum_{i=n_d^{(k)}+1}^{n} \widetilde{L}_{i-n_d^{(k)}}^{(k)}) \sum_{j=1}^{n_d^{(k)}} \Big\| \nabla f_{\pi_j^{(k)}}(\rvx_j^{(k)}) - \nabla f_{\pi_j^{(k)}}(\rvz)\Big\|^2\\
        \nonumber
        &\quad + 5\eta^2 n (\sum_{i=n_d^{(k)}+1}^{n} \widetilde{L}_{i-n_d^{(k)}}^{(k)}) \sum_{j=n_d^{(k)}+1}^{n}\Big\| \nabla f_{j-n_d^{(k)}}^{(k, pub)}(\rvx_j^{(k)}) - \nabla f_{j-n_d^{(k)}}^{(k, pub)}(\rvz)\Big\|^2\\
    \nonumber
        &\quad + 5\eta^2 L^{(k)*} \sum_{i=n_d^{(k)} + 1}^{n} \Big\|
        \sum_{j=n_d^{(k)}+1}^{i-1} \nabla f_{i-n_d^{(k)}}^{(k, pub)}(\rvz) -
        \sum_{j=n_d^{(k)}+1}^{i-1} \nabla f_{\pi_j^{(k)}}(\rvz)
        \Big\|^2
        + 10 \eta^2 n^2 L B_F(\rvz, \rvx^*) \sum_{i=n_d^{(k)}+1}^{n} \widetilde{L}_{i-n_d^{(k)}}^{(k)}\\
        \nonumber
        &\quad + 5\eta^2 \sum_{i=n_d^{(k)}+1}^{n} \widetilde{L}_{i-n_d^{(k)}}^{(k)}\Big\| \sum_{j=1}^{i-1} \nabla f_{\pi_j^{(k)}}(\rvx^*)\Big\|^2
        + 5\eta^2 \sum_{i=n_d^{(k)}+1}^{n} \widetilde{L}_{i-n_d^{(k)}}^{(k)} \Big\| \sum_{j=1}^{i-1} \rho_j^{(k)}\Big\|^2 
    \end{align}

    Combining Eq.~\ref{eq:breg_div_interm_ub} and Eq.~\ref{eq:breg_div_interm_ub_pub}, there is
    \begin{align}
    \label{eq:breg_div_interm_ub_sum}
        &\frac{1}{n} (I_1 + I_2) = \frac{1}{n}\Big( \sum_{i=1}^{n_d^{(k)}} L_{\pi_i^{(k)}} \|\rvx_i^{(k)} - \rvx_1^{(k)} \|^2 
        + \sum_{i=n_d^{(k)}+1}^{n} \widetilde{L}_{i-n_d^{(k)}}^{(k)} \|\rvx_i^{(k)} - \rvx_1^{(k)}\|^2\Big) \\
        \nonumber
        &\leq 5\eta^2 n \widehat{L}^{(k)} \sum_{j=1}^{n_d^{(k)}}\Big\| \nabla f_{\pi_j^{(k)}}(\rvx_{j}^{(k)}) - \nabla f_{\pi_j^{(k)}}(\rvz)\Big\|^2
        + 5\eta^2 n \widehat{L}^{(k)} \sum_{j=n_d^{(k)}+1}^{n}\Big\| \nabla f_{j-n_d^{(k)}}^{(k, pub)}(
        \rvx_j^{(k)}) - \nabla f_{j-n_d^{(k)}}^{(k,pub)}(\rvz)\Big\|^2 \\
        \nonumber
        &\quad + 10 \eta^2 n^2 \widehat{L}^{(k)} L B_F(\rvz, \rvx^*)
        + 5 \eta^2 \frac{1}{n} \Big(\sum_{i=2}^{n_d^{(k)}} L_{\pi_i^{(k)}} \Big\| \sum_{j=1}^{i-1} \nabla f_{\pi_j^{(k)}}(\rvx^*)\Big\|^2
        + \sum_{i=n_d^{(k)}+1}^{n}\widetilde{L}_{i-n_d^{(k)}}^{(k)}\Big\| \sum_{j=1}^{i-1}\nabla f_{\pi_j^{(k)}}(\rvx^*)\Big\|^2 \Big)\\
        \nonumber
        &\quad 
        + 5\eta^2 \frac{1}{n} \Big( \sum_{i=2}^{n_d^{(k)}} L_{\pi_i^{(k)}} \Big\| \sum_{j=1}^{i-1} \rho_j^{(k)}\Big\|^2
        + \sum_{i=n_d^{(k)}+1}^{n} \widetilde{L}_{i-n_d^{(k)}}^{(k)}\Big\| \sum_{j=1}^{i-1} \rho_j^{(k)}\Big\|^2\Big)\\
        \nonumber
        &\quad + 5\eta^2 L^{(k)*} \frac{1}{n}\sum_{i=n_d^{(k)} + 1}^{n} \Big\|
        \sum_{j=n_d^{(k)}+1}^{i-1} \nabla f_{i-n_d^{(k)}}^{(k, pub)}(\rvz) -
        \sum_{j=n_d^{(k)}+1}^{i-1} \nabla f_{\pi_j^{(k)}}(\rvz)
        \Big\|^2
    \end{align}

    Hence, plugging Eq.~\ref{eq:breg_div_interm_ub_sum} back to Eq.~\ref{eq:breg_diff_interm_2}, there is
    \begin{align*}
        &\frac{1}{n} \Big(\sum_{i=1}^{n_d^{(k)}}
            \Big( B_{f_{\pi_i^{(k)}}}(\rvx_{1}^{(k+1)}, \rvx_i^{(k)})
            - B_{f_{\pi_i^{(k)}}}(\rvz, \rvx_i^{(k)}) \Big)
            + \sum_{i=n_d^{(k)}+1}^{n}\Big(
                B_{f_{i-n_d^{(k)}}^{(k,pub)}}(\rvx_1^{(k+1)}, \rvx_i^{(k)})
                - B_{f_{i-n_d^{(k)}}^{(k, pub)}}(\rvz, \rvx_i^{(k)})
            \Big)
        \Big) \\
        &= \widehat{L}^{(k)} \|\rvx_{1}^{(k+1)} - \rvx_1^{(k)}\|^2 
        \\
        \nonumber
        &\quad + 5\eta^2 n \widehat{L}^{(k)} \sum_{j=1}^{n_d^{(k)}}\Big\| \nabla f_{\pi_j^{(k)}}(\rvx_{j}^{(k)}) - \nabla f_{\pi_j^{(k)}}(\rvz)\Big\|^2
        + 5\eta^2 n \widehat{L}^{(k)} \sum_{j=n_d^{(k)}+1}^{n}\Big\| \nabla f_{j-n_d^{(k)}}^{(k, pub)}(
        \rvx_j^{(k)}) - \nabla f_{j-n_d^{(k)}}^{(k,pub)}(\rvz)\Big\|^2 \\
        \nonumber
        &\quad + 10 \eta^2 n^2 \widehat{L}^{(k)} L B_F(\rvz, \rvx^*)
        + 5 \eta^2 \frac{1}{n} \Big(\sum_{i=2}^{n_d^{(k)}} L_{\pi_i^{(k)}} \Big\| \sum_{j=1}^{i-1} \nabla f_{\pi_j^{(k)}}(\rvx^*)\Big\|^2
        + \sum_{i=n_d^{(k)}+1}^{n}\widetilde{L}_{i-n_d^{(k)}}^{(k)}\Big\| \sum_{j=1}^{i-1}\nabla f_{\pi_j^{(k)}}(\rvx^*)\Big\|^2 \Big)\\
        \nonumber
        &\quad 
        + 5\eta^2 \frac{1}{n} \Big( \sum_{i=2}^{n_d^{(k)}} L_{\pi_i^{(k)}} \Big\| \sum_{j=1}^{i-1} \rho_j^{(k)}\Big\|^2
        + \sum_{i=n_d^{(k)}+1}^{n} \widetilde{L}_{i-n_d^{(k)}}^{(k)}\Big\| \sum_{j=1}^{i-1} \rho_j^{(k)}\Big\|^2\Big)\\
        \nonumber
        &\quad + 5\eta^2 L^{(k)*} \frac{1}{n} \sum_{i=n_d^{(k)} + 1}^{n} \Big\|
        \sum_{j=n_d^{(k)}+1}^{i-1} \nabla f_{i-n_d^{(k)}}^{(k, pub)}(\rvz) -
        \sum_{j=n_d^{(k)}+1}^{i-1} \nabla f_{\pi_j^{(k)}}(\rvz)
        \Big\|^2\\
        \nonumber
        &\quad - \frac{1}{n} \Big( 
        \sum_{i=1}^{n_d^{(k)}} \frac{ \left\|\nabla f_{\pi_i^{(k)}}(\rvx_i^{(k)}) - \nabla f_{\pi_i^{(k)}}(\rvz) \right\|^2}{2 L_{\pi_{i}^{(k)}}}
        + \sum_{i=n_d^{(k)}+1}^{n} \frac{\left\| \nabla f_{i-n_d^{(k)}}^{(k, pub)}(\rvx_i^{(k)}) - \nabla f_{i-n_d^{(k)}}^{(k, pub)}(\rvz) \right\|^2}{2\widetilde{L}_{i-n_d^{(k)}}^{(k)}}
        \Big)
    \end{align*}

    If one sets the learning rate $\eta$ such that
    \begin{align*}
        &5 \eta^2 n \widehat{L}^{(k)} \leq \frac{1}{n} \cdot \frac{1}{2 \widehat{L}^{(k)*}}, \quad
        \Rightarrow \eta \leq \frac{1}{n\sqrt{10 \widehat{L}^{(k)} \widehat{L}^{(k)*}}}
    \end{align*}
    then there is
    \begin{align*}
        &\frac{1}{n} \Big(\sum_{i=1}^{n_d^{(k)}}
            \Big( B_{f_{\pi_i^{(k)}}}(\rvx_{1}^{(k+1)}, \rvx_i^{(k)})
            - B_{f_{\pi_i^{(k)}}}(\rvz, \rvx_i^{(k)}) \Big)
            + \sum_{i=n_d^{(k)}+1}^{n}\Big(
                B_{f_{i-n_d^{(k)}}^{(k,pub)}}(\rvx_1^{(k+1)}, \rvx_i^{(k)})
                - B_{f_{i-n_d^{(k)}}^{(k, pub)}}(\rvz, \rvx_i^{(k)})
            \Big)
        \Big) \\
        \nonumber
        &\leq \widehat{L}^{(k)} \|\rvx_1^{(k+1)} - \rvx_1^{(k)} \|^2
        + 10\eta^2 n^2 \widehat{L}^{(k)} L B_F(\rvz, \rvx^{*})\\
        \nonumber
        &\quad + 5 \eta^2 \frac{1}{n} \Big(\sum_{i=2}^{n_d^{(k)}} L_{\pi_i^{(k)}} \Big\| \sum_{j=1}^{i-1} \nabla f_{\pi_j^{(k)}}(\rvx^*)\Big\|^2
        + \sum_{i=n_d^{(k)}+1}^{n}\widetilde{L}_{i-n_d^{(k)}}^{(k)}\Big\| \sum_{j=1}^{i-1}\nabla f_{\pi_j^{(k)}}(\rvx^*)\Big\|^2 \Big)\\
        \nonumber
        &\quad + 5\eta^2 \frac{1}{n} \Big( \sum_{i=2}^{n_d^{(k)}} L_{\pi_i^{(k)}} \Big\| \sum_{j=1}^{i-1} \rho_j^{(k)}\Big\|^2
        + \sum_{i=n_d^{(k)}+1}^{n} \widetilde{L}_{i-n_d^{(k)}}^{(k)}\Big\| \sum_{j=1}^{i-1} \rho_j^{(k)}\Big\|^2\Big)\\
        &\quad + 5\eta^2 L^{(k)*} \frac{1}{n} \sum_{i=n_d^{(k)} + 1}^{n} \Big\|
        \sum_{j=n_d^{(k)}+1}^{i-1} \nabla f_{i-n_d^{(k)}}^{(k, pub)}(\rvz) -
        \sum_{j=n_d^{(k)}+1}^{i-1} \nabla f_{\pi_j^{(k)}}(\rvz)
        \Big\|^2
    \end{align*}

\end{proof}

\begin{lemma}[One Epoch Convergence]
\label{lemma:one_epoch_convergence}
    Under Assumptions~\ref{ass:convexity},~\ref{ass:reg},~\ref{ass:dissim_partial_lipschitzness},~\ref{ass:appendix_refined_smoothness} and~\ref{ass:H_smoothness}, for any epoch $k \in [K]$, $\beta > 0$, and $\forall \rvz \in \R^d$, 
    if $\eta \leq \frac{1}{n\sqrt{10 \widehat{L}^{(k)} \widehat{L}^{(k)*}}}$, 
    Algorithm~\ref{alg:generalized_shuffled_gradient_fm} guarantees
    \begin{align}
    \label{eq:one_epoch_convergence}
        &G(\rvx_1^{(k+1)}) - G(\rvz)\\
    \nonumber
        &\leq \frac{1}{2 n \eta} ( \|\rvz - \rvx_1^{(k)}\|^2  - \|\rvz - \rvx_1^{(k+1)}\|^2 )
        + ( \frac{L_H^{(k)} + \beta}{2} - \frac{\mu_{\psi}}{2} )\| \rvz - \rvx_1^{(k+1)}\|^2
         + 10\eta^2 n^2 \widehat{L}^{(k)} L B_F(\rvz, \rvx^*)
       \\
        \nonumber
        &\quad + \underbrace{ 5 \eta^2 \frac{1}{n} \Big(\sum_{i=2}^{n_d^{(k)}} L_{\pi_i^{(k)}} \Big\| \sum_{j=1}^{i-1} \nabla f_{\pi_j^{(k)}}(\rvx^*)\Big\|^2
        + \sum_{i=n_d^{(k)}+1}^{n}\widetilde{L}_{i-n_d^{(k)}}^{(k)}\Big\| \sum_{j=1}^{i-1}\nabla f_{\pi_j^{(k)}}(\rvx^*)\Big\|^2 \Big)
        }_{\text{Optimization Uncertainty}}\\
        \nonumber
        &\quad 
        + \underbrace{ \frac{1}{n}\sum_{i=1}^{n}\langle -\rho_i^{(k)}, \rvx_1^{(k+1)} - \rvz\rangle
        + 5\eta^2 \frac{1}{n} \Big( \sum_{i=2}^{n_d^{(k)}} L_{\pi_i^{(k)}} \Big\| \sum_{j=1}^{i-1} \rho_j^{(k)}\Big\|^2
        + \sum_{i=n_d^{(k)}+1}^{n} \widetilde{L}_{i-n_d^{(k)}}^{(k)}\Big\| \sum_{j=1}^{i-1} \rho_j^{(k)}\Big\|^2\Big)
        }_{\text{Injected Noise}}\\
    \nonumber
        &\quad 
         + \underbrace{ \frac{1}{2n^2\beta}(C_n^{(k)})^2
         }_{\text{Non-vanishing Dissimilarity}}
        + \underbrace{ 5\eta^2 L^{(k)*} \frac{1}{n} \sum_{i=n_d^{(k)} + 1}^{n} \Big\|
        \sum_{j=n_d^{(k)}+1}^{i-1} \nabla f_{i-n_d^{(k)}}^{(k, pub)}(\rvz) -
        \sum_{j=n_d^{(k)}+1}^{i-1} \nabla f_{\pi_j^{(k)}}(\rvz)
        \Big\|^2
        }_{\text{Vanishing Dissimilarity}}
    \end{align}

\end{lemma}

\begin{proof}[Proof of Lemma~\ref{lemma:one_epoch_convergence}]
    By Lemma~\ref{lemma:one_epoch_bg_3pid} and Lemma~\ref{lemma:one_epoch_bound_bregmandiv}, for any $k \in [K]$ and $\forall \rvz\in \R^d$, if $\eta \leq \frac{1}{n\sqrt{10 \widehat{L}^{(k)} \widehat{L}^{(k)*}}}$,
    \begin{align}
    \label{eq:one_epoch_convergence_interm}
        &G(\rvx_1^{(k+1)}) - G(\rvz)\\
        \nonumber
        &\leq H^{(k)}(\rvx_1^{(k+1)}) - H^{(k)}(\rvz)
        + \frac{\|\rvz - \rvx_1^{(k)}\|^2}{2n \eta} - (\frac{1}{2n \eta} + \frac{\mu_\psi}{2}) \|\rvz - \rvx_1^{(k+1)}\|^2
        - \frac{1}{2n \eta}\|\rvx_1^{(k+1)} - \rvx_1^{(k)}\|^2\\
        \nonumber
        &\quad + \frac{1}{n}\sum_{i=1}^{n}\langle -\rho_i^{(k)}, \rvx_1^{(k+1)} - \rvz\rangle
        + \widehat{L}^{(k)} \|\rvx_{1}^{(k+1)} - \rvx_1^{(k)}\|^2 
        + 10\eta^2 n^2 \widehat{L}^{(k)} L B_F(\rvz, \rvx^*) \\
        \nonumber
        &\quad + 5 \eta^2 \frac{1}{n} \Big(\sum_{i=2}^{n_d^{(k)}} L_{\pi_i^{(k)}} \Big\| \sum_{j=1}^{i-1} \nabla f_{\pi_j^{(k)}}(\rvx^*)\Big\|^2
        + \sum_{i=n_d^{(k)}+1}^{n}\widetilde{L}_{i-n_d^{(k)}}^{(k)}\Big\| \sum_{j=1}^{i-1}\nabla f_{\pi_j^{(k)}}(\rvx^*)\Big\|^2 \Big)\\
        \nonumber
        &\quad + 5\eta^2 \frac{1}{n} \Big( \sum_{i=2}^{n_d^{(k)}} L_{\pi_i^{(k)}} \Big\| \sum_{j=1}^{i-1} \rho_j^{(k)}\Big\|^2
        + \sum_{i=n_d^{(k)}+1}^{n} \widetilde{L}_{i-n_d^{(k)}}^{(k)}\Big\| \sum_{j=1}^{i-1} \rho_j^{(k)}\Big\|^2\Big)\\
    \nonumber
        &\quad + 5\eta^2 L^{(k)*} \frac{1}{n} \sum_{i=n_d^{(k)} + 1}^{n} \Big\|
        \sum_{j=n_d^{(k)}+1}^{i-1} \nabla f_{i-n_d^{(k)}}^{(k, pub)}(\rvz) -
        \sum_{j=n_d^{(k)}+1}^{i-1} \nabla f_{\pi_j^{(k)}}(\rvz)
        \Big\|^2
    \end{align}

    Since $\eta \leq \frac{1}{n \sqrt{10 \widehat{L}^{(k)} \widehat{L}^{(k)*}}}$, there is $\widehat{L}^{(k)} \leq \sqrt{\widehat{L}^{(k)} \widehat{L}^{(k)*}} \leq \frac{1}{\sqrt{10} n \eta} \leq \frac{1}{2 n \eta}$, 
    and so $(\widehat{L}^{(k)} - \frac{1}{2n \eta})\|\rvx_1^{(k+1)} - \rvx_1^{(k)}\|^2 \leq 0$.

    For any $\beta > 0$ and any $\rvz\in\R^{d}$, 
    \begin{align}
    \nonumber
        &H^{(k)}(\rvx_{1}^{(k+1)}) - H^{(k)}(\rvz)\\
    \nonumber
        &= H^{(k)}(\rvx_{1}^{(k+1)}) -  H^{(k)}(\rvz)
        - \langle \nabla H^{(k)}(\rvz), \rvx_{1}^{(k+1)} - \rvz\rangle + \langle \nabla H^{(k)}(\rvz), \rvx_{1}^{(k+1)} - \rvz\rangle\\
    \nonumber
        &\stackrel{\text{(a)}}{\leq} \frac{L_H^{(k)}}{2}  \|\rvx_{1}^{(k+1)} - \rvz\|^2 + \frac{1}{2n^2 \beta}(C_n^{(k)})^2 + \frac{\beta}{2} \|\rvx_{1}^{(k+1)} - \rvz\|^2 \\
    \label{eq:dissim_ub}
        &= \frac{L_H^{(k)} + \beta}{2} \|\rvx_{1}^{(k+1)} - \rvz\|^2 + \frac{1}{2n^2 \beta}(C_n^{(k)})^2
    \end{align}
    where (a) is by Assumption~\ref{ass:H_smoothness},~\ref{ass:dissim_partial_lipschitzness}, Lemma~\ref{lemma:breg_div_ub_lb} and Young's inequality. 

    We comment 
    that if $H^{(k)} = 0$ for epoch $s$, a tighter bound $H^{(k)}(\rvx_1^{(k+1)}) - H^{(k)}(\rvz) = 0$ holds, as Young's inequality is not tight in this case. Consequently, one can set $\beta = 0$. 

    Finally, plugging Eq.\ref{eq:dissim_ub} back to Eq.\ref{eq:one_epoch_convergence_interm} yields the inequality (\ref{eq:one_epoch_convergence}) stated in the lemma.


\end{proof}

\subsection{Expected One Epoch Convergence}
\label{subsec:appendix_expected_one_epoch_convergence}

There are two sources of randomness involved in each epoch: 1) the shuffling operator in optimization, and 2) injected Gaussian noise to perturb the gradient for privacy preservation. 
1) can be bounded using Lemma~\ref{lemma:opt_noise_bound}. To bound 2),
in this section, we show upper bounds on the expectation of the additional error term due to noise injection and the noise variance in Lemma~\ref{lemma:noise_bias} and Lemma~\ref{lemma:noise_variance}. We then give an expected one epoch convergence bound, where the expectation is taken over the two sources of randomness, in Lemma~\ref{lemma:expected_one_epoch_convergence}.


\begin{lemma}[Additional Error]
\label{lemma:noise_bias}
    For any epoch $s\in [K]$ and $\forall \rvz \in \R^{d}$, consider the injected noise $\rho_{i}^{(k)} \sim \gN(0, (\sigma^{(k)})^2 \sI_d)$, $\forall i\in [n]$, if the regularization function $\psi$ is twice differentiable and $\rvz$ is independent of $\rho_i^{(k)}, \forall i\in [n]$, then the error caused by noise injection in epoch $k$ is
    \begin{align}
        \E\left[\frac{1}{n}\sum_{i=1}^{n} \langle \rho_i^{(k)}, \rvx_1^{(k+1)} - \rvz \rangle \right] 
        \leq (\sigma^{(k)})^2 nd \eta^2 \widehat{L}^{(k)*}
    \end{align}
    where the expectation is taken w.r.t. the injected noise $\{\rho_i^{(k)}\}_{i=1}^{n}$.
\end{lemma}

\begin{proof}[Proof of Lemma~\ref{lemma:noise_bias}]
    First, note that if $\rvz$ is independent of $\rho_i^{(k)}$, $\forall i\in [n]$, 
    there is 
    
    $\E\left[\langle \frac{1}{n}\sum_{i=1}^{n} \rho_i^{(k)}, \rvx_1^{(k+1)} - \rvz \rangle \right] 
    = \frac{1}{n}\sum_{i=1}^{n}\E\left[ \langle \rho_i^{(k)}, \rvx_1^{(k+1)} - \rvz \rangle \right] = \frac{1}{n}\sum_{i=1}^{n}\E\left[ \langle \rho_i^{(k)}, \rvx_1^{(k+1)}  \rangle \right]$.

    Recall that the update rule in Algorithm~\ref{alg:generalized_shuffled_gradient_fm} in epoch $k\in [K]$ is
    \begin{align}
    \nonumber
        &\rvx_{i+1}^{(k)} = \rvx_{i}^{(k)} - \eta \Big(\nabla f_{\pi_i^{(k)}} + \rho_i^{(k)}\Big),\quad \forall i \in[n_d^{(k)}]\\
    \nonumber
        &\rvx_{i+1}^{(k)} = \rvx_i^{(k)} - \eta \Big(\nabla f_{i-n_d^{(k)}}^{(k, pub)} + \rho_i^{(k)}\Big), \quad \forall n_d^{(k)} < i \leq n\\
    \label{eq:prox_step}
        &\rvx_{1}^{(k+1)} = \argmin_{\rvx\in\R^d} n \psi(\rvx) + \frac{\| \rvx - \rvx_{n+1}^{(k)}\|^2}{2\eta}
    \end{align}

    Since $\psi$ is twice differentiable, by Stein's Lemma (Lemma~\ref{lemma:steins_lemma}), for any $i\in [n]$, conditional on $\rho_j^{(k)}, \forall j\neq i$,
    \begin{align}
    \label{eq:application_steins_lemma}
        \E_{\rho_i^{(k)}}\left[ \langle \rho_i^{(k)}, \rvx_{1}^{(k+1)} \rangle \mid \{\rho_j^{(k)}\}_{j\neq i}\right]
        = (\sigma^{(k)})^2\cdot \E_{\rho_i^{(k)}}\left[\text{tr}(\frac{\partial \rvx_1^{(k+1)}}{\partial \rho_i^{(k)}}) \mid \{\rho_j^{(k)}\}_{j\neq i} \right]
    \end{align}
    We proceed by computing $\frac{\partial \rvx_1^{(k+1)}}{\partial \rho_i^{(k)}}$.
    By the optimality condition of $\rvx_1^{(k+1)}$ as in Eq.~\ref{eq:prox_step},

    \begin{align*}
        &n \nabla \psi(\rvx_1^{(k+1)}) + \frac{1}{\eta}\cdot (\rvx_{1}^{(k+1)} - \rvx_{n+1}^{(k)})  = \mathbf{0}\\
        &n\eta \nabla \psi(\rvx_1^{(k+1)}) + \rvx_1^{(k+1)} - \rvx_{n+1}^{(k)} = \mathbf{0}
    \end{align*}

    And using implicit differentiation of the above optimality condition, 
    \begin{align}
    \nonumber
        &n \eta \frac{\partial \nabla \psi(\rvx_1^{(k+1)})}{\partial \rho_i^{(k)}}
        + \frac{\partial \rvx_1^{(k+1)}}{\partial \rho_i^{(k)}}
        - \frac{\partial \rvx_{n+1}^{(k)}}{\partial \rho_i^{(k)}} = \mathbf{0}\\
    \nonumber
        &n \eta \nabla^2\psi(\rvx_1^{(k+1)})\frac{\partial \rvx_1^{(k+1)}}{\partial \rho_i^{(k)}} + \frac{\partial \rvx_1^{(k+1)}}{\partial \rho_i^{(k)}} - \frac{\partial \rvx_{n+1}^{(k)}}{\partial \rho_i^{(k)}} = \mathbf{0}\\
    \nonumber
        &\Big( n \eta \nabla^2 \psi(\rvx_1^{(k+1)}) + \sI_d \Big) \frac{\partial \rvx_1^{(k+1)}}{\partial \rho_i^{(k)}}
        = \frac{\partial \rvx_{n+1}^{(k)}}{\partial \rho_i^{(k)}}\\
    \label{eq:diff_implicit}
        &\frac{\partial \rvx_{1}^{(k+1)}}{\partial \rho_i^{(k)}} = \Big(\eta n \nabla^2 \psi(\rvx_1^{(k+1)}) + \sI_d \Big)^{-1} \frac{\partial \rvx_{n+1}^{(k)}}{\partial \rho_i^{(k)}}
    \end{align}
    where $\nabla^2 \psi(\rvx_1^{(k+1)})$ is the Hessian of $\psi$ evaluated at $\rvx_1^{(k+1)}$.

    We proceed by computing $\frac{\partial \rvx_{n+1}^{(k)}}{\partial \rho_i^{(k)}}$.
    Note that $\rho_i^{(k)}$ directly affects the update of $\rvx_{i+1}^{(k)}$ and indirectly affects the subsequent updates of $\rvx_{j}^{(k)}$ for all  $j > i + 1$. Hence, we decompose $\rvx_{n+1}^{(k)}$ as follows: for $i \leq n_d^{(k)}$,
    \begin{align*}
        \rvx_{n+1}^{(k)} &= \underbrace{
            \rvx_1 - \eta \sum_{j=1}^{i-1}\Big( \nabla f_{\pi_j^{(k)}}(\rvx_j^{(k)}) + \rho_j^{(k)} \Big)
            - \nabla f_{\pi_i^{(k)}}(\rvx_i^{(k)})
        }_{\text{Independent of $\rho_i^{(k)}$}}\\
        \nonumber
        &\quad - \underbrace{
            \rho_i^{(k)}
        }_{\text{Direct dependency}}
        - \underbrace{
            \eta \sum_{j=i+1}^{n_d^{(k)}}\Big( \nabla f_{\pi_j^{(k)}}(\rvx_j^{(k)}) + \rho_j^{(k)}\Big) - \eta \sum_{j=n_d^{(k)}+1}^{n}\Big(\nabla f_{j-n_d^{(k)}}^{(k, pub)}(\rvx_j^{(k)}) + \rho_j^{(k)}
        \Big)
        }_{\text{Implicit dependency on $\rho_i^{(k)}$ through $\rvx_j^{(k)}$'s}}
    \end{align*}
    and for $n_d^{(k)} < i \leq n$,
    \begin{align*}
        \rvx_{n+1}^{(k)} 
        &= \underbrace{
            \rvx_1 - \eta \sum_{j=1}^{n_d^{(k)}} \Big(\nabla f_{\pi_j^{(k)}}(\rvx_j^{(k)}) + \rho_j^{(k)}\Big) - \eta \sum_{j=n_d^{(k)}+1}^{i-1}\Big(\nabla f_{j-n_d^{(k)}}^{(k, pub)}(\rvx_j^{(k)}) + \rho_j^{(k)}\Big)
            - \nabla f_i^{(k, pub)}(\rvx_i^{(k)})
        }_{\text{Independent of $\rho_i^{(k)}$ through $\rvx_j^{(k)}$'s}}\\
        \nonumber
        &\quad - \underbrace{\rho_i^{(k)}}_{\text{Direct dependency}}
        - \underbrace{
            \eta \sum_{j=i+1}^{n}\Big(\nabla f_{j-n_d^{(k)}}^{(k, pub)}(\rvx_j^{(k)}) + \rho_j^{(k)}\Big)
        }_{\text{Implicit dependency on $\rho_i^{(k)}$}}
    \end{align*}
    And so for $i\leq n_d^{(k)}$,
    \begin{align}
        \frac{\partial \rvx_{n+1}^{(k)}}{\partial \rho_i^{(k)}} 
    \nonumber
        &= -\eta \sI_d - \eta \sum_{j=i+1}^{n_d^{(k)}} \frac{\partial \nabla f_{\pi_j^{(k)}}(\rvx_j^{(k)})}{\partial \rho_i^{(k)}}
        - \eta \sum_{j=n_d^{(k)}+1}^{n} \frac{\partial \nabla f_{j-n_d^{(k)}}^{(k, pub)}(\rvx_j^{(k)})}{\partial \rho_i^{(k)}}\\
    \label{eq:partial_x_n1_small_i}
        &= -\eta \sI_d - \eta \sum_{j=i+1}^{n_d^{(k)}} \nabla^2 f_{\pi_j^{(k)}}(\rvx_j^{(k)})\frac{\partial \rvx_j^{(k)}}{\partial \rho_i^{(k)}}
        - \eta \sum_{j=n_d^{(k)}+1}^{n} \nabla^2 f_{j-n_d^{(k)}}^{(k, pub)}(\rvx_j^{(k)}) \frac{\partial \rvx_j^{(k)}}{\partial \rho_i^{(k)}}
    \end{align}
    and for $n_d^{(k)} < i \leq n$,
    \begin{align}
    \label{eq:partial_x_n1_large_i}
    \frac{\rvx_{n+1}^{(k)}}{\partial \rho_i^{(k)}}
        = -\eta \sI_d - \eta \sum_{j=i+1}^{n}\frac{\partial \nabla f_{j-n_d^{(k)}}^{(k, pub)}(\rvx_j^{(k)})}{\partial \rho_i^{(k)}}
        = -\eta \sI_d - \eta \sum_{j=i+1}^{n} \nabla^2 f_{j-n_d^{(k)}}^{(k, pub)}(\rvx_j^{(k)}) \frac{\partial \rvx_j^{(k)}}{\partial \rho_i^{(k)}}
    \end{align}

    We now compute $\frac{\partial \rvx_j^{(k)}}{\partial \rho_i^{(k)}}$ for $j > i$. First, note that by the update rule, 
    \begin{align}
    \label{eq:partial_x_i1}
        \frac{\partial \rvx_{i+1}^{(k)}}{\partial \rho_i^{(k)}}
    = -\eta \sI_d
    \end{align}
    
    and for any $i < j \leq n$,
    \begin{align*}
        &\frac{\partial \rvx_{j+1}^{(k)}}{\partial \rho_i^{(k)}}
        = \begin{cases}
            \frac{\partial}{\partial \rho_i^{(k)}}\Big(\rvx_{j}^{(k)} - \eta \Big(\nabla f_{\pi_{j}^{(k)}}(\rvx_{j}^{(k)}) + \rho_{j}^{(k)} \Big)
        \Big) & \text{if $j\leq n_d^{(k)}$}\\
            \frac{\partial}{\partial \rho_i^{(k)}}\Big(\rvx_{j-n_d^{(k)}}^{(k, pub)}(\rvx_{j}^{(k)}) + \rho_{j}^{(k)} \Big) & \text{Otherwise} 
        \end{cases}\\
        &\Rightarrow \frac{\partial \rvx_{j+1}^{(k)}}{\partial \rho_i^{(k)}}
        = \begin{cases}
            \Big(\sI_d - \eta \nabla^2 f_{\pi_{j}^{(k)}}(\rvx_{j}^{(k)})\Big) \frac{\partial \rvx_{j}^{(k)}}{\partial \rho_i^{(k)}} & \text{if $j \leq n_d^{(k)}$}\\
            \Big( \sI_d - \eta \nabla^2 f_{j-n_d^{(k)}}^{(k,pub)}(\rvx_{j}^{(k)})\Big)\frac{\partial \rvx_{j}^{(k)}}{\partial \rho_i^{(k)}} & \text{Otherwise}
        \end{cases}
    \end{align*}
    Therefore, by the above recursion, for any $i < j \leq n_d^{(k)}$,
    \begin{align}
    \label{eq:partial_x_j_small_i}
        \frac{\partial \rvx_{j+1}^{(k)}}{\partial \rho_i^{(k)}}
        = -\eta \cdot \prod_{l=i+1}^{j}\Big( \sI_d - \eta \nabla^2 f_{\pi_{l}^{(k)}}(\rvx_{l}^{(k)})\Big) 
    \end{align}
    and similarity, for any $n_d^{(k)} < j \leq n$,
    \begin{align}
    \label{eq:partial_x_j_large_i}
        \frac{\partial \rvx_{j+1}^{(k)}}{\partial \rho_i^{(k)}}
        = -\eta \cdot \Big(\prod_{l=i+1}^{n_d^{(k)}}\Big(\sI_d - \eta \nabla^2 f_{\pi_{l}^{(k)}}(\rvx_{l}^{(k)})\Big)
        \Big)\cdot \Big(\prod_{l=n_d^{(k)}+1}^{j} \Big(\sI_d - \eta \nabla^2 f_{l-n_d^{(k)}}^{(k,pub)}(\rvx_{l}^{(k)}) \Big) \Big)
    \end{align}

    Therefore, plugging Eq.~\ref{eq:partial_x_i1}, Eq.~\ref{eq:partial_x_j_small_i} and Eq.~\ref{eq:partial_x_j_large_i} back to Eq.~\ref{eq:partial_x_n1_small_i} or Eq.~\ref{eq:partial_x_n1_large_i}, there is, for $i \leq n_d^{(k)}$,

    \begin{align*}
        &\frac{\partial \rvx_{n+1}^{(k)}}{\partial \rho_i^{(k)}}
        = -\eta \sI_d
        + \eta^2 \nabla^2 f_{\pi_{i+1}^{(k)}}(\rvx_{i+1}^{(k)})
        + \eta^2 \sum_{j=i+2}^{n_d^{(k)}} \nabla^2 f_{\pi_j^{(k)}}(\rvx_j^{(k)})
        \prod_{l=i+1}^{j-1} \Big( \sI_d - \eta \nabla^2 f_{\pi_l^{(k)}}(\rvx_{l}^{(k)})\Big)\\
        \nonumber
        &\quad + \eta^2 \nabla^2 f_{1}^{(k, pub)}(\rvx_{n_d^{(k)}+1}^{(k)}) \prod_{l=i+1}^{n_d^{(k)}}\Big( \sI_d - \eta \nabla^2 f_{\pi_l^{(k)}}(\rvx_l^{(k)})\Big)\\
        \nonumber
        &\quad + \eta^2 \sum_{j=n_d^{(k)}+2}^{n} \nabla^2 f_{j-n_d^{(k)}}^{(k, pub)}(\rvx_j^{(k)}) \Big(\prod_{l=i+1}^{n_d^{(k)}}\Big(\sI_d - \eta\nabla^2 f_{\pi_l^{(k)}}(\rvx_l^{(k)})\Big)\Big) \cdot\Big( \prod_{l=n_d^{(k)}+1}^{j-1}\Big(\sI_d - \eta \nabla f_{l-n_d^{(k)}}^{(k, pub)}(\rvx_l^{(k)}) \Big) \Big)
    \end{align*}
    and for $n_d^{(k)} < i \leq n$,
    \begin{align*}
        \frac{\rvx_{n+1}^{(k)}}{\partial \rho_i^{(k)}}
        &= -\eta \sI_d + \eta^2 \nabla^2 f_{i+1-n_d^{(k)}}^{(k,pub)}(\rvx_{i+1}^{(k)})
        + \eta^2 \sum_{j=i+2}^{n} \nabla^2 f_{j-n_d^{(k)}}^{(k,pub)}(\rvx_j^{(k)})\\
        \nonumber
        &\quad \cdot \Big(\prod_{i+1}^{n_d^{(k)}}\Big(\sI_d  - \eta \nabla^2 f_{\pi_k^{(k)}}(\rvx_k^{(k)}) \Big)\Big)
        \cdot \Big(\prod_{l=n_d^{(k)}+1}^{j-1} \Big(\sI_d - \eta \nabla f_{l-n_d^{(k)}}^{(k, pub)}(\rvx_l^{(k)}) \Big)
    \end{align*}
    By Assumption~\ref{ass:convexity} and~\ref{ass:smoothness}, 
    $\|\nabla^2 f_i(\rvx)\|_{op} \leq \widehat{L}^{(k)*}$ and $\|\nabla^2 f_{j}^{(k,pub)}(\rvx)\|_{op} \leq \widehat{L}^{(k)*}$, $\forall i\in [n_d^{(k)}], j\in[n-n_d^{(k)}]$ and $\forall \rvx\in\R^d$, where $\|\cdot\|_{op}$ denotes the matrix operator norm and recall that $\widehat{L}^{(k)*} = \max\{ \{L_{\pi_i^{(k)}}\}_{i=1}^{n_d^{(k)}} \cup \{\widetilde{L}^{(k)}_i\}_{i=1}^{n-n_d^{(k)}} \}$ is the maximum smoothness parameter in epoch $k$.
    
    And so if $\eta \leq \frac{1}{\widehat{L}^{(k)*}}$, $\forall i \in [n]$,
    \begin{align}
    \label{eq:op_ub}
        \left\| \frac{\rvx_{n+1}^{(k)}}{\partial \rho_i^{(k)}} \right\|_{op}
        \leq n \eta^2 \widehat{L}^{(k)*}
    \end{align}

    Moreover, by Assumption~\ref{ass:reg}, $\lambda_{min}(\nabla^2 \psi(\rvx)) \geq 0$, $\forall \rvx \in \R^d$, 
    where $\lambda_{min}$ denotes the minimum eigenvalue. And so
    \begin{align*}
        \left\| \Big(\sI_d + \eta n \nabla^2 \psi(\rvx_1^{(k+1)})\Big)^{-1} \right\|_{op} 
        \leq 1
    \end{align*}
    Hence, by Eq.~\ref{eq:diff_implicit}, 
    \begin{align*}
        \left\| \frac{\partial \rvx_{1}^{(k)}}{\partial \rho_i^{(k)}} \right\|_{op} 
        = \left\| \Big(\eta n \nabla^2 \psi(\rvx_1^{(k+1)}) + \sI_d \Big)^{-1} \frac{\partial \rvx_{n+1}^{(k)}}{\partial \rho_i^{(k)}} \right\|_{op}
        \leq n\eta^2 \widehat{L}^{(k)*}
    \end{align*}
    Since for some symmetric real matrix $\mA$, $\text{tr}(\mA) \leq d \|\mA\|_{op}$, 
    \begin{align*}
        \text{tr}(\frac{\partial \rvx_{1}^{(k+1)}}{\partial \rho_i^{(k)}})
        \leq nd \eta^2 \widehat{L}^{(k)*}
    \end{align*}
    Hence, by Eq.~\ref{eq:application_steins_lemma}, for any $i\in [n]$, conditional on $\rho_j^{(k)}, \forall j\neq i$,
    \begin{align*}
        \E_{\rho_i^{(k)}}\left[ \langle \rho_i^{(k)}, \rvx_{1}^{(k+1)} \rangle \mid\{\rho_j^{(k)}\}_{j\neq i} \right]
        = (\sigma^{(k)})^2\cdot \E_{\rho_i^{(k)}}\left[\text{tr}(\frac{\partial \rvx_1^{(k+1)}}{\partial \rho_i^{(k)}}) \mid\{\rho_j^{(k)}\}_{j\neq i} \right]
        \leq (\sigma^{(k)})^2 nd \eta^2 \widehat{L}^{(k)*}
    \end{align*}
    and by law of total expectation,
    \begin{align*}
        \E\left[\langle \rho_i^{(k)}, \rvx_1^{(k+1)}\rangle \right]
        = \E\left[ \E_{\rho_i^{(k)}}\left[ \langle \rho_i^{(k)}, \rvx_1^{(k+1)}\rangle \mid \{\rho_j\}_{j\neq i} \right] \right] 
        \leq (\sigma^{(k)})^2 nd \eta^2 \widehat{L}^{(k)*}
    \end{align*}
    and so
    \begin{align*}
        \E\left[\frac{1}{n}\sum_{i=1}^{n} \langle \rho_i^{(k)}, \rvx_1^{(k+1)} - \rvz \rangle \right]
        \leq (\sigma^{(k)})^2 nd \eta^2 \widehat{L}^{(k)*}
    \end{align*}

\end{proof}

\begin{lemma}[Noise Variance]
\label{lemma:noise_variance}
    For any epoch $k\in [K]$ and $\forall \rvz \in \R^d$, consider the injected noise $\rho_i^{(k)} \sim \gN(0, (\sigma^{(k)})^2 \sI_d)$, $\forall i\in [n]$, the variance caused by noise injection in epoch $k$ is, $\forall i\in [n]$,
    \begin{align*}
        \E\left[\Big\| \sum_{j=1}^{i} \rho_j^{(k)}\Big\|^2\right]
        \leq id (\sigma^{(k)})^2
    \end{align*}
    where the expectation is taken w.r.t. the injected noise $\{\rho_i^{(k)}\}_{i=1}^{n}$.
\end{lemma}

\begin{proof}[Proof of Lemma~\ref{lemma:noise_variance}]
    First, note that $\rho_i^{(k)}$ and $\rho_j^{(k)}$, i.e., the noise injected at step $i$ and step $j$ in epoch $k$, are independent, for any $i \neq j$. Thus,
    \begin{align*}
        \E\left[ \Big\|\sum_{j=1}^{i} \rho_j^{(k)} \Big\|^2 \right]
        &= \sum_{j=1}^{i}\E\left[\Big\| \rho_j^{(k)}\Big\|^2\right]
        + 2 \sum_{j=1}^{i} \sum_{k=j+1}^{i}
        \E\left[\langle \rho_i^{(k)}, \rho_j^{(k)}\rangle \right]\\
        &= \sum_{j=1}^{i}\E\left[\Big\| \rho_j^{(k)}\Big\|^2\right]\\
        &\leq i d (\sigma^{(k)})^2
    \end{align*}
\end{proof}

\begin{lemma}[Expected One Epoch Convergence]
\label{lemma:expected_one_epoch_convergence}
    Under Assumptions~\ref{ass:convexity},~\ref{ass:reg},~\ref{ass:dissim_partial_lipschitzness},~\ref{ass:appendix_refined_smoothness} and~\ref{ass:H_smoothness}, for any epoch $k\in [K]$, $\beta > 0$ and $\forall \rvz \in \R^d$, 
    if $\eta \leq \frac{1}{n\sqrt{10 \widehat{L}^{(k)} \widehat{L}^{(k)*}}}$ and $\rvz$ is independent of $\rho_i^{(k)}$, $\forall i\in [n]$,
    Algorithm~\ref{alg:generalized_shuffled_gradient_fm} guarantees

    \begin{align}
    \label{eq:expected_one_epoch_convergence}
        &\E\left[G(\rvx_1^{(k+1)})\right] - \E\left[G(\rvz)\right]
        \leq  \frac{1}{2 n \eta}\Big( \E\left[\|\rvz - \rvx_1^{(k)}\|^2\right]  - \E\left[ \|\rvz - \rvx_1^{(k+1)}\|^2 \right] \Big)\\
        \nonumber
        &\quad + \Big( \frac{L_H^{(k)} + \beta}{2} - \frac{\mu_{\psi}}{2} \Big) \E\left[\| \rvz - \rvx_1^{(k+1)}\|^2 \right]
        + 10\eta^2 n^2 \widehat{L}^{(k)} L \E\left[  B_F(\rvz, \rvx^*)\right]\\
        \nonumber
        &\quad + \underbrace{
            5 \eta^2 n^2 \widehat{L}^{(k)}\sigma_{any}^2
        }_{\text{Optimization Uncertainty}}
        + \underbrace{
            \frac{1}{2n^2 \beta}(C_n^{(k)})^2 
        }_{\text{Non-vanishing Dissimilarity}}
         + \underbrace{
            5\eta^2 \widehat{L}^{(k)*} \frac{1}{n} \sum_{i=n_d^{(k)} + 1}^{n-1}(C_i^{(k)})^2
        }_{\text{Vanishing Dissimilarity}}
         + \underbrace{
            6 \eta^2 nd (\sigma^{(k)})^2 \widehat{L}^{(k)*}
        }_{\text{Injected Noise}}
    \end{align}
    where the expectation is taken w.r.t. both the injected noise within epoch $k$, i.e., $\{\rho_i^{(k)}\}_{i=1}^{n}$, and the shuffling operator $\pi^{(k)}$.
\end{lemma}

\begin{proof}[Proof of~\ref{lemma:expected_one_epoch_convergence}]
    By Assumption~\ref{ass:dissim_partial_lipschitzness},
    \begin{align}
    \label{eq:exp_dissimilarity}
        &\frac{1}{n} \sum_{i=n_d^{(k)} + 1}^{n} \E_{\pi^{(k)}} \Big\|
        \sum_{j=n_d^{(k)}+1}^{i-1} \nabla f_{i-n_d^{(k)}}^{(k, pub)}(\rvz) -
        \sum_{j=n_d^{(k)}+1}^{i-1} \nabla f_{\pi_j^{(k)}}(\rvz)
        \Big\|^2
        \leq \frac{1}{n}\sum_{i=n_d^{(k)} + 1}^{n-1} (C_i^{(k)})^2
    \end{align}
    
    and by Lemma~\ref{lemma:opt_noise_bound}, for any permutation $\pi^{(k)}\in \Pi_{n}$, there is
    \begin{align}
    \label{eq:exp_opt_uncertainty}
        &\E_{\pi^{(k)}} \Big[\frac{1}{n} \Big(\sum_{i=2}^{n_d^{(k)}} L_{\pi_i^{(k)}} \Big\| \sum_{j=1}^{i-1} \nabla f_{\pi_j^{(k)}}(\rvx^*)\Big\|^2
        + \sum_{i=n_d^{(k)}+1}^{n}\widetilde{L}_{i-n_d^{(k)}}^{(k)}\Big\| \sum_{j=1}^{i-1}\nabla f_{\pi_j^{(k)}}(\rvx^*)\Big\|^2 \Big)\Big]
        \leq n^2 \widehat{L}^{(k)} \sigma_{any}^2
    \end{align}


    where the expectation is taken w.r.t. the shuffling operator $\pi^{(k)}$.

    Moreover, by Lemma~\ref{lemma:noise_variance},
    \begin{align}
    \nonumber
        &\E_{\pi^{(k)}}\Big[\frac{1}{n} \Big( \sum_{i=2}^{n_d^{(k)}} L_{\pi_i^{(k)}} \Big\| \sum_{j=1}^{i-1} \rho_j^{(k)}\Big\|^2
        + \sum_{i=n_d^{(k)}+1}^{n} \widetilde{L}_{i-n_d^{(k)}}^{(k)}\Big\| \sum_{j=1}^{i-1} \rho_j^{(k)}\Big\|^2\Big)\Big]\\
    \nonumber
        &\leq \frac{1}{n} \Big( \sum_{i=2}^{n_d^{(k)}} L_{\pi_i^{(k)}} (i-1) d(\sigma^{(k)})^2
        + \sum_{i=n_d^{(k)}+1}^{n} \widetilde{L}_{i-n_d^{(k)}}^{(k)} (i-1) d (\sigma^{(k)})^2\Big)\\
    \label{eq:exp_noise}
        &\leq \frac{1}{n} nd (\sigma^{(k)})^2 \Big( \sum_{i=2}^{n_d^{(k)}} L_{\pi_i^{(k)}} + \sum_{i=n_d^{(k)}+1}^{n} \widetilde{L}_{i-n_d^{(k)}}^{(k)} \Big)
        \leq nd (\sigma^{(k)})^2 \widehat{L}^{(k)}
    \end{align}


    where the expectation is taken w.r.t. the injected noise $\{\rho_j^{(k)}\}_{j=1}^{n}$.

    Following Eq.~\ref{eq:exp_dissimilarity},~\ref{eq:exp_opt_uncertainty},~\ref{eq:exp_noise}, and Lemma~\ref{lemma:one_epoch_convergence},~\ref{lemma:noise_bias}, for any $\rvz\in\R^{d}$,
    
    \begin{align}
        &\E\left[ G(\rvx_1^{(k+1)})\right] - \E\left[ G(\rvz) \right]\\
        \nonumber
        &\leq \frac{1}{2 n \eta}\Big( \E\left[\|\rvz - \rvx_1^{(k)}\|^2\right]  - \E\left[ \|\rvz - \rvx_1^{(k+1)}\|^2 \right] \Big)
        + \Big( \frac{L_H^{(k)} + \beta}{2} - \frac{\mu_{\psi}}{2} \Big) \E\left[\| \rvz - \rvx_1^{(k+1)}\|^2 \right]\\
        \nonumber
        &\quad  + 10\eta^2 n^2 \widehat{L}^{(k)} L \E\left[  B_F(\rvz, \rvx^*)\right]
        + \frac{1}{2n^2 \beta}(C_n^{(k)})^2 
         + 5\eta^2 \widehat{L}^{(k)*} \frac{1}{n} \sum_{i=n_d^{(k)} + 1}^{n-1}(C_i^{(k)})^2
            + 5 \eta^2 n^2 \widehat{L}^{(k)}\sigma_{any}^2 \\
        \nonumber
        &\quad + 
        \eta^2 (\sigma^{(k)})^2 nd \widehat{L}^{(k)*}
        + 5\eta^2 nd (\sigma^{(k)})^2 \widehat{L}^{(k)}
    \end{align}
    where the expectation is taken w.r.t. both the injected noise within epoch $k$, i.e., $\{\rho_i^{(k)}\}_{i=1}^{n}$, and the shuffling operator $\pi^{(k)}$.
    Combining the last two terms and note that $\widehat{L}^{(k)} \leq \widehat{L}^{(k)*}$ yields the inequality (\ref{eq:expected_one_epoch_convergence}) in the above lemma statement. 
    
\end{proof}

\subsection{Convergence Across $K$ Epochs}
\label{subsec:appendix_k_epoch_convergence}

Now that we have the expected convergence rate for one epoch, we follow a similar approach as in~\cite{liu2024last_iterate_shuffled_gradient}, first showing the convergence for any arbitrary number of epochs $s\in [K]$ by picking proper $\rvz$ points as the virtual sequence and using a weighted telescoping sum in Lemma~\ref{lemma:convergence_across_arbitrary_epochs} and then showing the convergence for $K$ epochs in Theorem~\ref{thm:convergence_generalized_shufflg_fm_appendix}.

\begin{lemma}[Convergence Across Arbitrary Epochs]
\label{lemma:convergence_across_arbitrary_epochs}
    Under Assumptions~\ref{ass:convexity},~\ref{ass:reg},~\ref{ass:dissim_partial_lipschitzness},~\ref{ass:appendix_refined_smoothness} and~\ref{ass:H_smoothness}, for any number of epochs $s \in [K]$ and $\beta > 0$, 
    if $\mu_{\psi} \geq L_H^{(k)} + \beta$, $\forall k\in [s]$, and $\eta \leq \frac{1}{n\sqrt{10 \max_{k\in[s]} (\widehat{L}^{(k)} \widehat{L}^{(k)*}}) }$, Algorithm~\ref{alg:generalized_shuffled_gradient_fm} guarantees
    

    \begin{align*}
        &\E\left[G(\rvx_1^{(s+1)})\right] - G(\rvx^*)
        \leq \frac{1}{2\eta n s} \| \rvx^* - \rvx_1^{(1)}\|^2
        + 10 \eta^2 n^2 L\max_{k\in[s]}\widehat{L}^{(k)} \sum_{k=2}^{s} \frac{1}{s+2-k} \E\left[B_F(\rvx_1^{(k)}, \rvx^*)\right]\\
        \nonumber
        &\quad + 5\eta^2 n^2 \sigma_{any}^2 \sum_{k=1}^{s} \frac{ \widehat{L}^{(k)} }{s+1-k}
        + \frac{1}{2n^2 \beta}\sum_{k=1}^{s} \frac{ (C_n^{(k)})^2 }{s+1-k}\\
        \nonumber
        &\quad + 5\eta^2 \sum_{k=1}^{s} \frac{ \widehat{L}^{(k)*} \frac{1}{n}\sum_{i=n_d^{(k)}+1}^{n-1}(C_i^{(k)})^2} {s+1-k}
        + 6\eta^2 nd \sum_{k=1}^{s} \frac{ (\sigma^{(k)})^2 \widehat{L}^{(k)*} }{s+1-k}
    \end{align*}

\end{lemma}

\begin{proof}[Proof of Lemma~\ref{lemma:convergence_across_arbitrary_epochs}]
    Fix an arbitrary number of epochs $s\in [K]$.
    
    If $\mu_\psi \geq L_H^{(k)} + \beta$, 
    $\forall k\in [s]$, $\eta \leq \frac{1}{n\sqrt{10 \max_{k\in [s]} (\widehat{L}^{(k)} \widehat{L}^{(k)*}})}$, and $\rvz$ independent of $\{\rho_i^{(k)}\}_{i=1}^{n}$,
    then by Lemma~\ref{lemma:expected_one_epoch_convergence}, for any $k\in [s]$,
    \begin{align}
    \label{eq:expected_one_epoch_convergence_strong_reg}
    &\E\left[G(\rvx_1^{(k+1)})\right] - \E\left[G(\rvz)\right]\\
        \nonumber
        &\leq  \frac{1}{2 n \eta}\Big( \E\left[\|\rvz - \rvx_1^{(k)}\|^2\right]  - \E\left[ \|\rvz - \rvx_1^{(k+1)}\|^2 \right] \Big)
        + 10\eta^2 n^2 \widehat{L}^{(k)} L \E\left[  B_F(\rvz, \rvx^*)\right]\\
        \nonumber
        &\quad + 5 \eta^2 n^2 \widehat{L}^{(k)}\sigma_{any}^2
        + \frac{1}{2n^2 \beta}(C_n^{(k)})^2 
         +  5\eta^2 \widehat{L}^{(k)*} \frac{1}{n} \sum_{i=n_d^{(k)}+1}^{n-1}(C_i^{(k)})^2
         + 6 \eta^2 nd (\sigma^{(k)})^2 \widehat{L}^{(k)*}
    \end{align}

    Define the non-decreasing sequence
    \begin{align*}
        v_k = \frac{1}{s+1-k}, \forall k\in [s], \quad v_0 = v_1 = \frac{1}{s}
    \end{align*}
    and the auxiliary points
    \begin{align}
    \label{eq:z_def}
        \rvz^{(0)} = \rvx^*, \quad \rvz^{(k)} = \Big( 1- \frac{v_{k-1}}{v_k}\Big) \rvx_1^{(k)} + \frac{v_{k-1}}{v_k} \rvz^{(k-1)}, \forall k\in [s]
    \end{align}
    Equivalently, $\rvz^{(k)}$ can be re-written as
    \begin{align}
    \label{eq:z_eq_form}
        \rvz^{(k)} = \frac{v_0}{v_k}\rvx^* + \sum_{l=1}^{k} \frac{v_l - v_{l-1}}{v_k} \rvx_1^{(l)}, \forall k\in [0]\cup [s]
    \end{align}
    
    Note that for an epoch $k$, $\rvz^{(k)}$ only depends on $\rvx^*$ and $\rvx_1^{(l)}$, for $l \leq k$, and hence by the update rule, $\rvz^{(k)}$ is independent of $\rho_i^{(k)}$, $\forall i\in [n]$.
    And so for any $k\in [s]$, we can choose $\rvz = \rvz^{(k)}$ in Eq.~\ref{eq:expected_one_epoch_convergence_strong_reg} and this leads to

    \begin{align}
    \label{eq:expected_one_epoch_convergence_strong_reg_z}
    &\E\left[G(\rvx_1^{(k+1)})\right] - \E\left[G(\rvz)\right]\\
        \nonumber
        &\leq  \frac{1}{2 n \eta}\Big( \E\left[\|\rvz^{(k)} - \rvx_1^{(k)}\|^2\right]  - \E\left[ \|\rvz^{(k)} - \rvx_1^{(k+1)}\|^2 \right] \Big)
        + 10\eta^2 n^2 \widehat{L}^{(k)} L \E\left[  B_F(\rvz^{(k)}, \rvx^*)\right]\\
        \nonumber
        &\quad + 5 \eta^2 n^2 \widehat{L}^{(k)}\sigma_{any}^2
        + \frac{1}{2n^2 \beta}(C_n^{(k)})^2 
         +  5\eta^2 \widehat{L}^{(k)*} \frac{1}{n} \sum_{i=n_d^{(k)}+1}^{n-1}(C_i^{(k)})^2
         + 6 \eta^2 nd (\sigma^{(k)})^2 \widehat{L}^{(k)*}
    \end{align}

    Note that by Eq.~\ref{eq:z_def},
    \begin{align}
        \|\rvz^{(k)} - \rvx_1^{(k)}\|^2 = \frac{v_{k-1}^2}{v_k^2} \| \rvz^{(k-1)} - \rvx_1^{(k)}\|^2
        \leq \frac{v_{k-1}}{v_k}\|\rvz^{(k-1)} - \rvx_1^{(k)}\|^2
    \end{align}
    where the last inequality is due to $v_{k-1} \leq v_k$. Hence, following Eq.~\ref{eq:expected_one_epoch_convergence_strong_reg_z},

    \begin{align}
    \label{eq:convergence_across_epochs_interm}
        &v_k \cdot \left(\E\left[G(\rvx_1^{(k+1)})\right] - \E\left[G(\rvz)\right] \right)\\
        \nonumber
        &\leq  \frac{1}{2 n \eta}\Big( \E\left[ v_{k-1} \|\rvz^{(k-1)} - \rvx_1^{(k)}\|^2\right]  - v_k \E\left[ \|\rvz^{(k)} - \rvx_1^{(k+1)}\|^2 \right] \Big)
        + 10 v_k \eta^2 n^2 \widehat{L}^{(k)} L \E\left[  B_F(\rvz^{(k)}, \rvx^*)\right]\\
        \nonumber
        &\quad + 5 v_k \eta^2 n^2 \widehat{L}^{(k)}\sigma_{any}^2
        + v_k \frac{1}{2n^2 \beta}(C_n^{(k)})^2 
         +  5 v_k \eta^2 \widehat{L}^{(k)*} \frac{1}{n} \sum_{i=n_d^{(k)}+1}^{n-1}(C_i^{(k)})^2
         + 6 v_k \eta^2 nd (\sigma^{(k)})^2 \widehat{L}^{(k)*}
    \end{align}
    Summing Eq.~\ref{eq:convergence_across_epochs_interm} from $k = 1$ to $s$ to obtain
    \begin{align}
    \label{eq:telescoping_sum_result}
        &\sum_{k=1}^{s} v_k \cdot \left(\E\left[G(\rvx_1^{(k+1)})\right] - \E\left[G(\rvz)\right] \right)
        \leq \frac{1}{2 n \eta} \Big(\E\left[ v_0 \| \rvz^{(0)} - \rvz_1^{(1)}\|^2 \right]
        - v_{s}\E\left[\| \rvz^{(s)} - \rvx_1^{(s+1)}\|^2\right]\Big)\\
        \nonumber
        &\quad + 10 \eta^2 n^2 L \sum_{k=1}^{s} \widehat{L}^{(k)} v_k \E\left[B_F(\rvz^{(k)}, \rvx^*)\right]
        + 5\eta^2 n^2 \sigma_{any}^2 \sum_{k=1}^{s} v_k \widehat{L}^{(k)}
        + \frac{1}{2n^2 \beta}\sum_{k=1}^{s} v_k (C_n^{(k)})^2\\
    \nonumber
        &\quad + 5\eta^2 \sum_{k=1}^{s} v_k \widehat{L}^{(k)*} \frac{1}{n}\sum_{i=n_d^{(k)}+1}^{n-1}(C_i^{(k)})^2
        + 6\eta^2 nd \sum_{k=1}^{s} v_k (\sigma^{(k)})^2 \widehat{L}^{(k)*}
    \end{align}

     Note since $\|\rvz^{(s)} - \rvx_1^{(s+1)}\|^2 \geq 0$ and $\rvz^{(0)} = \rvx^*, v_0 = \frac{1}{s}$,
    \begin{align}
    \label{eq:convergence_across_epochs_tele_sum}
        \frac{1}{2\eta n}\Big( v_0 \E\left[ \| \rvz^{(0)} - \rvx_1^{(1)}\|^2 \right] - v_s \E\left[ \| \rvz^{(s)} - \rvx_1^{(s+1)} \|^2 \right] \Big)
        \leq \frac{1}{2\eta n s} \| \rvx^* - \rvx_1^{(1)}\|^2 
    \end{align}

    We first bound the L.H.S. of Eq.~\ref{eq:telescoping_sum_result}.
    By Assumption~\ref{ass:convexity} and~\ref{ass:reg}, the objective function $G(\rvx) = F(\rvx) + \psi(\rvx)$ is convex, and hence, by Eq.~\ref{eq:z_eq_form}, there is
    \begin{align*}
        G(\rvz^{(k)}) \leq \frac{v_0}{v_k} G(\rvx^*) + \sum_{l=1}^{k}\frac{v_l - v_{l-1}}{v_k} G(\rvx_1^{(l)})
        = G(\rvx^*) + \sum_{l=1}^{k}\frac{v_l - v_{l-1}}{v_k} \Big( G(\rvx_1^{(l)}) - G(\rvx^*)\Big)
    \end{align*}
    which implies
    \begin{align}
    \nonumber
        &\sum_{k=1}^{s} v_k \Big( G(\rvx_1^{(k+1)}) - G(\rvz^{(k)})\Big)\\
        \nonumber
        &\geq \sum_{k=1}^{s} \Big( v_k \Big(G(\rvx_1^{(k+1)}) - G(\rvx^*)\Big) - \sum_{l=1}^{k} (v_l - v_{l-1}) \Big(G(\rvx_1^{(l)}) - G(\rvx^*)\Big) \Big)\\
    \nonumber
        &\geq \sum_{k=1}^{s} v_k \Big(G(\rvx_1^{(k+1)}) - G(\rvx^*)\Big)
        - \sum_{k=1}^{s}\sum_{l=1}^{k} (v_l - v_{l-1}) \Big(G(\rvx_1^{(l)}) - G(\rvx^*)\Big)\\
    \nonumber
        &= \sum_{k=1}^{s} v_k \Big(G(\rvx_1^{(k+1)}) - G(\rvx^*)\Big)
        - \sum_{l=1}^{s}(s+1-l) (v_l - v_{l-1}) \Big(G(\rvx_1^{(l)}) - G(\rvx^*)\Big)\\
    \nonumber
        &= v_s \Big( G(\rvx_1^{(s+1)}) - G(\rvx^*) \Big)
        + \sum_{k=1}^{s-1} \frac{1}{s+1-k} \Big(G(\rvx_1^{(k+1)}) - G(\rvx^*)\Big)\\
        \nonumber
        &\quad - s(v_1 - v_0)\Big( G(\rvx_1^{(1)}) - G(\rvx^*) \Big)
        - \sum_{l=2}^{s}(s+1-l)(\frac{1}{s+1-l} - \frac{1}{s+2-l})\Big(G(\rvx_1^{(l)}) - G(\rvx^*)\Big)\\
    \nonumber
        &= v_s \Big( G(\rvx_1^{(s+1)}) - G(\rvx^*) \Big)
        + \sum_{k=2}^{s} \frac{1}{s+2-k} \Big( G(\rvx_1^{(k)}) - G(\rvx^*) \Big)\\
        \nonumber
        &\quad - s(v_1 - v_0) \Big( G(\rvx_1^{(1)}) - G(\rvx^*)\Big)
        - \sum_{l=2}^{s}\frac{1}{s+2-l}\Big(G(\rvx_1^{(l)}) - G(\rvx^*)\Big)\\
    \nonumber
        &= v_s \Big( G(\rvx_1^{(s+1)}) - G(\rvx^*) \Big) - s(v_1 - v_0)\Big( G(\rvx_1^{(1)}) - G(\rvx^*)\Big)
    \end{align}
    Note that $v_1 = v_0$ and $v_s = 1$ by definition, and so taking expectation of both sides,
    \begin{align}
    \label{eq:convergence_across_arbitrary_epochs_lhs}
        \sum_{k=1}^{s} v_k\Big( \E\left[G(\rvx_1^{(k+1)}) \right] - \E\left[G(\rvz^{(k)})\right]\Big)
        \geq \E\left[G(\rvx_1^{(s+1)})\right] - G(\rvx^*)
    \end{align}

    We now bound the term involving $\E\left[B_F(\rvz^{(k)}, \rvx^*)\right]$ in the R.H.S. of Eq.~\ref{eq:telescoping_sum_result}. 
    By the convexity of $B_F(\cdot, \rvx^*)$ fixing the second argument (due to $F$ being convex), and Eq.~\ref{eq:z_eq_form}, 
    \begin{align*}
        B_F(\rvz^{(k)}, \rvx^*) \leq \frac{v_0}{v_k}B_F(\rvx^*, \rvx^*) + \sum_{l=1}^{k}\frac{v_l - v_{l-1}}{v_k}B_F(\rvx_1^{(l)}, \rvx^*)
        = \sum_{l=1}^{k}\frac{v_l - v_{l-1}}{v_k}B_F(\rvx_1^{(l)}, \rvx^*)
    \end{align*}
    which implies
    \begin{align}
    \nonumber
        &10 \eta^2n^2 L \sum_{k=1}^{s} v_k \widehat{L}^{(k)} \E\left[B_F(\rvz^{(k)}, \rvx^*)\right]
        \leq 10 \eta^2 n^2
        L \max_{k\in[s]}\widehat{L}^{(k)}
        \sum_{k=1}^{s} \sum_{l=1}^{k} (v_l - v_{l-1})\E\left[B_F(\rvx_1^{(l)}, \rvx^*)\right]\\
    \nonumber
        &= 10 \eta^2 n^2 L \max_{k\in[s]}\widehat{L}^{(k)}
        \sum_{l=1}^{s} (s+1-l)(v_l - v_{l-1}) \E\left[B_F(\rvx_1^{(l)}, \rvx^*)\right] \\
    \nonumber
        &= 10 \eta^2 n^2 L \max_{k\in[s]}\widehat{L}^{(k)}
        \sum_{l=2}^{s} (s+1-l)(\frac{1}{s+1-l} - \frac{1}{s+2-l}) \E\left[B_F(\rvx_1^{(l)}, \rvx^*)\right] \\
    \nonumber
        &= 10 \eta^2 n^2 L \max_{k\in[s]}\widehat{L}^{(k)} \sum_{l=2}^{s} \frac{1}{s+2-l} \E\left[B_F(\rvx_1^{(l)}, \rvx^*)\right]\\
    \label{eq:convergence_across_epochs_bregmandiv}
        &= 10 \eta^2 n^2 L \max_{k\in[s]}\widehat{L}^{(k)}   \sum_{l=2}^{s} v_{l-1} \E\left[B_F(\rvx_1^{(l)}, \rvx^*)\right]
    \end{align}

    Therefore, plugging Eq.~\ref{eq:convergence_across_epochs_tele_sum}, Eq.~\ref{eq:convergence_across_arbitrary_epochs_lhs} and Eq.~\ref{eq:convergence_across_epochs_bregmandiv} back to Eq.~\ref{eq:telescoping_sum_result}, there is, for any fixed epoch $s\in [K]$
    \begin{align*}
        &\E\left[G(\rvx_1^{(s+1)})\right] - G(\rvx^*)\\
        \nonumber
        &\leq \frac{1}{2\eta n s} \| \rvx^* - \rvx_1^{(1)}\|^2
        + 10 \eta^2 n^2 L\max_{k\in[s]}\widehat{L}^{(k)} \sum_{k=2}^{s} \frac{1}{s+2-k} \E\left[B_F(\rvx_1^{(k)}, \rvx^*)\right]\\
        \nonumber
        &\quad + 5\eta^2 n^2 \sigma_{any}^2 \sum_{k=1}^{s} \frac{ \widehat{L}^{(k)} }{s+1-k}
        + \frac{1}{2n^2 \beta}\sum_{k=1}^{s} \frac{ (C_n^{(k)})^2 }{s+1-k}\\
        \nonumber
        &\quad + 5\eta^2 \sum_{k=1}^{s} \frac{ \widehat{L}^{(k)*} \frac{1}{n}\sum_{i=n_d^{(k)}+1}^{n-1}(C_i^{(k)})^2} {s+1-k}
        + 6\eta^2 nd \sum_{k=1}^{s} \frac{ (\sigma^{(k)})^2 \widehat{L}^{(k)*} }{s+1-k}
    \end{align*}
\end{proof}

\begin{theorem}[Convergence of Generalized Shuffled Gradient Framework (Re-statement of Theorem~\ref{thm:convergence_generalized_shufflg_fm})]
\label{thm:convergence_generalized_shufflg_fm_appendix}

    Under Assumptions~\ref{ass:convexity},~\ref{ass:reg},~\ref{ass:dissim_partial_lipschitzness},~\ref{ass:appendix_refined_smoothness} and~\ref{ass:H_smoothness}, for $\beta > 0$, 
    if $\mu_{\psi} \geq L_H^{(k)} + \beta$, $\forall k\in [K]$, and 
     $\eta \leq \frac{1}{2n \sqrt{10 \bar{L}^* \max_{k\in [K]}\widehat{L}^{(k)*} (1+\log K)}}$, 
    where $\bar{L}^* = \max\{L, \max_{k\in[K]} \widehat{L}^{(k)}\}$,
    Algorithm~\ref{alg:generalized_shuffled_gradient_fm} guarantees

    \begin{align}
    \label{eq:general_convergence_appendix}
        &\E\left[ G(\rvx_1^{(K+1)}) \right] - G(\rvx^*)
        \leq \underbrace{ \frac{\| \rvx_1^{(1)} - \rvx^*\|^2 }{\eta n K} 
        }_{\text{Initialization}}
        + \underbrace{ 10 \eta^2 n^2 \sigma_{any}^2 (1+\log K)\max_{k\in[K]} \widehat{L}^{(k)} 
        }_{\text{Optimization Uncertainty}} + 2 M
    \end{align}
    where
    \begin{align*}
        M = \max_{s\in [K]} \Big(
        \underbrace{ \frac{1}{2n^2 \beta}\sum_{k=1}^{s} \frac{ (C_n^{(k)})^2 }{s+1-k}
        }_{\text{Non-vanishing Dissimilarity}}
        + \underbrace{5\eta^2 \sum_{k=1}^{s} \frac{ \widehat{L}^{(k)*} \frac{1}{n}\sum_{i=n_d^{(k)}+1}^{n-1}(C_i^{(k)})^2} {s+1-k} }_{\text{Vanishing Dissimilarity}}
        + \underbrace{ 6\eta^2 nd \sum_{k=1}^{s} \frac{ (\sigma^{(k)})^2 \widehat{L}^{(k)*} }{s+1-k} }_{\text{Injected Noise}} \Big)
    \end{align*}
    and the expectation is taken w.r.t. the injected noise $\{\rho_i^{(k)}\}$ and the order of samples $\pi^{(k)}$, $\forall i\in [n], k\in [K]$.
\end{theorem}

\begin{proof}[Proof of Theorem~\ref{thm:convergence_generalized_shufflg_fm_appendix}]
    Taking the learning rate $\eta \leq \frac{1}{2n \sqrt{20 \bar{L}^* \max_{k \in [K]}\widehat{L}^{(k)*} (1+\log K)}}$, 
    where $\bar{L}^* = \max\{L, \max_{k\in[K]} \widehat{L}^{(k)}\}$, i.e., the max average smoothness parameters, satisfies the condition of Lemma~\ref{lemma:convergence_across_arbitrary_epochs}, and so for any number of epochs $s \in [K]$,
    \begin{align*}
        &\E\left[G(\rvx_1^{(s+1)})\right] - G(\rvx^*)
        \leq \frac{1}{2\eta n s} \| \rvx^* - \rvx_1^{(1)}\|^2
        + 10 \eta^2 n^2 L\max_{k\in[s]}\widehat{L}^{(k)} \sum_{k=2}^{s} \frac{1}{s+2-k} \E\left[B_F(\rvx_1^{(k)}, \rvx^*)\right]\\
        \nonumber
        &\quad + 5\eta^2 n^2 \sigma_{any}^2 \sum_{k=1}^{s} \frac{ \widehat{L}^{(k)} }{s+1-k}
        + \frac{1}{2n^2 \beta}\sum_{k=1}^{s} \frac{ (C_n^{(k)})^2 }{s+1-k}\\
        \nonumber
        &\quad + 5\eta^2 \sum_{k=1}^{s} \frac{ \widehat{L}^{(k)*} \frac{1}{n}\sum_{i=n_d^{(k)}+1}^{n-1}(C_i^{(k)})^2} {s+1-k}
        + 6\eta^2 nd \sum_{k=1}^{s} \frac{ (\sigma^{(k)})^2 \widehat{L}^{(k)*} }{s+1-k}
    \end{align*}

    Note that $\sum_{k=1}^{s} v_k = \sum_{k=1}^{s} \frac{1}{s+1-k} = \sum_{k=1}^{s} \frac{1}{k}\leq 1+\log s$, and so
    \begin{align*}
        5 \eta^2 n^2 \sigma_{any}^2 \sum_{k=1}^{s} \frac{\widehat{L}^{(k)}}{s+1-k}
        \leq \eta^2 n^2 \sigma_{any}^2  (1+\log s) \max_{k\in [s]} \widehat{L}^{(k)}
    \end{align*}
    Hence,
    \begin{align*}
        &\E\left[G(\rvx_1^{(s+1)})\right] - G(\rvx^*) 
        \leq \frac{1}{2\eta n s}  \| \rvx^* - \rvx_1^{(1)}\|^2 
        + 10 \eta^2 n^2 L\max_{k\in[s]}\widehat{L}^{(k)} \sum_{k=2}^{s} \frac{1}{s+2-k} \E\left[B_F(\rvx_1^{(k)}, \rvx^*)\right]\\
        \nonumber
        &\quad + 5\eta^2 n^2 \sigma_{any}^2 (1+\log s) \max_{k\in [s]}\widehat{L}^{(k)}
        + \frac{1}{2n^2 \beta}\sum_{k=1}^{s} \frac{ (C_n^{(k)})^2 }{s+1-k}\\
        \nonumber
        &\quad + 5\eta^2 \sum_{k=1}^{s} \frac{ \widehat{L}^{(k)*} \frac{1}{n}\sum_{i=n_d^{(k)}+1}^{n-1}(C_i^{(k)})^2} {s+1-k}
        + 6\eta^2 nd \sum_{k=1}^{s} \frac{ (\sigma^{(k)})^2 \widehat{L}^{(k)*} }{s+1-k}
    \end{align*}

    By the optimality condition, $\nabla G(\rvx^*) = \nabla F(\rvx^*) + \nabla \psi(\rvx^*) = \mathbf{0}$. Thus, for any $s\in [K]$,
    \begin{align*}
        \E\left[G(\rvx_1^{(s+1)})\right] - G(\rvx^*) 
        &\geq \E\left[ G(\rvx_1^{(s+1)})\right] - \E\left[ G(\rvx^*)\right] - \E\left[\langle \nabla F(\rvx^*) + \nabla \psi(\rvx^*), \rvx_1^{(s+1)} - \rvx^* \rangle \right] \\
        &= \E\left[ B_F(\rvx_1^{(s+1)},\rvx^*)\right] + \E\left[ B_{\psi}(\rvx_1^{(s+1)}, \rvx^*)\right]
        \geq \E\left[ B_F(\rvx_1^{(s+1)}, \rvx^*)\right]
    \end{align*}
    which implies that for any $s\in [K]$,
    \begin{align*}
        &\E\left[B_F(\rvx_1^{(s+1)}), \rvx^*)\right]
        \leq \frac{1}{2\eta n s} \| \rvx^* - \rvx_1^{(1)}\|^2
        + 10 \eta^2 n^2 L\max_{k\in[s]}\widehat{L}^{(k)} \sum_{k=2}^{s} \frac{1}{s+2-k} \E\left[B_F(\rvx_1^{(k)}, \rvx^*)\right]\\
    \nonumber
        &\quad + 5\eta^2 n^2 \sigma_{any}^2 (1+\log s) \max_{k\in [s]}\widehat{L}^{(k)}
        + \frac{1}{2n^2 \beta}\sum_{k=1}^{s} \frac{ (C_n^{(k)})^2 }{s+1-k}\\
    \nonumber
        &\quad + 5\eta^2 \sum_{k=1}^{s} \frac{ \widehat{L}^{(k)*} \frac{1}{n}\sum_{i=n_d^{(k)}+1}^{n-1}(C_i^{(k)})^2} {s+1-k}
        + 6\eta^2 nd \sum_{k=1}^{s} \frac{ (\sigma^{(k)})^2 \widehat{L}^{(k)*} }{s+1-k}
    \end{align*}
    Note that $\max_{k\in [s]}\widehat{L}^{(k)} \leq \max_{k\in [K]} \widehat{L}^{(k)}$.
    
    Now we apply Lemma~\ref{lemma:bregmandiv_relationship} with
    \begin{itemize}[itemsep=0mm]
        \item $d^{(s+1)}=\begin{cases}
                \E\left[ B_F(\rvx_1^{(s+1)}, \rvx^*)\right] &s \in [K-1]\\
                \E\left[ F(\rvx_1^{(K+1)})  \right] - \E\left[ F(\rvx^*) \right]
                &s = K
            \end{cases}$
        \item $e^{(k)} = \frac{1}{2n^2 \beta}(C_n^{(k)})^2 + 5\eta^2 \widehat{L}^{(k)*}\frac{1}{n}\sum_{i=1}^{n-1}(C_i^{(k)})^2 + 6\eta^2 nd (\sigma^{(k)})^2 \widehat{L}^{(k)*}, \forall k\in [K]$
        \item $a = \frac{1}{2\eta n}\| \rvx_1^{(1)} - \rvx^*\|^2$
        \item $b = 5\eta^2 n^2 \sigma_{any}^2(1+\log s) \max_{k\in [K]} \widehat{L}^{(k)}$
        \item $c = 10\eta^2 n^2 L \max_{k\in[K]} \widehat{L}^{(k)}$
    \end{itemize}
    to obtain
    \begin{align*}
        &\E\left[ G(\rvx_1^{(K+1)}) \right] - G(\rvx^*)\\
        \nonumber
        &\leq \Big( \frac{1}{2\eta n K} \|\rvx^* - \rvx_1^{(1)}\|^2
        + 5\eta^2 n^2 \sigma_{any}^2 (1+\log K)\max_{k\in[K]} \widehat{L}^{(k)} + M\Big)
        \sum_{i=0}^{K-1} \Big(20 \eta^2 n^2 L \max_{k\in [K]}\widehat{L}^{(k)}(1+\log K) \Big)^{i}
    \end{align*}
    where 
    \begin{align*}
        M = \max_{s\in [K]} \Big(
        \frac{1}{2n^2 \beta}\sum_{k=1}^{s} \frac{ (C_n^{(k)})^2 }{s+1-k}
        + 5\eta^2 \sum_{k=1}^{s} \frac{ \widehat{L}^{(k)*} \frac{1}{n}\sum_{i=n_d^{(k)}+1}^{n-1}(C_i^{(k)})^2} {s+1-k}
        + 6\eta^2 nd \sum_{k=1}^{s} \frac{ (\sigma^{(k)})^2 \widehat{L}^{(k)*} }{s+1-k} \Big)
    \end{align*}

    By setting $\eta \leq \frac{1}{2n \sqrt{10 \bar{L}^{*} \max_{k\in [K]}\widehat{L}^{(k)*} (1+\log K)}}$, 
    where $\bar{L}^{*} = \max\{L, \max_{k\in[K]} \widehat{L}^{(k)}\}$,
    there is
    \begin{align*}
        \sum_{i=0}^{K-1} \Big(20 \eta^2 n^2 L \max_{k\in [K]}\widehat{L}^{(k)}(1+\log K) \Big)^{i}
        \leq \sum_{i=0}^{K-1} \Big(
            \frac{n^2 L \max_{k\in[K]}\widehat{L}^{(k)} (1+\log K)}{2 n^2 \bar{L}^* \max_{k\in[K]} \widehat{L}^{(k)*} (1+\log K)}
        \Big)^{i}
        \leq \sum_{i=0}^{\infty} \frac{1}{2^{i}} = 2
    \end{align*}

    Therefore,
    \begin{align*}
        &\E\left[ G(\rvx_1^{(K+1)}) \right] - G(\rvx^*)
        \leq \frac{\| \rvx_1^{(1)} - \rvx^*\|^2 }{\eta n K} 
        + 10 \eta^2 n^2 \sigma_{any}^2 (1+\log K)\max_{k\in[K]} \widehat{L}^{(k)} + 2 M
    \end{align*}
    
\end{proof}

\subsection{Recovering Non-private Shuffled Gradient Methods}
\label{subsec:appendix_non_private_shuffled_g}

We note that the convergence bound presented in Theorem~\ref{thm:convergence_generalized_shufflg_fm_appendix} recovers, as a special case, the convergence result for non-private shuffled gradient methods in Theorem 4.4 of~\cite{liu2024last_iterate_shuffled_gradient}.
This is formalized in the following corollary. Recall $L = \frac{1}{n}\sum_{i=1}^{n} L_i$ and $L^{*} = \max_{i\in[n]}\{L_i\}_{i=1}^{n}
$ are the average and maximum smoothness constants of the target objective.

\begin{corollary}[Convergence of Non-private Shuffled Gradient Methods]
\label{corollary:convergence_non_private_shuffled_g}
    Under Assumptions \ref{ass:convexity}, \ref{ass:reg}, \ref{ass:dissim_partial_lipschitzness} and \ref{ass:appendix_refined_smoothness}, if the learning rate satisfies $\eta \leq \frac{1}{2n\sqrt{10 L L^{*}}}$, Algorithm~\ref{alg:generalized_shuffled_gradient_fm} guarantees
    \begin{align}
        \label{eq:raw_convergence_non_private_shuffled_g}&\E\left[G(\rvx_1^{(K+1)})\right] - G(\rvx^*)
        \leq \underbrace{\frac{\| \rvx_1^{(1)} - \rvx^{*}\|^2}{\eta n K}}_{\text{Initialization}}
        + \underbrace{10 \eta^2 n^2 \sigma_{any}^2 (1+\log K) L }_{\text{Optimization Uncertainty}}
    \end{align}
   Furthermore, choosing the learning rate as $\eta = \min\{\frac{1}{n \sqrt{L L^{*} (1+\log K)}}, \frac{\|\rvx^{*} - \rvx_1^{(1)}\|^{2/3}}{n L^{1/3} \sigma_{any}^{2/3} K^{1/3}(1+\log K)^{1/3} }\}$ yields the following convergence bound:
    \begin{align}
    \label{eq:convergence_non_private_shuffled_g}
    &\E\left[G(\rvx_1^{(K+1)})\right] - G(\rvx^{*})\\
        \nonumber
        &=\gO\left( \frac{\|\rvx^{*} - \rvx_1^{(1)} \|^{2}  \sqrt{L L^{*} (1+\log K)}}{K} \right)
        + \gO\left( \frac{\| \rvx^{*} - \rvx_1^{(1)} \|^{4/3} L^{1/3} \sigma_{any}^{2/3} (1+\log K)^{1/3} }{K^{2/3}} \right)
    \end{align}
\end{corollary}

\begin{proof}[Proof of Corollary~\ref{corollary:convergence_non_private_shuffled_g}]

    The convergence bound in Eq.~\ref{eq:raw_convergence_non_private_shuffled_g} follows directly from the general result in Theorem~\ref{thm:convergence_generalized_shufflg_fm_appendix} (Eq.~\ref{eq:general_convergence_appendix}) by setting: the number of private samples per epoch $n_d^{(k)} = n$, dissimilarity $C_n^{(k)} = 0$, noise $\sigma^{(k)} = 0$, and using the appropriate smoothness constants—maximum $L^{*} = \max_{i\in[n]} L_i$ and average per epoch $\widehat{L}^{(k)} = \frac{1}{n} \sum_{i=1}^{n} L_i = L$ for all $k \in [K]$. 
    With these settings, Eq.~\ref{eq:raw_convergence_non_private_shuffled_g} recovers the convergence bound for non-private shuffled gradient methods under smooth objectives prior to specifying the learning rate $\eta$, as shown in Theorem C.1 of~\cite{liu2024last_iterate_shuffled_gradient}. Adopting the same learning rate choice as in~\cite{liu2024last_iterate_shuffled_gradient} yields the exact result stated in their Theorem 4.4 and full version Theorem C.1.
\end{proof}

\section{Private Shuffled Gradient Methods}
\label{sec:appendix_private_shuffled_g}

\subsection{Convergence Analysis}

We give the full convergence bound of $\DPSG$ as follows.
Since the full $\gD$ is used across all epochs, $n_d^{(k)} = n$ and the dissimilarity measure (see Assumption~\ref{ass:dissim_partial_lipschitzness}) is $C_n^{(k)} = 0$, $\forall k\in [K]$. 
By Theorem~\ref{thm:convergence_generalized_shufflg_fm}, the convergence of $\DPSG$ is

\begin{corollary}[Convergence of $\DPSG$\footnote{One can set $\beta = 0$ when $L_H^{(k)} = 0$, which is the case here since no surrogate datasets is used. This implies $\mu_{\psi} \geq 0$, as indicated in Assumption~\ref{ass:reg}, suffices to ensure convergence.} (Re-statement of Corollary~\ref{corollary:conv_dpsg})]
\label{corollary:appendix_conv_dpsg}
    Under the conditions in Theorem~\ref{thm:convergence_generalized_shufflg_fm}, with $n_d^{(k)} = n$, $\gP^{(k)} = \emptyset$, and 
    $\sigma^{(k)} = \sigma$ for all $k \in [K]$,
    Algorithm~\ref{alg:generalized_shuffled_gradient_fm} ($\DPSG$) guarantees
    \begin{align*}
        &\E\left[ G(\rvx_1^{(K+1)}) \right] - G(\rvx^*) \leq \underbrace{
        \frac{\| \rvx_1^{(1)} - \rvx^*\|^2}{\eta n K}
        }_{\text{Initialization}}
        + \underbrace{ 10 \eta^2 n^2 \sigma_{any}^2 L (1+\log K)
        }_{\text{Optimization Uncertainty}}
        + \underbrace{ 12 \eta^2 nd \sigma^2 L^{*} (1+\log K) }_{\text{Noise Injection}}
    \end{align*}
\end{corollary}

\subsection{Privacy Analysis}
\label{subsec:appendix_private_shuffled_g_privacy}

\begin{lemma}[Privacy of $\DPSG$ (Restatement of Lemma~\ref{lemma:privacy_private_shuffled_g})]
\label{lemma:appendix_privacy_private_shuffled_g}
    Under Assumptions
    ~\ref{ass:convexity},~\ref{ass:appendix_refined_smoothness} and~\ref{ass:lipschitzness}, if the learning rate is $\eta \leq 1/L^{*}$, $\DPSG$ is $(\frac{2\alpha G^2 K}{\sigma^2} + \frac{\log 1/\delta}{\alpha - 1}, \delta)$-DP, for $\alpha > 1, \delta \in (0, 1)$.
\end{lemma}

\begin{proof}[Proof of Lemma~\ref{lemma:appendix_privacy_private_shuffled_g}]
    We first show the privacy loss per epoch by using privacy amplification by iteration (PABI) in Renyi Differential Privacy (RDP, see Definition~\ref{def:RDP}), and then use the composition theorem of RDP (see Proposition~\ref{prop:rdp_composition}) to derive the total privacy loss across $K$ epochs. Finally, we convert the privacy loss in RDP to DP based on Proposition~\ref{prop:rdp_to_dp}. 

    By Remark~\ref{remark:contractive_operators}, if one sets the learning rate in $\DPSG$ as $\eta \leq 1/L^{*}$, each gradient update step in epoch $k \in [K]$, i.e., $\rvx_{i+1}^{(k)} = \rvx_{i}^{(k)} - \eta (\nabla f_{\pi_i^{(k)}}(\rvx_i^{(k)}) + \rho_i^{(k)} )$, $\forall i\in [n]$, (line 8 of Algorithm~\ref{alg:generalized_shuffled_gradient_fm}) satisfies a ``noisy contractive sequence'' (CNI, see Definition~\ref{def:cni}) and hence, we can apply Theorem~\ref{thm:pabi} to reason about the privacy loss of epoch $k$, $\forall k\in[K]$.

    Let $\gD = \{\rvd_1, \dots, \rvd_t, \dots, \rvd_n\}$ and $\gD' = \{\rvd_1, \dots, \rvd_{t}', \dots, \rvd_n\}$ be two neighboring datasets that differ at some index $t \in [n]$. 
    On dataset $\gD$, the CNI is defined by the initial point $\rvx_1^{(k)}$, sequence of functions $g_i(\rvx) = \rvx_i^{(k)} - \eta \nabla f(\rvx; \rvd_{\pi_i^{(k)}})$, for all $\rvx$, and sequence of noise distributions $\gN(0, (\eta \sigma)^2 \sI_d)$. Similarly, on dataset $\gD'$, the CNI is defined 
    in the same way with the exception of $g_j'(\rvx) = \rvx - \eta \nabla f(\rvx; \rvd_{\pi_j^{(k)}}')$ for the index $j$ such that $\pi_j^{(k)} = t$. 
    Let $\rvx_{n+1}^{(k)}$, $(\rvx_{n+1}^{(k)})'$ be the output of the CNI on dataset $\gD$ and $\gD'$, respectively.
    
    By Assumption~\ref{ass:lipschitzness}, $f(\rvx; \rvd_i)$ is $G$-Lipschitz, $\forall i\in [n]$, and hence,
    \begin{align*}
        \sup_{\rvw}\| g_j(\rvx) - g_j'(\rvx) \| = \sup_{\rvw}\| \eta \nabla f(\rvx; \rvd_{t}) - \eta \nabla f(\rvx; \rvd_{t}') \| \leq 2 \eta G    
    \end{align*}

    We now apply Theorem~\ref{thm:pabi} with $a_1, \dots, a_{j-1} = 0$ and $a_{j}, \dots, a_n = \frac{2\eta G}{n - j + 1}$. Note that $s_{\pi_j^{(k)}} = 2 \eta G$ and $s_{i} = 0$, $\forall i \neq \pi_j^{(k)}$. In addition, $z_i \geq 0$ for all $i \leq n$ and $z_n = 0$. Hence,
    \begin{align*}
        \infdivalpha{\rvx_{n+1}^{(k)}}{(\rvx_{n+1}^{(k)})'} \leq \sum_{i=j}^{n} \frac{\alpha}{2 \eta^2 \sigma^2} \cdot \frac{4 \eta^2 G^2}{(n - j + 1)^2}
        = \frac{2 \alpha G^2}{\sigma^2 (n-j + 1)}
    \end{align*}
    The maximum privacy loss happens when $j = n$, that is, when the sample $\rvd_{t}$ is the last one being processed in epoch $k$. 
    And it is not hard to see that $\max_{j\in [n]} \infdivalpha{\rvx_{n+1}^{(k)}}{(\rvx_{n+1}^{(k)})'} \leq \frac{2\alpha G^2}{\sigma^2}$,
    which implies $\rvx_{n+1}^{(k)}$ in Algorithm~\ref{alg:generalized_shuffled_gradient_fm} is $(\alpha, \frac{2\alpha G^2}{\sigma^2})$-RDP, for $\alpha > 1$.
    The output of epoch $k$, $\rvx_1^{(k+1)}$ can be seen as a post-processing step of $\rvx_{n+1}^{(k)}$, and hence, $\rvx_1^{(k+1)}$ is also $(\alpha, \frac{2\alpha G^2}{\sigma^2})$-RDP.
    
    By Proposition~\ref{prop:rdp_composition}, the output $\rvx_1^{(K+1)}$ is $(\alpha, \frac{2\alpha G^2 K}{\sigma^2})$-RDP. And by Proposition~\ref{prop:rdp_to_dp}, $\rvx_1^{(K+1)}$ is $(\frac{2\alpha G^2 K}{\sigma^2} + \frac{\log 1/\delta}{\alpha - 1}, \delta)$-DP, for $\alpha > 1$ and $\delta \in (0, 1)$.
    
\end{proof}

\textbf{Remark on the Privacy Loss of Random Reshuffling (RR).}
For Incremental Gradient Methods (IG) and Shuffle Once (SO), it is not hard to see that the worst-case privacy loss bound presented above is tight. One might argue that, in the case of Random Reshuffling (RR), where the permutation $\pi_i^{(k)}$ is re-generated at the beginning of each epoch, leading to a reshuffling of the sample order in $\gD$, this additional randomness could amplify privacy, thereby reducing the privacy loss. However, we argue that this potential improvement is limited to a constant level. 
Deriving a significantly smaller privacy loss bound in the RR setting -- such as one that scales proportionally to $1/n$ -- is unlikely without additional assumptions.

We use the following lemma to derive a tighter bound on the privacy loss of RR per epoch, taking into account the randomness introduced by shuffling:

\begin{lemma}[Joint convexity of scaled exponentiation of Rényi divergence, Lemma 4.1 of~\cite{ye2022singapore_paper}]
\label{lemma:joint_convexity_renyi_div}
    Let $\mu_1, \dots, \mu_m$ and $\nu_1, \dots, \nu_m$ be distributions over $\R^d$. Then, for any RDP order $\alpha \geq 1$, and any coefficients $p_1, \dots, p_m \geq 0$ that satisfy $p_1 + \dots + p_m = 1$, the following inequality holds
    \begin{align*}
        e^{(\alpha-1) \infdivalpha{\sum_{j=1}^{m} p_j \mu_j}{\sum_{j=1}^{m} p_j \nu_j} }
        \leq \sum_{j=1}^{m} p_j \cdot e^{(\alpha-1) \infdivalpha{\mu_j}{\nu_j} }
    \end{align*}
\end{lemma}

From the previous proof and the PABI bound (Theorem~\ref{thm:pabi}), we observe that the privacy loss for a single epoch is primarily determined by the index $j$ such that $\pi_j^{(k)} = t$, where $t$ is the index of the sample at which the two neighboring datasets
$\gD$ and $\gD'$ differ ($\rvd_t \in \gD$ and $\rvd_{t}' \in \gD'$).
Since shuffling ensures that $j$ can take any value in $\{1,2,\dots, n\}$ with equal probability, $j$ is a random variable uniformly distributed over $[n]$.

We apply Lemma~\ref{lemma:joint_convexity_renyi_div} by instantiating the distributions $\mu_i$ as the CNI's on $\gD$ with $j = i$ for value $i\in [n]$ and similarly, the distributions $\nu_i$ as the CNI's on $\gD'$ with $j = i$ for value $i\in [n]$. It easy to see that $p_i = \frac{1}{n}$. Hence, privacy loss of RR in epoch $k$, $\eps_{\text{per-epoch}}^{(k)}$, is given by
\begin{align*}
    \eps_{\text{per-epoch}}^{(k)} &= \frac{1}{\alpha-1} \log e^{(\alpha-1) \cdot \infdivalpha{\rvx_{n+1}^{(k)}}{\rvx_{n+1}^{(k)}}}\\
    &\leq \frac{1}{\alpha-1} \log \left(
        \frac{1}{n}\sum_{j=1}^{n} e^{(\alpha-1)\cdot \frac{2\alpha G^2}{\sigma^2 (n-j+1)}}
    \right)
    = \frac{1}{\alpha -1}\log \left(
        \underbrace{ \frac{1}{n}\sum_{j=1}^{n} e^{(\alpha-1) \cdot \frac{2\alpha G^2}{\sigma^2\cdot j}}
        }_{:=S_n}
    \right)
\end{align*}

In shuffled gradient methods, the dataset size $n$ is finite and usually small to allow for $K \geq 2$ epochs over the dataset to ensure convergence. Therefore, we cannot asymptotically approximate the bound by treating $n\rightarrow \infty$. 
When $n$ is small, the term $S_n$ in bound is dominated by $e^{(\alpha-1) \frac{2\alpha G^2}{\sigma^2}}$ and $\frac{1}{n}$,
leading to the approximation
$S_n \approx \frac{1}{n}e^{(\alpha-1) \frac{2\alpha G^2}{\sigma^2}}$.
Consequently, the upper bound on $\eps_{\text{per-epoch}}^{(k)}$ becomes $\frac{1}{\alpha - 1}\log S_n \approx \frac{2\alpha G^2}{\sigma^2} - \log n$.
This indicates the privacy loss bound for random reshuffling (RR) is nearly identical to that of IG and SO. As a result, the shuffling operation provides only a marginal improvement in privacy loss in this case.

Similar privacy loss bounds occur in the PABI-based privacy analysis of (impractical) variants of SGD and one can of course apply strong assumptions on $\infdivalpha{\rvx_{n+1}^{(k)}}{(\rvx_{n+1}^{(k)})'}$ to reduce the above upper bound. See, for example, Lemma 25 of the seminal work on PABI~\cite{Feldman2018privacy_amp_iter}.

\subsection{Computing the Empirical Excess Risk}
\label{subsec:appendix_private_shuffle_g_empirical_excess_risk}

To ensure the output $\rvx_1^{(K+1)}$ is $(\eps, \delta)$-DP, we set $\alpha = \frac{\sigma \sqrt{\log(1/\delta)}}{G \sqrt{2K}}$ based on Lemma~\ref{lemma:appendix_privacy_private_shuffled_g} to minimize the overall privacy loss. This choice implies $\sigma = \gO\left(\frac{G \sqrt{K \log(1/\delta)}}{\eps}\right)$.
The learning rate is set to be $\eta = \gO\left(\frac{\|\rvx_1^{(1)} - \rvx^{*} \|^{2/3}}{n L^{*} \left(K (1 + \log K)\right)^{1/3}}\right)$ to optimize the convergence bound, while satisfying the conditions of both Corollary~\ref{corollary:appendix_conv_dpsg} (convergence) and Lemma~\ref{lemma:appendix_privacy_private_shuffled_g} (privacy). After choosing the learning rate, the convergence bound of $\DPSG$ is now given by
\begin{align*}
    &\E\left[G(\rvx_1^{(K+1)})\right] - \E\left[G(\rvx^*)\right]\\
    &\leq \gO\Big(\|\rvx_1^{(1)} - \rvx^*\|^{4/3} (1+\log K)^{1/3} 
    \Big( \frac{L^*}{K^{2/3}} + \frac{1}{L^* K^{2/3}} + \frac{d G^2 K^{1/3}\log 1/\delta}{n L^* \eps^2} \Big) \Big)
\end{align*}
Finally, setting the number of epochs as
$K = \gO\left(\frac{n\eps^2}{d} \right)$ to minimize the above bound, resulting in
the following empirical excess risk:
\begin{align*}
    \E\left[G(\rvx_1^{(K+1)})\right] - \E\left[G(\rvx^*)\right]
    = \widetilde{O}\left( \frac{1}{n^{2/3}}(\frac{\sqrt{d}}{\eps})^{4/3} \right)
\end{align*}
with respect to $n$, $d$, and $\eps$, while ignoring other terms. 
Here, $\widetilde{\gO}$ hides logarithmic factors in $(n, d, 1/\delta)$.

\section{Private Shuffled Gradient Methods with Public Data}
\label{sec:appendix_algo_pub_data}

In this section, we give more details on algorithms that leverage public data samples to improve the privacy-convergence trade-offs.

\textbf{Setting.} 
In addition to the private dataset $\gD$, which defines the target objective (Eq.~\ref{eq:main_problem}), we assume access to a single public dataset $\gP = \{\rvp_1, \dots, \rvp_n\}$. Recall that for each sample $\rvp_i$, we denote the corresponding objective function as $f(\rvx; \rvp_i)$, which is $\widetilde{L}_i$-smooth and $\widetilde{G}$-Lipschitz. We define the maximum and average smoothness parameters over the public dataset as $\widetilde{L}^{*} = \max_{i \in [n]} \widetilde{L}_i$ and $\widetilde{L} = \frac{1}{n} \sum_{i=1}^{n} \widetilde{L}_i$, respectively. The maximum Lipschitz constant is denoted by $G^{*} = \max\{G, \widetilde{G}\}$.

\subsection{Parameters and Convergence Analysis}
\label{subsec:appendix_algo_pub_data_param_convergence}

Each algorithm that leverages public data samples considered in this section --- $\privpub$ (see Algorithm~\ref{alg:privpub}), $\pubpriv$ (see Algorithm~\ref{alg:pubpriv}) and $\interleaved$ (see Algorithm~\ref{alg:interleaved}) --- is an instantiation of the generalized shuffled gradient framework (Algorithm~\ref{alg:generalized_shuffled_gradient_fm}) with specific parameters.
We summarize the key parameters of each algorithm in Table~\ref{tab:algo_pub_samples}.

The specific parameter choices for each algorithm result in different dissimilarity measures and the maximum smoothness parameter, both of which are critical factors in the convergence bound. Their values are summarized in Table~\ref{tab:dissim_max_smoothness}.

\begin{algorithm}[h]
\caption{$\privpub$}
\label{alg:privpub} 
    \begin{algorithmic}[1]
    \STATE Input: 
    Initial point $\rvx_1^{(1)}$, learning rate $\eta$, number of epochs $K$.
    Private dataset $\gD = \{\rvd_i\}_{i=1}^{n}$. Public datasets $\gP= \{\rvp_i\}_{i=1}^{n}$.
    Number of epochs $S$ using private samples only.
    Noise standard deviation $\sigma_{\text{prp}}$.
    \STATE \textit{IG}: Fix an order $\pi$, and set $\pi^{(k)} = \pi, \forall k \in [K]$
    \STATE \textit{SO}: Generate permutation $\pi$ of $[n]$, and set $\pi^{(k)} = \pi, \forall k\in [K]$
    \FOR{Private epochs $k = 1,2,\dots,S$}
        \STATE \textit{RR}: Generate permutation $\pi^{(k)}$ of $[n]$
        \FOR{$i = 1,2,\dots, n$}
            \STATE $\rvx_{i+1}^{(k)} \leftarrow \rvx_i^{(k)} - \eta \Big(\nabla f(\rvx_i^{(k)};  \rvd_{\pi_i^{(k)}} ) + \rho_i^{(k)} \Big)$, where noise $\rho_i^{(k)} \sim \gN(0, (\sigma_{\text{prp}})^2 \sI_d)$
        \ENDFOR
        \STATE $\rvx_1^{(k+1)} \leftarrow \argmin_{\rvx\in \R^{d}} n  \psi(\rvx) + \frac{\|\rvx - \rvx_{n+1}^{(k)}\|^2}{2\eta}$ 
    \ENDFOR
    \FOR{Public epochs $k = S+1, \dots, K$}
        \FOR{$i = 1,2, \dots, n$ }
            \STATE $\rvx_{i+1}^{(k)} \leftarrow \rvx_i^{(k)} - \eta \nabla f(\rvx_i^{(k)}; \rvp_{i})$
        \ENDFOR
        \STATE $\rvx_1^{(k+1)} \leftarrow \argmin_{\rvx\in \R^{d}} n  \psi(\rvx) + \frac{\|\rvx - \rvx_{n+1}^{(k)}\|^2}{2\eta}$ 
    \ENDFOR
    \STATE \textbf{return} $\rvx_1^{(K+1)}$
    \end{algorithmic}
\end{algorithm}

\begin{algorithm}[h]
\caption{$\pubpriv$}
\label{alg:pubpriv} 
    \begin{algorithmic}[1]
    \STATE Input: 
    Initial point $\rvx_1^{(1)}$, learning rate $\eta$, number of epochs $K$.
    Private dataset $\gD = \{\rvd_i\}_{i=1}^{n}$. Public datasets $\gP = \{\rvp_i\}_{i=1}^{n}$.
    Number of epochs $S$ using public samples only.
    Noise standard deviation $\sigma_{\text{pup}}$.
    \STATE \textit{IG}: Fix an order $\pi$, and set $\pi^{(k)} = \pi, \forall k \in [K]$
    \STATE \textit{SO}: Generate permutation $\pi$ of $[n]$, and set $\pi^{(k)} = \pi, \forall k\in [K]$
    \FOR{Public epochs $k = 1, 2, \dots, S$}
        \FOR{$i = 1,2, \dots, n$ }
            \STATE $\rvx_{i+1}^{(k)} \leftarrow \rvx_i^{(k)} - \eta \nabla f(\rvx_i^{(k)}; \rvp_{i})$
        \ENDFOR
        \STATE $\rvx_1^{(k+1)} \leftarrow \argmin_{\rvx\in \R^{d}} n  \psi(\rvx) + \frac{\|\rvx - \rvx_{n+1}^{(k)}\|^2}{2\eta}$ 
    \ENDFOR
    \FOR{Private epochs $k = S+1,\dots, K$}
        \STATE \textit{RR}: Generate permutation $\pi^{(k)}$ of $[n]$
        \FOR{$i = 1,2,\dots, n$}
            \STATE $\rvx_{i+1}^{(k)} \leftarrow \rvx_i^{(k)} - \eta \Big(\nabla f(\rvx_i^{(k)};  \rvd_{\pi_i^{(k)}} ) + \rho_i^{(k)} \Big)$, where noise $\rho_i^{(k)} \sim \gN(0, (\sigma_{\text{pup}})^2 \sI_d)$
        \ENDFOR
        \STATE $\rvx_1^{(k+1)} \leftarrow \argmin_{\rvx\in \R^{d}} n  \psi(\rvx) + \frac{\|\rvx - \rvx_{n+1}^{(k)}\|^2}{2\eta}$ 
    \ENDFOR
    \STATE \textbf{return} $\rvx_1^{(K+1)}$
    \end{algorithmic}
\end{algorithm}

\begin{algorithm}[h]
\caption{$\interleaved$}
\label{alg:interleaved} 
    \begin{algorithmic}[1]
    \STATE Input: 
    Initial point $\rvx_1^{(1)}$, learning rate $\eta$, number of epochs $K$.
    Private dataset $\gD = \{\rvd_i\}_{i=1}^{n}$. Number of private samples to use per epoch $n_d \in (0, n)$. Public datasets $\gP = \{\rvp_i\}_{i=1}^{n-n_d}$.
    Noise standard deviation $\sigma_{\text{int}}$.
    \STATE \textit{IG}: Fix an order $\pi$, and set $\pi^{(k)} = \pi, \forall k \in [K]$
    \STATE \textit{SO}: Generate permutation $\pi$ of $[n]$, and set $\pi^{(k)} = \pi, \forall k\in [K]$
    \FOR{$k = 1,2,\dots,K$}
        \STATE \textit{RR}: Generate permutation $\pi^{(k)}$ of $[n]$
        \FOR{$i = 1,2,\dots, n_d$}
            \STATE $\rvx_{i+1}^{(k)} \leftarrow \rvx_i^{(k)} - \eta \Big(\nabla f(\rvx_i^{(k)};  \rvd_{\pi_i^{(k)}} ) + \rho_i^{(k)} \Big)$, where noise $\rho_i^{(k)} \sim \gN(0, (\sigma_{\text{int}})^2 \sI_d)$
        \ENDFOR
        \FOR{$i = n_d+1, n_d+ 2, \dots, n$ }
            \STATE $\rvx_{i+1}^{(k)} \leftarrow \rvx_i^{(k)} - \eta \Big( \nabla f(\rvx_i^{(k)}; \rvp_{i-n_d}) + \rho_i^{(k)}\Big)$, where noise $\rho_i^{(k)} \sim \gN(0, (\sigma_{\text{int}})^2\sI_d)$ 
        \ENDFOR
        \STATE $\rvx_1^{(k+1)} \leftarrow \argmin_{\rvx\in \R^{d}} n  \psi(\rvx) + \frac{\|\rvx - \rvx_{n+1}^{(k)}\|^2}{2\eta}$ 
    \ENDFOR
    \STATE \textbf{return} $\rvx_1^{(K+1)}$
    \end{algorithmic}
\end{algorithm}

\textbf{Convergence.} We now present the convergence of each algorithm, as corollaries of Theorem~\ref{thm:convergence_generalized_shufflg_fm}, in Corollary~\ref{corollary:convergence_pub_data_opt}.
Note that to ensure the following bounds are tight, we enforce that the number of pre-determined epochs to be $S \in \{1,2,\dots,K-1\}$.

\begin{table}[]
    \centering
\begin{threeparttable}
\begin{adjustbox}{width=1\textwidth}
    \begin{tabular}{|c|l|l|l|}
    \hline
         \diagbox{Parameters}{Algorithm} & $\privpub$ & $\pubpriv$ & $\interleaved$\\
    \hline
        \# private samples used: $n_d^{(k)}$
        & $=\begin{cases}
            n & \text{if $k \leq S$\tnote{\S}}\\
            0 & \text{if $S < k \leq K$}
        \end{cases}$
        & $=\begin{cases}
            0 & \text{if $k \leq S$}\\
            n & \text{if $S < k \leq K$}
        \end{cases}$
        & $n_d^{(k)} = n_d$\tnote{\dag}, $\forall k\in [K]$\\
    \hline
        Noise variance: $(\sigma^{(k)})^2$ 
         & $= \begin{cases}
             (\sigma_{\text{prp}})^2 & \text{if $k \leq S$}\\
             0 & \text{if $S < k \leq K$}
         \end{cases}$
         & $= \begin{cases}
             0 & \text{if $k \leq S$}\\
             (\sigma_{\text{pup}})^2 & \text{if $S < k \leq K$}
         \end{cases}$
         & $= (\sigma_{\text{int}})^2, \forall k\in [K]$ \\
    \hline
    \end{tabular}
    \end{adjustbox}
    \begin{tablenotes}
        \item[\S] $S \in \{1,2,\dots, K-1\}$ is a pre-determined number of epochs.
        \item[\dag] $n_d\in [n]$ is a pre-determined number of private samples to use in every epoch.
    \end{tablenotes}
\end{threeparttable}
    \caption{Parameters of different algorithms that leverage public data samples.}
    \label{tab:algo_pub_samples}
\end{table}

\begin{table*}[]
    \centering
\begin{threeparttable}
\begin{adjustbox}{width=1\textwidth}
    \begin{tabular}{|c|l|l|l|}
    \hline
        \diagbox{Terms}{Algorithm} & $\privpub$ & $\pubpriv$ & $\interleaved$\\
    \hline
         Dissimilarity (Non-vanishing): $C_n^{(k)}$
         & $=\begin{cases}
             0 & \text{if $k \leq S$}\\
             C_n^{\text{full}} & \text{if $S < k \leq K$}
         \end{cases}$ 
         & $=\begin{cases}
             C_n^{\text{full}} & \text{if $k \leq S$}\\
             0 & \text{if $k < S \leq K$}
         \end{cases}$ 
         & $= C_n^{\text{part}}$, $\forall k \in [K]$ \\
    \hline
         Dissimilarity: $\frac{1}{n}\sum_{i=1}^{n-1} (C_i^{(k)})^2 $
         & $= \begin{cases}
             0 & \text{if $k \leq S$}\\
             \frac{1}{n}\sum_{i=1}^{n-1}(C_i^{\text{full}})^2  & \text{if $S < k \leq K$}
         \end{cases}$
         & $= \begin{cases}
            \frac{1}{n}\sum_{i=1}^{n-1}(C_i^{\text{full}})^2 & \text{if $k \leq S$}\\
             0 & \text{if $S < k \leq K$}
         \end{cases}$
         & $= \frac{1}{n}\sum_{i= n_d+1}^{n-1} (C_{i}^{\text{part}})^2$, $\forall k\in [K]$ \\
    \hline
         Max smoothness parameter $\widehat{L}^{(k)*}$
         & $= \begin{cases}
            L^* & \text{if $k \leq S$}\\
            \widetilde{L}^{*} & \text{if $S < k \leq K$}
         \end{cases}$
         & $= \begin{cases}
             \widetilde{L}^{*} & \text{if $k \leq S$}\\
             L^* & \text{if $S < k \leq K$}
         \end{cases}$
         & $= \max\{L^*, \widetilde{L}^*\}$ \\
    \hline
    \end{tabular}
\end{adjustbox}
\end{threeparttable}
    \caption{The resulting dissimilarity measures and the maximum smoothness parameters of different algorithms. Here, $C_n^{\text{full}}$ measures the dissimilarity between $\gD$ and $\gP$ over the full datasets. $C_n^{\text{part}}$ measures the dissimilarity between $\gD$ and using the first $n-n_d$ samples from $\gP$. 
    This notion similarly extends to $C_i^{\text{part}}$ and $C_i^{\text{full}}$, for $i< n$.
    }
    \label{tab:dissim_max_smoothness}
\end{table*}

\begin{corollary}[Convergence of Public Data Assisted Optimization]
\label{corollary:convergence_pub_data_opt}
    If one instantiates Algorithm~\ref{alg:generalized_shuffled_gradient_fm} with parameters indicated in Table~\ref{tab:algo_pub_samples}, then
    under Assumptions~\ref{ass:convexity},~\ref{ass:appendix_refined_smoothness},~\ref{ass:reg},~\ref{ass:H_smoothness} and~\ref{ass:dissim_partial_lipschitzness}, 
    for $\beta > 0$, 
    if $\mu_{\psi} \geq L_H^{(k)} + \beta$, $\forall k\in [K]$, and 
     $\eta \leq \frac{1}{2n \max\{L^*, \widetilde{L}^{*}\} \sqrt{10 (1+\log K)}}$,     Algorithm~\ref{alg:generalized_shuffled_gradient_fm}
    guarantees convergence 
    \begin{align*}
        &\E\left[ G(\rvx_1^{(K+1)}) \right] - \E\left[G(\rvx^*)\right]
        \leq \underbrace{ \frac{\|\rvx_1^{(1)} - \rvx^*\|^2}{\eta n K}  }_{\text{Initialization}}
        + \underbrace{ 10 \eta^2 n^2 \sigma_{any}^2 (1+\log K)\max\{L, \widetilde{L}\} }_{\text{Optimization Uncertainty}} + 2 M
    \end{align*}
    where
    \begin{itemize}[itemsep=0mm]
        \item For \textbf{$\privpub$},
            \begin{align*}
                M &= \underbrace{
                    \frac{1 + \log (K-S)}{2 n^2 \beta} (C_n^{\text{full}})^2
                }_{\text{Non-vanishing Dissimilarity}}
                + \underbrace{
                    5\eta^2 (1+ \log (K - S)) \frac{1}{n}\sum_{i=1}^{n-1}(C_i^{\text{full}})^2\widetilde{L}^{*}
                }_{\text{Vanishing Dissimilarity}}\\
                \nonumber
                &\quad + \underbrace{
                    6 \eta^2 nd (1 + \log S) (\sigma_{\text{prp}})^2 L^{*}
                }_{\text{Injected Noise}}
            \end{align*}

        \item For \textbf{$\pubpriv$},
            \begin{align*}
                M &= \underbrace{
                    \frac{1 + \log S}{2n^2 \beta} (C_n^{\text{full}})^2
                }_{\text{Non-vanishing Dissimilarity}}
                + \underbrace{
                    5 \eta^2 (1+\log S) \frac{1}{n}\sum_{i=1}^{n-1} (C_i^{\text{full}})^2 \widetilde{L}^{*}
                }_{\text{Vanishing Dissimilarity}}\\
                \nonumber
                &\quad + \underbrace{
                    6 \eta^2 nd (1 + \log (K-S)) (\sigma_{\text{pup}})^2 L^{*}
                }_{\text{Injected Noise}}
            \end{align*}

        \item For \textbf{$\interleaved$},
            \begin{align*}
                M &= \underbrace{
                    \frac{1 + \log K}{2n^2 \beta} (C_n^{\text{part}})^2
                }_{\text{Non-vanishing Dissimilarity}}
                + \underbrace{
                    5 \eta^2 (1 + \log K) \frac{1}{n}\sum_{i=n_d+1}^{n-1} (C_{i}^{\text{part}})^2 \max\{L^*, \widehat{L}^*\}
                }_{\text{Vanishing Dissimilarity}}\\
                \nonumber
                &\quad + \underbrace{
                    6 \eta^2 nd (1+\log K) (\sigma_{\text{int}})^2 \max\{ L^{*}, \widehat{L}^{*}\}
                }_{\text{Injected Noise}}
            \end{align*}
    \end{itemize}

\end{corollary}

\subsection{Privacy Analysis}
\label{subsec:appendix_algo_pub_data_privacy}

In this section, we derive the privacy guarantees of the three algorithms that leverage public samples: $\privpub$, $\pubpriv$ and $\interleaved$.

\begin{lemma}[Privacy of Public Data Assisted Optimization]
\label{lemma:appendix_privacy_pub_data_opt}
    Under Assumptions~~\ref{ass:convexity},~\ref{ass:appendix_refined_smoothness} and~\ref{ass:lipschitzness},
    if the learning rate is $\eta \leq 1/L^{*}$,
    \begin{itemize}[itemsep=0mm]
        \item \textit{$\privpub$} is $(\frac{2 \alpha G^2 S}{(\sigma_{\text{prp}})^2} + \frac{\log 1/\delta}{\alpha-1}, \delta)$-DP.
        \item \textit{$\pubpriv$} is $(\frac{2\alpha G^2 (K-S)}{(\sigma_{\text{pup}})^2} + \frac{\log 1/\delta}{\alpha-1}, \delta)$-DP.
    \end{itemize}
    and if the learning rate is $\eta \leq 1/\max\{L^{*}, \widetilde{L}^{*}\}$,
    \begin{itemize}[itemsep=0mm]
        \item \textit{$\interleaved$} is $(\frac{2\alpha (G^{*})^2 K}{(n+1 - n_d) (\sigma_{\text{int}})^2} + \frac{\log 1/\delta}{\alpha-1}, \delta)$-DP.
    \end{itemize}
\end{lemma}

\begin{proof}[Proof of Lemma~\ref{lemma:appendix_privacy_pub_data_opt}]
    The proof is similar to the proof of Lemma~\ref{lemma:appendix_privacy_private_shuffled_g} showing the privacy loss of $\DPSG$. If the learning rate is set as $\eta \leq 1/L^*$ for $\privpub$ and $\pubpriv$ and $\eta \leq 1/\max\{L, L^{*}\}$ for $\interleaved$, it is then guaranteed that each gradient step in one epoch is ``contractive'', which enables us to apply PABI (Theorem~\ref{thm:pabi}) to reason about the per-epoch privacy loss. 

    The per-epoch privacy loss of $\privpub$ and $\pubpriv$ is the same as the per-epoch privacy loss in $\DPSG$ whenever the private dataset $\gD$ is used during the epoch. Specifically, for $\alpha > 1$,
    \begin{itemize}[itemsep=0mm]
        \item For $\privpub$, $\rvx_{1}^{(k+1)}$ is $(\alpha, \frac{2\alpha G^2}{(\sigma_{\text{prp}})^2})$-RDP, if $k \leq S$ and there is no privacy loss 0 otherwise.
        \item For $\pubpriv$, $\rvx_{1}^{(k+1)}$ is $(\alpha, \frac{2\alpha G^2}{(\sigma_{\text{pup}})^2})$-RDP, if $S+1 \leq k \leq K$, and there is no privacy loss 0 otherwise.
    \end{itemize}

    By applying composition (Proposition~\ref{prop:rdp_composition}) across $K$ and $K-S$ epochs for $\privpub$ and $\pubpriv$, respectively, and subsequently converting the RDP bound to a DP bound using Proposition~\ref{prop:rdp_to_dp}, we obtain the overall privacy loss as stated in the lemma statement.

    For $\interleaved$, we consider two neighboring datasets $\gD = \{\rvd_1,\dots,\rvd_t,\dots, \rvd_n\}$ and $\gD' = \{\rvd_1, \dots, \rvd_t', \dots, \rvd_n\}$ that differ at some index $t \in [n]$. 
    In every epoch, only the first $n_d$ steps use samples from the private dataset $\gD$, and the remaining steps all use public samples. Hence, $t$ can only occur at step $j \leq n_d$ in the sequence of updates in every epoch. 
    
    We apply Theorem~\ref{thm:pabi} with $a_1, \dots, a_{j-1} = 0$ and $a_j,\dots, a_n = \frac{2\eta G^{*}}{n-j+1}$, where $j \leq n_d$. Note that $s_{\pi_j^{(k)}} = 2\eta G^{*}$ and $s_i = 0$, $\forall i\neq \pi_j^{(k)}$.  In addition, $z_i \geq 0$ for all $i\leq n$ and $z_n = 0$. Hence,
    \begin{align*}
        \infdivalpha{\rvx_{n+1}^{(k)}}{(\rvx_{n+1}^{(k)})'} \leq \sum_{i=j}^{n} \frac{\alpha}{2 \eta^2 (\sigma_{\text{int}})^2} \cdot \frac{4\eta^2 (G^{*})^2}{(n-j+1)^2}
        = \frac{2\alpha (G^{*})^2}{(\sigma_{\text{int}})^2(n-j+1)}
    \end{align*}
    where $(\rvx_{n+1}^{(k)})'$ is the point obtained by optimizing on the neighboring dataset $\gD'$.

    Since $j \leq n_d$, the maximum privacy loss happens at $j = n_d$, that is, when the sample $\rvd_t$ is used at step $n_d$ in one epoch. And so
    \begin{align*}
        \max \infdivalpha{\rvx_{n+1}^{(k)}}{(\rvx_{n+1}^{(k)})'}
        \leq \frac{2\alpha (G^{*})^2}{(\sigma_{\text{int}})^2 (n-n_d+1)}
    \end{align*}
    which implies $\rvx_{n+1}^{(k)}$ and hence, $\rvx_{1}^{(k+1)}$, is $(\alpha, \frac{2\alpha (G^{*})^2}{(\sigma_{\text{int}})^2 (n-n_d+1)} )$-RDP.

    Applying composition (Proposition~\ref{prop:rdp_composition}) across $K$ epochs and converting the RDP bound to a DP bound using Proposition~\ref{prop:rdp_to_dp} leads to the overall privacy loss as stated in the lemma statement.

\end{proof}

\subsection{Computing the Empirical Excess Risk}
\label{subsec:appendix_algo_pub_data_risk}

In this section, we derive the empirical excess risk of the three algorithms that leverage public samples: $\privpub$, $\pubpriv$ and $\interleaved$.

We begin by determining the optimal order $\alpha$ in in the RDP bound that minimizes the privacy loss, and the resulting amount of noise required for each algorithm to ensure the algorithm satisfies $(\eps, \delta)$-DP, as summarized in Table~\ref{tab:algo_pub_data_order_noise}.

\begin{table}[H]
    \centering
    \begin{tabular}{|c|c|c|}
    \hline
        \diagbox{Algorithm}{Term} & Renyi Order $\alpha$ & Noise Variance \\
    \hline
         \privpub & $\alpha = \frac{\sigma_{\text{prp}}\sqrt{\log 1/\delta}}{G \sqrt{2S}}$ 
         & $(\sigma_{\text{prp}})^2 = \gO\left( \frac{G^2 S \log 1/\delta}{\eps^2} \right)$ \\
    \hline
         \pubpriv & $\alpha = \frac{\sigma_{\text{pup}} \sqrt{\log 1/\delta}}{G \sqrt{2(K-S)}}$ 
         & $(\sigma_{\text{pup}})^2 = \gO\left( \frac{ G^2 (K-S) \log 1/\delta}{\eps^2} \right)$ \\
    \hline
         \interleaved & 
         $\alpha = \frac{\sigma_{\text{int}} \sqrt{(n+1-n_d) \log 1/\delta}}{G^* \sqrt{2K}}$ 
         & $(\sigma_{\text{int}})^2 = \gO\left( \frac{(G^{*})^2 K \log 1/\delta}{\eps^2 (n+1-n_d) } \right)$ \\
    \hline
    \end{tabular}
    \caption{Choices of the order $\alpha$ in the RDP bound (Lemma~\ref{lemma:appendix_privacy_pub_data_opt}) and the resulting amount of noise required for each algorithm to ensure the output $\rvx_1^{(K+1)}$ satisfies $(\eps,\delta)$-DP.}
    \label{tab:algo_pub_data_order_noise}
\end{table}

Based on Corollary~\ref{corollary:convergence_pub_data_opt},
we set the learning rate as $\eta = \gO\left( \frac{\|\rvx_1^{(1)} - \rvx^*\|^{2/3}}{n \max\{L^{*}, \widetilde{L}^{*}\} (K(1+\log K))^{1/3}} \right)$ in $\privpub$, $\pubpriv$ and $\interleaved$
to minimize the convergence bound, while satisfying the conditions of both Corollary~\ref{corollary:convergence_pub_data_opt} (convergence) and Lemma~\ref{lemma:appendix_privacy_pub_data_opt} (privacy). 
After choosing the learning rate, the convergence bounds of the algorithms are now given by
\begin{itemize}
    \item $\privpub$:
    \begin{align*}
        &\E\left[ G(\rvx_1^{(K+1)}) \right] - \E\left[G(\rvx^*)\right]\\
        &\leq \underbrace{ \gO \left( 
                \frac{\|\rvx_1^{(1)} - \rvx^*\|^{4/3} \max\{L^{*}, \widetilde{L}^{*}\} (1+\log K)^{1/3}}{K^{2/3}} \right)
            }_{\text{Initialization}}
        + \underbrace{ \gO \left( 
            \frac{\|\rvx_1^{(1)} - \rvx^*\|^{4/3} \sigma_{any}^2 (1+\log K)^{1/3}}
            {K^{2/3} \max\{L^{*}, \widetilde{L}^{*} \} } 
            \right)
        }_{\text{Optimization Uncertainty}}\\
        &\quad + \underbrace{
            \gO\left( 
            \frac{1 + \log (K-S) }{n^2 \beta} (C_n^{\text{full}})^2 \right)
        }_{\text{Non-vanishing Dissimilarity}}
            + \underbrace{
                \gO \left( \frac{\|\rvx_1^{(1)} - \rvx^*\|^{4/3} \frac{1}{n}\sum_{i=1}^{n}(C_i^{\text{full}})^2}
                {n^2 \max\{L^{*}, \widetilde{L}^{*}\} K^{2/3}}
                \cdot \frac{(1+ \log (K- S))}{(1+ \log K)^{1/3}} \right)
        }_{\text{Vanishing Dissimilarity}} \\
        &\quad + \underbrace{ \gO\left( 
            \frac{\|\rvx_1^{(1)} - \rvx^{*}\|^{4/3}d G^2 \log (1/\delta) S }
            {n \max\{L^{*}, \widetilde{L}^{*} \} \eps^2 K^{2/3}} \cdot \frac{(1+ \log S )}{(1 + \log K)^{2/3}}
        \right)
        }_{\text{Noise Injection}}
    \end{align*}

    \item $\pubpriv$:
    \begin{align*}
        &\E\left[ G(\rvx_1^{(K+1)}) \right] - \E\left[G(\rvx^*)\right]\\
        &\leq \underbrace{ \gO \left( 
                \frac{\|\rvx_1^{(1)} - \rvx^*\|^{4/3} \max\{L^{*}, \widetilde{L}^{*}\} (1+\log K)^{1/3}}{K^{2/3}} \right)
            }_{\text{Initialization}}
        + \underbrace{ \gO \left( 
            \frac{\|\rvx_1^{(1)} - \rvx^*\|^{4/3} \sigma_{any}^2 (1+\log K)^{1/3}}
            {K^{2/3} \max\{L^{*}, \widetilde{L}^{*} \} } 
            \right)
        }_{\text{Optimization Uncertainty}}\\
        &\quad + \underbrace{ \gO\left(
            \frac{1 + \log S}{n^2 \beta}(C_n^{\text{full}})^2 \right)
        }_{\text{Non-vanishing Dissimilarity}}
        + \underbrace{ 
            \gO\left( \frac{\|\rvx_1^{(1)} - \rvx^{*}\|^{4/3} \frac{1}{n}\sum_{i=1}^{n}(C_i^{\text{part}})^2}
            {n^2 \max\{L^{*}, \widetilde{L}^{*}\} K^{2/3} } \cdot \frac{1 + \log S}{(1+\log K)^{2/3}} \right)
        }_{\text{Vanishing Dissimilarity}} \\
        &\quad + \underbrace{ \gO\left( \frac{\|\rvx_1^{(1)} - \rvx^{*}\|^{4/3} d G^2  \log (1/\delta) (K-S)}{n \max\{L^{*}, \widetilde{L}^{*}\} \eps^2 K^{2/3}} \cdot \frac{1 + \log (K - S)}{(1+\log K)^{2/3}} \right)
        }_{\text{Noise Injection}}
    \end{align*}

    \item $\interleaved$:
    \begin{align*}
        &\E\left[ G(\rvx_1^{(K+1)}) \right] - \E\left[G(\rvx^*)\right]\\
        &\leq \underbrace{ \gO \left( 
                \frac{\|\rvx_1^{(1)} - \rvx^*\|^{4/3} \max\{L^{*}, \widetilde{L}^{*}\} (1+\log K)^{1/3}}{K^{2/3}} \right)
            }_{\text{Initialization}}
        + \underbrace{ \gO \left( 
            \frac{\|\rvx_1^{(1)} - \rvx^*\|^{4/3} \sigma_{any}^2 (1+\log K)^{1/3}}
            {K^{2/3} \max\{L^{*}, \widetilde{L}^{*} \} } 
            \right)
        }_{\text{Optimization Uncertainty}}\\
        &\quad + \underbrace{ \gO\left( \frac{1 + \log K}{n^2 \beta} (C_n^{\text{part}})^{2} \right) 
        }_{\text{Non-vanishing Dissimilarity}}
        + \underbrace{ \gO\left( \frac{\|\rvx_1^{(1)} -\rvx^{*}\|^{4/3} \frac{1}{n}\sum_{i=n_d+1}^{n-1}(C_{i}^{\text{part}})^2}{n^2 \max\{L^{*}, \widetilde{L}^{*} \} K^{2/3}  } (1+\log K)^{1/3} \right)
        }_{\text{Vanishing Dissimilarity}}\\
        &\quad + \underbrace{ \gO\left( \frac{\|\rvx_1^{(1)} - \rvx^{*}\|^{4/3} d (G^{*})^2 \log (1/\delta) K^{1/3}}
        {n \max\{L^{*}, \widetilde{L}^{*} \} \eps^2 (n-1+n_d)} (1+\log K)^{1/3} \right)
        }_{\text{Noise Injection}}
    \end{align*}
\end{itemize}

\textbf{Comparison.} Recall that to ensure a fair comparison, we fix the total number of gradient steps using private samples from $\gD$ and public samples from $\gP$ to be identical across $K$ epochs. We use $p\in (0,1]$ to denote the fraction of gradient steps computed using private samples. Specifically,
\begin{enumerate}[itemsep=0mm]
    \item For \pubpriv, we set the number of epochs using the private dataset $\gD$ as $S = p K$. Since $S \in [K]$, here, $p \in [\frac{1}{K}, 1]$.
    \item For \privpub, we set the number of epochs using private dataset $\gD$ as $K - S = p K$. Since $K - S \in [K]$, again, $p\in [\frac{1}{K}, 1]$.
    \item For \interleaved, we set the number of steps using samples from the private dataset $\gD$ within every epoch as $n_d = p n$.
\end{enumerate}

Based on the above, we consider the fraction of steps of using private samples as $p\in [\frac{1}{K}, 1]$.
For simplicity, we assume that both the number of epochs using the private dataset ($pK$) and the number of steps using private samples within a single epoch ($pn$) are integers.
We summarize the optimal number of epochs $K$ for each algorithm to minimize the convergence bound, along with the resulting empirical excess risk in Table~\ref{tab:empirical_excess_risk_algo_pub_data}. 
To keep the comparison clean, the empirical excess risk bounds are represented in $n,d,\eps, p$ and the non-vanishing dissimilarity term,
while treating other terms as constants.

\begin{table}[H]
    \centering
\begin{threeparttable}
    \begin{tabular}{|c|c|c|}
    \hline
         \diagbox{Algorithm}{Term} & \# Epochs $K$ & Empirical Excess Risk \\
    \hline
         \privpub & $\gO\left(\frac{n\eps^2}{d p} \right)$ 
         & $\widetilde{\gO}\left( 
         \left(\frac{p}{n}\right)^{2/3} \left(\frac{\sqrt{d}}{\eps} \right)^{4/3}
         + \frac{(C_n^{\text{full}})^2}{n^2 \beta}
         \right)$ \\
    \hline
        \pubpriv & $\gO\left( \frac{n\eps^2}{d p} \right)$
        & $\widetilde{\gO}\left( 
         \left( \frac{p}{n} \right)^{2/3} \left(\frac{\sqrt{d}}{\eps} \right)^{4/3}
         + \frac{(C_n^{\text{full}})^2}{n^2 \beta}
         \right)$ \\
    \hline
        \interleaved & $\gO\left( \frac{n\eps^2 [ (1-p)n+1 ]}{d} \right)$
        & $\widetilde{\gO}\left( 
        \left(\frac{1}{n[(1-p)n + 1]}\right)^{2/3}
        \left( \frac{\sqrt{d}}{\eps} \right)^{4/3}
        + \frac{(C_n^{\text{part}})^2}{n^2\beta}
        \right)$\\
    \hline
    \end{tabular}
\end{threeparttable}
    \caption{
    Empirical excess risk for each algorithm using public samples, represented in terms of $n,d,\eps,p$ and the non-vanishing dissimilarity term.
    Here, $p\in [\frac{1}{K},1]$ is the fraction of gradient steps that use private samples. 
    The dissimilarity measures $C_n^{\text{full}}$ and $C_n^{\text{part}}$ reflects the difference between the private dataset $\gD$ and the public dataset $\gP$, when using full and partial samples from $\gP$ in a single epoch, respectively. 
    The notation $\widetilde{\gO}$ suppresses logarithmic factors in $n, d, 1/\eps^2$ and $1/\delta$.}
    \label{tab:empirical_excess_risk_algo_pub_data}
\end{table}

\section{Using Other Regularization Functions $\psi$}
\label{sec:appendix_other_reg}

In this section, we consider using regularization functions $\psi$ that are not twice differentiable.
Specifically, we consider $\psi$ being the $\ell_1$ regularizer or the projection operator onto a convex set $\gB$.

The convergence proof for these cases remains the same as when $\psi$ is twice differentiable up to Section~\ref{subsec:appendix_one_epoch_convergence}, which analyzes one-epoch convergence prior to taking the expectation with respect to the injected noise.
We need to re-compute the expected additional error term introduced by noise injection (Lemma~\ref{lemma:noise_bias}), which was bounded directly through Stein's lemma when $\psi$ is twice differentiable. However, stein's lemma does not apply when the function of the noise, which involves $\psi$, exists points at which it is not twice differentiable.

\subsection{The $\ell_1$ Regularizer}
\label{subsec:appendix_l1_reg}

Instead of using Stein's lemma out of the shelf, we follow a similar idea and directly apply integration by parts. An additional offset term appears in this case, due to the non-differentiable points in the soft thresholding function after applying the $\ell_1$ regularization. To simplify the bound on the additional offset term, we make an assumption here:
\begin{assumption}
\label{ass:l1_offset}
    Let the regularization function in the objective (Eq.(\ref{eq:main_problem})) be $\psi(\rvx) = \lambda \|\rvx\|_1$ for some $\lambda > 0$.
    For all $k\in [K]$, there exists a constant $B > 0$ such that the model parameter $\rvx_{n+1}^{(k)}$ satisfies
    \begin{align*}
        \E\left[ \left\| \rvx_{n+1}^{(k)} - \eta \lambda n \mathbf{1} \right\|_{\infty}\right] \leq B
    \end{align*}
    where the expectation is taken w.r.t. the injected noise vectors, $\mathbf{1} \in \R^d$ is the all one vector, and recall that $\rvx_{n+1}^{(k)}$ is a function of the noise vectors. 
\end{assumption}

\begin{lemma}[Additional Error ($\ell_1$ Regularization)]
\label{lemma:noise_bias_l1_reg}
    For any epoch $k\in [K]$ and $\forall \rvz \in \R^{d}$, consider the injected noise $\rho_{i}^{(k)} \sim \gN(0, (\sigma^{(k)})^2 \sI_d)$, $\forall i\in [n]$, if the regularization function is $\psi(\rvx) = \lambda \|\rvx\|_1$ for some parameter $\lambda > 0$ and if $\rvz$ is independent of $\rho_i^{(k)}, \forall i\in [n]$, then the error caused by noise injection in epoch $k$ is
    \begin{align*}
        \E\left[\frac{1}{n}\sum_{i=1}^{n} \langle \rho_i^{(k)}, \rvx_1^{(k+1)} - \rvz \rangle \right] 
        \leq (\sigma^{(k)})^2 nd \eta^2 \widehat{L}^{(k)*}
        + \frac{2 d(\sigma^{(k)})^2}{\sqrt{2\pi}} B
    \end{align*}
    where the expectation is taken w.r.t. the injected noise $\{\rho_i^{(k)}\}_{i=1}^{n}$.
\end{lemma}

\begin{proof}[Proof of Lemma~\ref{lemma:noise_bias_l1_reg}]

For epoch $k\in [K]$,
\begin{align*}
    \rvx_{1}^{(k+1)} = \argmin_{\rvx\in\R^{d}} n \lambda \|\rvx\|_1 + \frac{\|\rvx - \rvx_{n+1}^{(k)}\|^2}{2\eta}   
\end{align*}
and for $j\in [d]$, 
\begin{align*}
    \rvx_{1, j}^{(k+1)} = \begin{cases}
        \rvx_{n+1,j}^{(k)} - \eta \lambda n & \text{if $\rvx_{n+1, j}^{(k)} > \eta \lambda n$}\\
        \rvx_{n+1, j}^{(k)} + \eta \lambda n & \text{if $\rvx_{n+1, j}^{(k)} < -\eta \lambda n$}\\
        0 & \text{if $|\rvx_{n+1, j}^{(k)}| \leq \eta \lambda n$}
    \end{cases}
    := g(\rvx_{n+1,j}^{(k)})
\end{align*}
where $\rvx_{n+1,j}^{(k)}$ and $\rvx_{1,j}^{(k+1)}$ denote the $j$-th coordinate of $\rvx_{n+1}^{(k)}$ and $\rvx_{1}^{(k+1)}$, respectively.

For $i\in [n]$, conditional on $\rho_k^{(k)}, \forall k\neq i$,
\begin{align}
\label{eq:l1_reg_cond_exp}
    \E_{\rho_i^{(k)}}\left[\left\langle \rvx_{1}^{(k+1)}, \rho_i^{(k)} \right\rangle \mid \{\rho_k^{(k)}\}_{k\neq i}\right]
    = \sum_{j=1}^{d} \E_{\rho_{i,j}^{(k)}}\left[ \rvx_{1, j}^{(k+1)} \rho_{i, j}^{(k)} \mid \{\rho_{l, j}^{(k)}\}_{l \neq i}\right]
\end{align}
where $\rho_{i, j}^{(k)} \in \gN(0, (\sigma^{(k)})^2)$ denotes the $j$-th coordinate of the noise vector $\rho_i^{(k)}$, $\forall i\in [n], j\in[d]$.

For $i\in [n]$ and $j\in [d]$, let $m_{i,j}^{(k)}$ denote the value of the random variable $\rho_{i, j}^{(k)}$. 
\begin{align*}
    &\E_{\rho_{i,j}^{(k)}}\left[\rvx_{1,j}^{(k+1)} \rho_{i, j}^{(k)} \mid \{\rho_{k, j}^{(k)}\}_{k \neq i} \right]
    = \E_{\rho_{i,j}^{(k)}}\left[ g(\rvx_{n+1,j}^{(k)}) \rho_{i, j}^{(k)} \mid \{\rho_{l, j}\}_{l \neq i}^{(k)} \right]\\
    &= \frac{1}{\sigma^{(k)} \sqrt{2\pi}}
    \Bigg(
    \int_{\rvx_{n+1,j}^{(k)} < -\eta \lambda n} \underbrace{(\rvx_{n+1,j}^{(k)} + \eta \lambda n)}_{:=u_1}
    \underbrace{
    m_{i, j}^{(k)} e^{-\frac{(m_{i,j}^{(k)})^2}{2 (\sigma^{(k)})^2}}
    }_{:=d v_1} dm_{i, j}^{(k)}\\
    \nonumber
    &\quad + \int_{\rvx_{n+1,j}^{(k)} > \eta \lambda n} \underbrace{(\rvx_{n+1,j}^{(k)} - \eta \lambda n)}_{:= u_2} 
    \underbrace{m_{i, j}^{(k)} e^{-\frac{(m_{i, j}^{(k)})^2}{2 (\sigma^{(k)})^2}}}_{:=d v_2} dm_{i, j}^{(k)}
    \Bigg)
\end{align*}

Let $u_1 = \rvx_{n+1,j}^{(k)} + \eta \lambda n$ and $d v_1 = m_{i,j}^{(k)} e^{-\frac{(m_{i,j}^{(k)})^2}{2 (\sigma^{(k)})^2}} dm_{i, j}^{(k)}$. 
Then, $d u_1 = \frac{\partial \rvx_{n+1,j}^{(k)}}{\partial \rho_{i, j}^{(k)}} d m_{i,j}^{(k)}$ and $v_1 = -(\sigma^{(k)})^2 e^{-\frac{(m_{i,j}^{(k)})^2}{2 (\sigma^{(k)})^2}}$.
Similarly, let $u_2 = \rvx_{n+1,j}^{(k)} - \eta \lambda n$ and $d v_2 = m_{i,j}^{(k)} e^{-\frac{(m_{i,j}^{(k)})^2}{2 (\sigma^{(k)})^2}}$. Then, $d u_2 = \frac{\partial \rvx_{n+1,j}^{(k)}}{\partial \rho_{i, j}^{(k)}}$ and $v_2 = -(\sigma^{(k)})^2 e^{-\frac{(m_{i,j}^{(k)})^2}{2 (\sigma^{(k)})^2}}$.
Using integration by parts,
\begin{align}
\nonumber
    &\frac{1}{\sigma^{(k)} \sqrt{2\pi}}
    \int_{\rvx_{n+1, j}^{(k)} < -\eta \lambda n} 
        (\rvx_{n+1,j}^{(k)} + \eta \lambda n)
        m_{i,j}^{(k)} e^{-\frac{(m_{i,j}^{(k)})^2}{2 (\sigma^{(k)})^2}} d m_{i,j}^{(k)}\\
    \nonumber
    &= \frac{1}{\sigma^{(k)} \sqrt{2\pi}}
    \Big[ - (\rvx_{n+1,j}^{(k)} + \eta \lambda n) (\sigma^{(k)})^2 e^{-\frac{(m_{i,j}^{(k)})^2}{2(\sigma^{(k)})^2}} \Big]_{\rvx_{n+1,j}^{(k)} < -\eta \lambda n}\\
    \nonumber
    &\quad - \frac{1}{\sigma^{(k)} \sqrt{2\pi}}
    \int_{\rvx_{n+1, j}^{(k)} < -\eta \lambda n}
    - (\sigma^{(k)})^2 e^{-\frac{(m_{i,j}^{(k)} )^2}{2 (\sigma^{(k)})^2}} \frac{\partial \rvx_{n+1, j}^{(k)}}{\partial \rho_{i, j}} dm_{i,j}^{(k)}\\
\label{eq:appendix_l1_1}
    &= \frac{1}{\sigma^{(k)} \sqrt{2\pi}}
    \Big[ - (\rvx_{n+1,j}^{(k)} + \eta \lambda n) (\sigma^{(k)})^2 e^{-\frac{(m_{i,j}^{(k)})^2}{2(\sigma^{(k)})^2}} \Big]_{\rvx_{n+1,j}^{(k)} < -\eta \lambda n}\\
    \nonumber
    &\quad + \frac{ (\sigma^{(k)})^2}{\sigma^{(k)} \sqrt{2\pi}} \int_{\rvx_{n+1, j}^{(k)} < - \eta \lambda n}
    \frac{\partial \rvx_{n+1,j}^{(k)}}{\partial \rho_{i, j}^{(k)}} e^{-\frac{(m_{i,j}^{(k)})^2}{2 (\sigma^{(k)})^2}} dm_{i,j}^{(k)}
\end{align}
and
\begin{align}
\nonumber
    &\frac{1}{\sigma^{(k)} \sqrt{2\pi}} 
        \int_{\rvx_{n+1,j}^{(k)} > \eta \lambda n} (\rvx_{n+1,j}^{(k)} - \eta \lambda n) 
        m_{i,j}^{(k)} e^{-\frac{(m_{i,j}^{(k)})^2}{2 (\sigma^{(k)})^2}} d m_{i,j}^{(k)}\\
    \nonumber
    &= \frac{1}{\sigma^{(k)} \sqrt{2\pi}}\left[
    - (\rvx_{n+1,j}^{(k)} - \eta \lambda n) (\sigma^{(k)})^2 e^{-\frac{(m_{i,j}^{(k)})^2}{2 (\sigma^{(k)})^2}}
    \right]_{\rvx_{n+1,j}^{(k)} > \eta \lambda n}\\
    \nonumber
    &\quad - \frac{1}{\sigma^{(k)} \sqrt{2\pi}}
    \int_{\rvx_{n+1,j}^{(k)} > \eta \lambda n}
    - (\sigma^{(k)})^2 e^{-\frac{(m_{i,j}^{(k)})^2}{2 (\sigma^{(k)})^2}} \frac{\partial \rvx_{n+1,j}^{(k)}}{\partial \rho_{i, j}} dm_{i,j}^{(k)}\\
\label{eq:appendix_l1_2}
    &= \frac{1}{\sigma^{(k)} \sqrt{2\pi}}\left[
    - (\rvx_{n+1,j}^{(k)} - \eta \lambda n) (\sigma^{(k)})^2 e^{-\frac{(m_{i,j}^{(k)})^2}{2 (\sigma^{(k)})^2}}
    \right]_{\rvx_{n+1,j}^{(k)} > \eta \lambda n}\\
    \nonumber
    &\quad + \frac{(\sigma^{(k)})^2}{\sigma^{(k)} \sqrt{2\pi}}
    \int_{\rvx_{n+1,j}^{(k)} > \eta \lambda n} \frac{\partial \rvx_{n+1,j}^{(k)}}{\partial \rho_{i, j}^{(k)}}
    e^{-\frac{(m_{i,j}^{(k)})^2}{2 (\sigma^{(k)})^2}} d m_{i,j}^{(k)}
\end{align}

Summing up Eq.~\ref{eq:appendix_l1_1} and Eq.~\ref{eq:appendix_l1_2},
\begin{align}
\label{eq:l1_reg_exp_single_coord}
    &\E_{\rho_{i,j}^{(k)}}\left[ \rvx_{1,j}^{(k+1)} \rho_{i, j}^{(k)} \mid \{\rho_{k,j}^{(k)}\}_{k\neq i} \right]\\
    \nonumber
    &= \frac{1}{\sigma^{(k)} \sqrt{2\pi}} \Bigg( \Big[ -(\rvx_{n+1,j}^{(k)} + \eta \lambda n) (\sigma^{(k)})^2 e^{-\frac{(m_{i,j}^{(k)})^2}{2 (\sigma^{(k)})^2}}  \Big]_{\rvx_{n+1,j}^{(k)} < - \eta \lambda n} + \Big[ -(\rvx_{n+1,j}^{(k)} - \eta \lambda n) (\sigma^{(k)})^2 e^{-\frac{(m_{i,j}^{(k)})^2}{2 (\sigma^{(k)})^2}} \Big]_{\rvx_{n+1,j}^{(k)} > \eta \lambda n}\Bigg) \\
    \nonumber
    &\quad + \frac{ (\sigma^{(k)})^2}{ (\sigma^{(k)}) \sqrt{2\pi}} \Bigg(
    \int_{\rvx_{n+1,j}^{(k)} < - \eta \lambda n} \frac{\partial \rvx_{n+1,j}^{(k)}}{\partial \rho_{i, j}^{(k)} } e^{-\frac{(m_{i,j}^{(k)})^2}{2 (\sigma^{(k)})^2}}dm_{i,j}^{(k)}
    + \int_{\rvx_{n+1,j}^{(k)} > \eta \lambda n} \frac{\partial \rvx_{n+1,j}^{(k)}}{\partial \rho_{i, j}^{(k)} } e^{-\frac{(m_{i,j}^{(k)})^2}{2 (\sigma^{(k)})^2}} dm_{i,j}^{(k)}
    \Bigg)
\end{align}

We use $\rvx_{n+1,j}^{(k)}(m_{i,j}^{(k)})$ to denote the value of $\rvx_{n+1,j}^{(k)}$ as a function of the noise value $m_{i,j}^{(k)}$.

By Eq.~\ref{eq:op_ub}, $\left\| \frac{\rvx_{n+1}^{(k)}}{\partial \rho_i^{(k)}} \right\|_{op} \leq n \eta^2 \widehat{L}^{(k)*}$ is bounded, $\forall i\in [n], k\in [K]$, and hence, $|\frac{\partial \rvx_{n+1,j}^{(k)}(m_{i,j}^{(k)})}{\partial m_{i,j}^{(k)}}|$ is also bounded.
This implies that $\frac{1}{\sqrt{2\pi}} \cdot -(\rvx_{n+1,j}^{(k)}(m_{i,j}^{(k)}) + \eta \lambda n) (\sigma^{(k)})^2 e^{-\frac{(m_{i,j}^{(k)})^2}{2 (\sigma^{(k)})^2}}\rightarrow 0$ 
and $\frac{1}{\sqrt{2\pi}}\cdot -(\rvx_{n+1,j}^{(k)}(m_{i,j}^{(k)}) - \eta \lambda n) (\sigma^{(k)})^2 e^{-\frac{(m_{i,j}^{(k)})^2}{2(\sigma^{(k)})^2}} \rightarrow 0$
as $m_{i,j}^{(k)} \rightarrow \infty$ or $m_{i,j}^{(k)} \rightarrow -\infty$. 

We now compute the terms in the above Eq.~\ref{eq:l1_reg_exp_single_coord} separately.
\begin{align*}
    &\frac{1}{\sigma^{(k)} \sqrt{2\pi}}\left[ -(\rvx_{n+1,j}^{(k)} + \eta \lambda n) (\sigma^{(k)})^2 e^{-\frac{(m_{i,j}^{(k)})^2}{2 (\sigma^{(k)})^2}}  \right]_{\rvx_{n+1,j}^{(k)} < - \eta \lambda n}\\
    \nonumber
    &= \frac{1}{\sigma^{(k)} \sqrt{2\pi}}\left[ -(\rvx_{n+1,j}^{(k)}(m_{i,j}^{(k)}) + \eta \lambda n) (\sigma^{(k)})^2 e^{-\frac{(m_{i,j}^{(k)})^2}{2 (\sigma^{(k)})^2}}  \right]_{\inf\{m: \rvx_{n+1,j}^{(k)}(m) < - \eta \lambda n\}}^{\sup\{m: \rvx_{n+1,j}^{(k)}(m) < - \eta \lambda n\}}
\end{align*}

Note that $\sup\{m: \rvx_{n+1,j}^{(k)}(m) < - \eta \lambda n\}$ is finite and $\inf\{m: \rvx_{n+1,j}^{(k)}(m) < - \eta \lambda n\} \rightarrow -\infty$. Let $m_{+} = \sup\{m: \rvx_{n+1,j}^{(k)}(m) < - \eta \lambda n\}$, and the above is then
\begin{align}
\nonumber
    &\frac{1}{\sigma^{(k)} \sqrt{2\pi}}\left[ -(\rvx_{n+1,j}^{(k)} + \eta \lambda n) (\sigma^{(k)})^2 e^{-\frac{(m_{i,j}^{(k)})^2}{2 (\sigma^{(k)})^2}}  \right]_{\rvx_{n+1,j}^{(k)} < - \eta \lambda n}\\
    \nonumber
    &= \frac{1}{\sigma^{(k)} \sqrt{2\pi}}\cdot -(\rvx_{n+1,j}^{(k)}(m_{+}) + \eta \lambda n) (\sigma^{(k)})^2 e^{-\frac{m_{+}^2}{2 (\sigma^{(k)})^2}} - 0\\
\label{eq:l1_reg_t1}
    &= -\frac{1}{\sigma^{(k)} \sqrt{2\pi}}(\rvx_{n+1,j}^{(k)}(m_{+}) + \eta \lambda n) (\sigma^{(k)})^2 e^{-\frac{m_{+}^2}{2 (\sigma^{(k)})^2}}
\end{align}

Similarly, note that $\sup\{m: \rvx_{n+1,j}^{(k)}(m) > \eta \lambda n\}\rightarrow \infty$ and $\inf\{m: \rvx_{n+1,j}^{(k)}(m) > \eta \lambda n\}$ is finite. Let $m_{-} = \inf\{m: \rvx_{n+1,j}^{(k)}(m) > \eta \lambda n\}$.
\begin{align}
\nonumber
    &\frac{1}{(\sigma^{(k)}) \sqrt{2\pi}}
    \left[ -(\rvx_{n+1,j}^{(k)} - \eta \lambda n) (\sigma^{(k)})^2 e^{-\frac{(m_{i,j}^{(k)})^2}{2 (\sigma^{(k)})^2}} \right]_{\rvx_{n+1,j}^{(k)} > \eta \lambda n}\\
    \nonumber
    &= \frac{1}{\sigma^{(k)} \sqrt{2\pi}}
    \left[ -(\rvx_{n+1,j}^{(k)}(m_{i,j}^{(k)}) - \eta \lambda n) (\sigma^{(k)})^2 e^{-\frac{(m_{i,j}^{(k)})^2}{2 (\sigma^{(k)})^2}} \right]_{\inf\{m: \rvx_{n+1,j}^{(k)}(m) > \eta \lambda n\}}^{\sup\{m: \rvx_{n+1,j}^{(k)}(m) > \eta \lambda n\} }\\
\nonumber
    &= 0 - \frac{1}{\sigma^{(k)} \sqrt{2\pi}}\cdot -(\rvx_{n+1,j}^{(k)}(m_{-}) - \eta \lambda n) (\sigma^{(k)})^2 e^{-\frac{(m_{-})^2}{2 (\sigma^{(k)})^2}}\\
\label{eq:l1_reg_t2}
    &= \frac{1}{\sigma^{(k)} \sqrt{2\pi}} (\rvx_{n+1,j}^{(k)}(m_{-}) - \eta \lambda n) (\sigma^{(k)})^2 e^{-\frac{(m_{-})^{2}}{2 (\sigma^{(k)})^2}}
\end{align}

Furthermore,
\begin{align}
\nonumber
    &\frac{(\sigma^{(k)})^2}{\sigma^{(k)} \sqrt{2\pi}}\left(
    \int_{\rvx_{n+1,j}^{(k)} < - \eta \lambda n} \frac{\partial \rvx_{n+1,j}^{(k)}}{\partial \rho_{i, j}^{(k)} } e^{-\frac{(m_{i,j}^{(k)})^2}{2 (\sigma^{(k)})^2}}dm_{i,j}^{(k)}
    + \int_{\rvx_{n+1,j}^{(k)} > \eta \lambda n} \frac{\partial \rvx_{n+1,j}^{(k)}}{\partial \rho_{i, j}^{(k)} } e^{-\frac{(m_{i,j}^{(k)})^2}{2(\sigma^{(k)})^2}} dm_{i,j}^{(k)}
    \right)\\
    \nonumber
    &\leq \frac{(\sigma^{(k)})^2}{\sigma^{(k)} \sqrt{2\pi}}\left(
    \int_{\rvx_{n+1,j}^{(k)} < - \eta \lambda n} \left|\frac{\partial \rvx_{n+1,j}^{(k)}}{\partial \rho_{i, j}^{(k)} }\right| e^{-\frac{(m_{i,j}^{(k)})^2}{2(\sigma^{(k)})^2}}dm_{i,j}^{(k)}
    + \int_{\rvx_{n+1,j}^{(k)} > \eta \lambda n} \left|\frac{\partial \rvx_{n+1,j}^{(k)}}{\partial \rho_{i, j}} \right| e^{-\frac{(m_{i,j}^{(k)})^2}{2(\sigma^{(k)})^2}} dm_{i,j}^{(k)}
    \right)\\
\nonumber
    &\leq \frac{(\sigma^{(k)})^2}{\sigma^{(k)} \sqrt{2\pi}}
    \int_{\R} \left| \frac{\partial \rvx_{n+1,j}^{(k)}}{\partial \rho_{i,j}^{(k)}} \right| e^{-\frac{(m_{i,j}^{(k)})^2}{2 (\sigma^{(k)})^2}}
    d m_{i,j}^{(k)}\\
\nonumber
    &\leq (\sigma^{(k)})^2 \max\left| \frac{\partial \rvx_{n+1,j}^{(k)}}{\partial \rho_{i, j}^{(k)}} \right|\cdot \frac{1}{\sigma^{(k)}\sqrt{2\pi}} \int_{\R} e^{-\frac{(m_{i,j}^{(k)})^2}{2(\sigma^{(k)})^2}} d m_{i,j}^{(k)}\\
\label{eq:l1_reg_t3}
    &= (\sigma^{(k)})^2 \max\left| \frac{\partial \rvx_{n+1,j}^{(k)}}{\partial \rho_{i, j}^{(k)}} \right|
\end{align}

Plugging Eq.~\ref{eq:l1_reg_t1}, Eq.~\ref{eq:l1_reg_t2} and Eq.~\ref{eq:l1_reg_t3} back to Eq.~\ref{eq:l1_reg_exp_single_coord},
\begin{align*}
    &\E_{\rho_{i,j}^{(k)}}\left[\rvx_{1,j}^{(k+1)} \rho_{i, j}^{(k)} \mid \{\rho_{k, j}^{(k)}\}_{k \neq i} \right]
    \leq (\sigma^{(k)})^2 \max\left| \frac{\partial \rvx_{n+1,j}^{(k)}}{\partial \rho_{i, j}^{(k)} } \right|\\
    \nonumber
    &\quad \underbrace{-\frac{1}{\sigma^{(k)}\sqrt{2\pi}}(\rvx_{n+1,j}^{(k)}(m_{+}) + \eta \lambda n) (\sigma^{(k)})^2 e^{-\frac{m_{+}^2}{2(\sigma^{(k)})^2}}
    + \frac{1}{\sigma^{(k)} \sqrt{2\pi}} (\rvx_{n+1,j}^{(k)}(m_{-}) - \eta \lambda n) (\sigma^{(k)})^2 e^{-\frac{(m_{-})^{2}}{2 (\sigma^{(k)})^2}}
    }_{:= \Delta_{i,j}^{(k)}}
\end{align*}

where $m_{+} = \sup\{m: \rvx_{n+1,j}^{(k)}(m) < -\eta \lambda n\}$ and $m_{-} = \inf\{m: \rvx_{n+1,j}^{(k)}(m) > \eta \lambda n\}$.
We note that $-(\rvx_{n+1,j}^{(k)}(m_{+}) + \eta \lambda n) \geq 0$ and $(\rvx_{n+1,j^{(k)}}(m_{-}) - \eta \lambda n) \geq 0$. 

Let $\Delta_{i, j}^{(k)} = -\frac{1}{\sigma^{(k)} \sqrt{2\pi}}(\rvx_{n+1,j}^{(k)}(m_{+}) + \eta \lambda n) (\sigma^{(k)})^2 e^{-\frac{m_{+}^2}{2 (\sigma^{(k)})^2}}
+ \frac{1}{\sigma^{(k)} \sqrt{2\pi}} (\rvx_{n+1,j}^{(k)}(m_{-}) - \eta \lambda n) (\sigma^{(k)})^2 e^{-\frac{(m_{-})^{2}}{2 (\sigma^{(k)})^2}}$.
Note that the subscript $i$ here indicates in $\Delta_{i, j}^{(k)}$, the value $\rvx_{n+1,j}^{(k)}(m)$ is a deterministic function of the noise $\rho_{i}^{(k)}$'s value $m$, by fixing the values of $\{\rho_{k}^{(k)}\}$, $\forall i\neq k$.
Also, when $\eta \lambda n = 0$, we can integrate over $\R$ and in this case $\Delta_j^{(k)} = 0$. This case is essentially what Stein's lemma addresses.

Following Eq.~\ref{eq:l1_reg_cond_exp}, for $i\in [n]$, conditional on $\rho_k^{(k)}, \forall k\neq i$,
\begin{align*}
    \E_{\rho_i^{(k)}}\left[\left\langle \rvx_{1}^{(k+1)}, \rho_i^{(k)} \right\rangle \mid \{\rho_k^{(k)}\}_{k\neq i}\right]
    &= \sum_{j=1}^{d} \E_{\rho_{i,j}^{(k)}}\left[ \rvx_{1, j}^{(k+1)} \rho_{i, j}^{(k)} \mid \{\rho_{k, j}^{(k)}\}_{k \neq i}\right]\\
    &\leq (\sigma^{(k)})^2 \sum_{j=1}^{d} \max\left| \frac{\partial \rvx_{n+1,j}^{(k)}}{\partial \rho_{i, j}^{(k)}} \right|
    + \sum_{j=1}^{d} \E\left[\Delta_{i,j}^{(k)} \mid \{\rho_k^{(k)}\}_{k\neq i} \right]
\end{align*}

Note that $\sum_{j=1}^{d} \max \left| \frac{\partial \rvx_{n+1,j}^{(k)}}{\partial \rho_{i, j}^{(k)}} \right| = \sum_{j=1}^{d} \max \left| \left[\frac{\partial \rvx_{n+1}^{(k)}}{\partial \rho_{i}^{(k)}}\right]_{j,j} \right|$, i.e., sum of the absolute value of the $j$-th diagonal entry of the Jacobian matrix $\frac{\partial \rvx_{n+1}^{(k)}}{\partial \rho_i^{(k)}}$. 
Again, by Eq.~\ref{eq:op_ub}, $\left\| \frac{\rvx_{n+1}^{(k)}}{\partial \rho_i^{(k)}} \right\|_{op} \leq n \eta^2 \widehat{L}^{(k)*}$ is bounded, $\forall i\in [n], k\in [K]$.
Therefore, for $i\in [n]$,
\begin{align*}
    \sum_{j=1}^{d}\max\left|\frac{\partial \rvx_{n+1,j}^{(k)}}{\partial \rho_{i, j}^{(k)}}\right|
    \leq \max \left\| \frac{\partial \rvx_{n+1}^{(k)}}{\partial \rho_{i}^{(k)}} \right\|_{tr}
    \leq d \cdot \max \left\| \frac{\partial \rvx_{n+1}^{(k)}}{\partial \rho_{i}^{(k)}} \right\|_{op}
    \leq d n \eta^2 \widehat{L}^{(k)*}
\end{align*}
where $\|\cdot\|$ denotes the trace norm of a matrix. Hence, by law of total expectation,
\begin{align*}
    \E_{\rho_{i}^{(k)}}\left[ \left\langle \rvx_1^{(k+1)}, \rho_i^{(k)} \right\rangle \right]
    &= \E\left[\E_{\rho_i^{(k)}}\left[\left\langle \rvx_{1}^{(k+1)}, \rho_i^{(k)} \right\rangle \mid \{\rho_k^{(k)}\}_{k\neq i}\right]\right]\\
    &\leq (\sigma^{(k)})^2 d n \eta^2 \widehat{L}^{(k)*}
    + \sum_{j=1}^{d}\E\left[\E_{\rho_i^{(k)}}\left[ \Delta_{i,j}^{(k)} \mid  \{\rho_{k}^{(k)}\}_{k\neq i}\right]\right]\\
    &\leq (\sigma^{(k)})^2 d n \eta^2 \widehat{L}^{(k)*}
    + \sum_{j=1}^{d} \frac{2 (\sigma^{(k)})^2}{\sqrt{2\pi}} B
    = (\sigma^{(k)})^2 d n \eta^2 \widehat{L}^{(k)*}
    + \frac{2 d (\sigma^{(k)})^2 }{\sqrt{2\pi}} B
\end{align*}
where the last inequality is by Assumption~\ref{ass:l1_offset} and the fact that $\frac{1}{(\sigma^{(k)})\sqrt{2\pi}} e^{-\frac{m^2}{2(\sigma^{(k)})^2}}\leq 1$. Therefore,
\begin{align*}
    \E\left[\frac{1}{n}\sum_{i=1}^{n} \langle \rho_i^{(k)}, \rvx_1^{(k+1)} - \rvz \rangle \right] 
    \leq (\sigma^{(k)})^2 nd \eta^2 \widehat{L}^{(k)*}
    + \frac{2 d (\sigma^{(k)})^2}{\sqrt{2\pi}} B
\end{align*}

\end{proof}

The rest of the proof follows section~\ref{subsec:appendix_expected_one_epoch_convergence} and section~\ref{subsec:appendix_k_epoch_convergence}. The final bound we get in the case where $\psi(\rvx) = \frac{\lambda}{2}\|\rvx_1\|$, for $\lambda > 0$, is

\begin{theorem}[Convergence under $\ell_1$ regularization]
\label{thm:convergence_l1}
    Under Assumption~\ref{ass:convexity},~\ref{ass:l1_offset},~\ref{ass:H_smoothness},~\ref{ass:dissim_partial_lipschitzness},~\ref{ass:appendix_refined_smoothness}, for $\beta > 0$, 
    if $\mu_{\psi} \geq L_H^{(k)} + \beta$, $\forall k\in [K]$, and 
     $\eta \leq \frac{1}{2n \sqrt{10 \bar{L}^* \max_{k\in [K]}\widehat{L}^{(k)*} (1+\log K)}}$, 
    where $\bar{L}^* = \max\{L, \max_{k\in[K]} \widehat{L}^{(k)}\}$,
    Algorithm~\ref{alg:generalized_shuffled_gradient_fm} guarantees

    \begin{align*}
        &\E\left[ G(\rvx_1^{(K+1)}) \right] - G(\rvx^*)
        \leq \frac{\|\rvx_1^{(1)} - \rvx^* \|^2}{\eta n K}
        + 10 \eta^2 n^2 \sigma_{any}^2 (1+\log K)\max_{k\in[K]} \widehat{L}^{(k)} + 2 M
    \end{align*}
    where
    \begin{align*}
        M &= \max_{s\in [K]} \Big(
        \frac{1}{2n^2 \beta}\sum_{k=1}^{s} \frac{ (C_n^{(k)})^2 }{s+1-k}
        + 5\eta^2 \sum_{k=1}^{s} \frac{ \widehat{L}^{(k)*} \frac{1}{n}\sum_{i=n_d^{(k)}+1}^{n-1}(C_i^{(k)})^2} {s+1-k}\\
        \nonumber
        &\quad + 6\eta^2 nd \sum_{k=1}^{s} \frac{ (\sigma^{(k)})^2 \widehat{L}^{(k)*} }{s+1-k} 
        + 2\sum_{k=1}^{s} \frac{d (\sigma^{(k)})^2 B}{\sqrt{2\pi}(s+1-k)}\Big)
    \end{align*}
    and the expectation is taken w.r.t. the injected noise $\{\rho_i^{(k)}\}$ and the order of samples $\pi^{(k)}$, $\forall i\in [n], k\in [K]$.
\end{theorem}

The convergence of $\DPSG$ in this case is:
\begin{corollary}[Convergence of $\DPSG$ under $\ell_1$ regularization]
\label{corollary:convergence_dpsg_l1}
    Under the conditions in Theorem~\ref{thm:convergence_l1}, with $n_d^{(k)} = n, \gP^{(k)} = \emptyset$, and $\sigma^{(k)} = \sigma$, for all $k\in [K]$, Algorithm~\ref{alg:generalized_shuffled_gradient_fm} ($\DPSG$ under $l_1$ regularization) guarantees
    \begin{align*}
        & \E [G(\rvx_1^{(K+1)})] - G(\rvx^*)\\
        \nonumber
        &\lesssim \frac{\|\rvx_1^{(1)} - \rvx^{*}\|^2}{\eta n K} 
        + \eta^2 n^2 \sigma_{any}^2 L (1+\log K)
        + \eta^2 n d \sigma^2 L^{*} (1 + \log K)
        + d \sigma^2 B (1+\log K)
    \end{align*}
    and the expectation is taken w.r.t. the injected noise $\{\rho_i^{(k)}\}$ and the order of samples $\pi^{(k)}$, $\forall i\in [n], k\in [K]$.
\end{corollary}

Since the additional term is non-vanishing, the choice of $\sigma, \eta$ and $K$ is the same as in the case where $\psi$ is twice differentiable, with $\sigma = \widetilde{\gO}(\frac{G^{*}\sqrt{K}}{\eps}), \eta = \widetilde{\gO}(\frac{1}{n L^{*}K^{1/3}}), K= \gO(\frac{n\eps^2}{d})$.
The empirical excess risk in this case is then $\E\left[G(\rvx_1^{(K+1)})\right] - G(\rvx^*) = \widetilde{O}\left(\frac{1}{n^{2/3}} \Big( \frac{\sqrt{d}}{\eps} \Big)^{4/3} + nB \right)$.
We leave it as an open question whether this additional $nB$ term can be  eliminated under $\ell_1$ regularization.

\subsection{The Projection Operator}
\label{subsec:appendix_proj_op}

In this section, we consider the regularization function being the projection operator, i.e., $\psi(\rvx) = \gI\{\rvx \in \gB\}$, where $\gI$ is the indicator function and $\gB$ is a convex set.
We again need to re-compute the expectation of the additional error term introduced by noise injection (Lemma~\ref{lemma:noise_bias}).
We cannot apply Stein's lemma or directly use integration by parts in this case. Instead, we use Young's inequality to break the correlation between $\rho_i^{(k)}$ and $\rvx_1^{(k+1)}$, where recall that $\rvx_1^{(k+1)}$ is a function of $\rho_i^{(k)}$. This leads to a non-vanishing variance term (one that does not scale with the learning rate $\eta$) due to the variance of $\rho_i^{(k)}$. We leave as an open question whether this term can further be reduced when $\psi$ is the projection operator.

Unlike in the $\ell_1$ regularization case, the additional error term due to noise injection here introduces other terms that can be subsumed into the convergence bound. 
After deriving the additional error term in Lemma~\ref{lemma:noise_bias_proj_op}, we derive the convergence bound of one epoch in expectation in Lemma~\ref{lemma:expected_one_epoch_convergence_proj}
and finally the full convergence bound across $K$ epochs in Lemma~\ref{thm:convergence_proj}.

\begin{lemma}[Additional Error (Projection Operator)]
\label{lemma:noise_bias_proj_op}
    For any epoch $k\in [K]$ and $\forall \rvz \in \R^{d}$, consider the injected noise $\rho_{i}^{(k)} \sim \gN(0, (\sigma^{(k)})^2 \sI_d)$, $\forall i\in [n]$, if the regularization function is $\psi(\rvx) = \gI\{\rvx \in \gB\}$ for a convex set $\gB$ and if $\rvz$ is independent of $\rho_i^{(k)}, \forall i\in [n]$, then the error caused by noise injection in epoch $k$ is
    \begin{align}
    \label{eq:noise_bias_proj_op_eq}
        &\E\left[\frac{1}{n}\sum_{i=1}^{n} \langle \rho_i^{(k)}, \rvx_1^{(k+1)} - \rvz \rangle \right] 
        \leq (\sigma^{(k)})^2 nd \eta^2 \widehat{L}^{(k)*}
        + \frac{1}{2}d (\sigma^{(k)})^2\\
        \nonumber
        &\quad + \frac{5\eta^2}{2} \Bigg(
        n\sum_{j=1}^{n_d^{(k)}} \E\left[\Big\| \nabla f_{\pi_j^{(k)}}(\rvx_j^{(k)}) - \nabla f_{\pi_j^{(k)}}(\rvz)\Big\|^2\right]
        + n \sum_{j=n_d^{(k)}+1}^{n} \E\left[\Big\| \nabla f_{j-n_d^{(k)}}^{(k,pub)}(\rvx_j^{(k)})
        - \nabla f_{j-n_d^{(k)}}^{(k,pub)}(\rvz)
        \Big\|^2 \right]\\
    \nonumber
        &\quad + (C_n^{(k)})^2
    + n L \E[B_F(\rvz, \rvx^*)] + n^2 \sigma_{any} + n d (\sigma^{(k)})^2
    \Bigg)
\end{align}
    where the expectation is taken w.r.t. the injected noise $\{\rho_i^{(k)}\}_{i=1}^{n}$.
\end{lemma}

\begin{proof}[Proof of Lemma~\ref{lemma:noise_bias_proj_op}]

For epoch $k\in [K]$,
\begin{align*}
    \rvx_1^{(k+1)} = \argmin_{\rvx\in \R^{d}} n \gI\{\rvx \in \gB\} + \frac{\|\rvx - \rvx_{n+1}^{(k)}\|^2}{2 \eta}
    = \argmin_{\rvx \in \gB} \|\rvx - \rvx_{n+1}^{(k)}\|
\end{align*}
$\rvx_{1}^{(k+1)}$ is essentially the projection of $\rvx_{n+1}^{(k)}$ onto $\gB$.

For $i\in [n]$, 
\begin{align*}
    \E\left[\left\langle \rvx_{1}^{(k+1)}, \rho_i^{(k)} \right\rangle \right]
    &= \E\left[\left\langle \rvx_{n+1}^{(k)}, \rho_i^{(k)} \right\rangle\right] + \E\left[\left\langle \rvx_{1}^{(k+1)} - \rvx_{n+1}^{(k)}, \rho_i^{(k)} \right\rangle\right]\\
    &\leq \E\left[\left\langle \rvx_{n+1}^{(k)}, \rho_i^{(k)} \right\rangle\right]
    + \frac{1}{2}\E\left[\left\| \rho_i^{(k)}\right\|^2\right]
    + \frac{1}{2}\E\left[\left\| \rvx_{n+1}^{(k)} - \rvx_{1}^{(k)}\right\|^2 \right]
\end{align*}
where the last step is by Young's inequality.

We apply Stein's lemma to bound $\E\left[\left\langle \rvx_{n+1}^{(k)}, \rho_i^{(k)} \right\rangle\right]$ as follows: Conditional on $\rho_{j}^{(k)}, \forall j\neq i$,
\begin{align*}
    \E_{\rho_i^{(k)}}\left[\left\langle \rvx_{n+1}^{(k)}, \rho_i^{(k)} \right\rangle \mid \{\rho_j^{(k)}\}_{j\neq i}\right]
    = (\sigma^{(k)})^2\cdot \E_{\rho_i^{(k)}}\left[\text{tr}\left( \frac{\partial \rvx_{n+1}^{(k)}}{\partial \rho_i^{(k)}}\right) \mid \{\rho_j^{(k)}\}_{j\neq i}\right]
\end{align*}

By Eq.~\ref{eq:op_ub}, if $\eta \leq \frac{1}{\widehat{L}^{(k)*}}$, $\forall i \in [n]$ and $s\in [K]$, $\left\| \frac{\rvx_{n+1}^{(k)}}{\partial \rho_i^{(k)}} \right\|_{op}
\leq n \eta^2 \widehat{L}^{(k)*}$. And so for $i\in [n]$,
\begin{align}
\label{eq:noise_bias_reg_proj_t1}
    \E\left[\left\langle \rvx_{n+1}^{(k)}, \rho_i^{(k)}\right\rangle\right]
    = \E\left[\E_{\rho_i^{(k)}}\left[ \left\langle \rvx_{n+1}^{(k)}, \rho_i^{(k)} \right\rangle \mid \{\rho_j^{(k)}\}_{j\neq i}\right]\right]
    \leq (\sigma^{(k)})^2 nd \eta^2 \widehat{L}^{(k)*}
\end{align}

By Lemma~\ref{lemma:noise_variance}, 
\begin{align}
\label{eq:noise_bias_reg_proj_t2}
    \E\left[\left\|\rho_i^{(k)} \right\|^2\right] \leq d (\sigma^{(k)})^2
\end{align}

We use similar techniques to bound $\E\left[\left\| \rvx_{n+1}^{(k)} - \rvx_1^{(k)} \right\|^2\right]$ as in the convergence proof in section~\ref{subsec:appendix_one_epoch_convergence} and in Lemma~\ref{lemma:expected_one_epoch_convergence}. Specifically, based on the update in Algorithm~\ref{alg:generalized_shuffled_gradient_fm},

\begin{align}
\nonumber
    &\E\left[ \left\| \rvx_{n+1}^{(k)} - \rvx_1^{(k)}\right\|^2 \right]\\
    \nonumber
    &= \eta^2 \E\Big[\Big\| \sum_{j=1}^{n_d^{(k)}} \Big(\nabla f_{\pi_j^{(k)}}(\rvx_j^{(k)}) + \rho_j^{(k)} \Big)
    + \sum_{j=n_d^{(k)}+1}^{n} (\nabla f_{j-n_d^{(k)}}^{(k, pub)}\Big(\rvx_{j}^{(k)}) + \rho_j^{(k)} \Big) \Big\|^2
    \Big]\\
\nonumber
    &\leq 5\eta^2 \Bigg(
        \E\left[\Big\|\sum_{j=1}^{n_d^{(k)}} \nabla f_{\pi_j^{(k)}}(\rvx_j^{(k)}) + \sum_{j=n_d^{(k)}+1}^{n} \nabla f_{j-n_d^{(k)} }^{(k, pub)}(\rvx_j^{(k)})
        - \sum_{j=1}^{n_d^{(k)} }\nabla f_{\pi_j^{(k)}}(\rvz)
        - \sum_{j=n_d^{(k)}+1}^{n} \nabla f_{j-n_d^{(k)}}^{(k, pub)}(\rvz)
        \Big\|^2\right]\\
    \nonumber
    &\quad + \E\left[ \Big\| \sum_{j=n_d^{(k)}+1}^{n} \nabla f_{j-n_d^{(k)}}^{(k,pub)}(\rvz)
    - \sum_{j=n_d^{(k)}+1}^{n}\nabla f_{\pi_j^{(k)}}(\rvz)
    \Big\|^2 \right]\\
    \nonumber
    &\quad + \E\left[\Big\| \sum_{j=1}^{n} \nabla f_{\pi_j^{(k)}}(\rvz) - \sum_{j=1}^{n} \nabla f_{\pi_j^{(k)}}(\rvx^*)
    \Big\|^2\right]
    + \E\left[\Big\| \sum_{j=1}^{n} \nabla f_{\pi_j^{(k)}}(\rvx^*) \Big\|^2\right]
    + \E\left[\Big\| \sum_{j=1}^{n} \rho_j^{(k)} \Big\|^2 \right]
    \Bigg)\\
\label{eq:noise_bias_reg_proj_t3}
    &\leq 5\eta^2 \Bigg(
        n\sum_{j=1}^{n_d^{(k)}} \E\left[\Big\| \nabla f_{\pi_j^{(k)}}(\rvx_j^{(k)}) - \nabla f_{\pi_j^{(k)}}(\rvz)\Big\|^2\right]
        + n \sum_{j=n_d^{(k)}+1}^{n} \E\left[\Big\| \nabla f_{j-n_d^{(k)}}^{(k,pub)}(\rvx_j^{(k)})
        - \nabla f_{j-n_d^{(k)}}^{(k,pub)}(\rvz)
        \Big\|^2 \right]\\
    \nonumber
    &\quad + (C_n^{(k)})^2
    + n L \E[ B_F(\rvz, \rvx^*) ] + n^2 \sigma_{any}^2 + n d (\sigma^{(k)})^2
    \Bigg)
\end{align}
where the last inequality is due to Jensen's inequality, Assumption~\ref{ass:dissim_partial_lipschitzness}, 
Lemma~\ref{lemma:breg_div_ub_lb}, the definition of $\sigma_{any}^2$ and Lemma~\ref{lemma:noise_variance}.

Combining Eqs.~\ref{eq:noise_bias_reg_proj_t1}, \ref{eq:noise_bias_reg_proj_t2}, and \ref{eq:noise_bias_reg_proj_t3}, and noting that $\rvz$ is independent of $\rho_i^{(k)}$ for all $i \in [n]$, we obtain the inequality (\ref{eq:noise_bias_proj_op_eq}) stated in the lemma statement.

\end{proof}

The additional error term due to noise injection when $\psi$ is the projection operator stated above slightly changes the constants in the convergence bound. We give the expected one-epoch convergence bound in this case in Lemma~\ref{lemma:expected_one_epoch_convergence_proj} as follows.

\begin{lemma}[Expected One Epoch Convergence (Projection)]
\label{lemma:expected_one_epoch_convergence_proj}
    Under Assumptions~\ref{ass:convexity},~\ref{ass:dissim_partial_lipschitzness},~\ref{ass:appendix_refined_smoothness} and~\ref{ass:H_smoothness}, for any epoch $k\in [K]$, $\beta > 0$ and $\forall \rvz \in \R^d$, 
    if $\psi(\rvx) = \gI\{\rvx\in \gB\}$ for a convex set $\gB$, $\eta \leq \frac{1}{n\sqrt{10 (\widehat{L}^{(k)} +1) \widehat{L}^{(k)*}}}$ and $\rvz$ is independent of $\rho_i^{(k)}$, $\forall i\in [n]$,
    Algorithm~\ref{alg:generalized_shuffled_gradient_fm} guarantees
    \begin{align*}
        &\E\left[G(\rvx_1^{(k+1)}) - G(\rvz)\right]
        \leq \frac{1}{2n\eta}\Big(\E\left[\|\rvz - \rvx_1^{(k)}\|^2 \right] - \E\left[\| \rvz - \rvx_1^{(k+1)}\|^2 \right] \Big)\\
    \nonumber
        &\quad + \Big( \frac{L_H^{(k)} + \beta}{2} - \frac{\mu_{\psi}}{2} \Big) \E\left[\| \rvz - \rvx_1^{(k+1)}\|^2 \right] 
        + 10 \eta^2 n^2 L (\widehat{L}^{(k)} +1) \E\left[B_F(\rvz, \rvx^*)\right]\\
        \nonumber
        &\quad
        + \underbrace{ 5\eta^2 n^2 (\widehat{L}^{(k)} +1) \sigma_{any}^2 }_{\text{Opt. Uncertainty}}
        + \underbrace{ 5\eta^2 \frac{1}{n}\sum_{i=n_d^{(k)}+1}^{n-1} \widehat{L}^{(k)*} (C_i^{(k)})^2
        + \frac{5}{2}\eta^2 (C_n^{(k)})^2
        }_{\text{Vanishing Dissimilarity}}
        + \underbrace{ \frac{1}{2n^2 \beta}(C_n^{(k)})^2 }_{\text{Non-vanishing Dissimilarity}}\\
    \nonumber
        &\quad + \underbrace{ 6 \eta^2 nd (\sigma^{(k)})^2 (\widehat{L}^{(k)*}+1) }_{\text{Injected Noise}}
        + \underbrace{ \frac{1}{2}d (\sigma^{(k)})^2}_{\text{Add. Error}}
    \end{align*}
    and the expectation is taken w.r.t. the injected noise $\{\rho_i^{(k)}\}$ and the order of samples $\pi^{(k)}$, $\forall i\in [n], k\in [K]$.
\end{lemma}

\begin{proof}[Proof of Lemma~\ref{lemma:expected_one_epoch_convergence_proj}]
Following Lemma~\ref{lemma:one_epoch_bg_3pid},~\ref{lemma:one_epoch_bound_bregmandiv} and~\ref{lemma:one_epoch_convergence}, after taking expectation, there is for any $\rvz\in \R^{d}$, and for $k\in [K]$,
\begin{align}
\label{eq:proj_interm}
    &\E\left[G(\rvx_1^{(k+1)}) - G(\rvz)\right]
    \leq \E\left[ H^{(k)}(\rvx_1^{(k+1)}) - H^{(k)}(\rvz)\right]
        + \frac{\E\left[\|\rvz - \rvx_1^{(k)}\|^2 \right]}{2n \eta}\\
    \nonumber
    &\quad - (\frac{1}{2n \eta} + \frac{\mu_\psi}{2}) \E\left[\|\rvz - \rvx_1^{(k+1)}\|^2\right]
        - \frac{1}{2n \eta}\E\left[\|\rvx_1^{(k+1)} - \rvx_1^{(k)}\|^2 \right]\\
    \nonumber
    &\quad + \E\left[\frac{1}{n}\sum_{i=1}^{n}\langle -\rho_i^{(k)}, \rvx_1^{(k+1)} - \rvz\rangle\right]
    + \widehat{L}^{(k)}\E\left[\| \rvx_1^{(k+1)} - \rvx_1^{(k)}\|^2\right]
    + 5\eta^2 \frac{1}{n}\sum_{i=n_d^{(k)}+1}^{n-1} \widehat{L}^{(k)*} (C_i^{(k)})^2\\
    \nonumber
    &\quad + 10 \eta^2 n^2 \widehat{L}^{(k)} L \E\left[B_F(\rvz, \rvx^*)\right]
    + 5\eta^2 n^2 \widehat{L}^{(k)} \sigma_{any}^2 + 5\eta^2 nd (\sigma^{(k)})^2 \widehat{L}^{(k)}\\
    \nonumber
    &\quad + 5\eta^2 n \widehat{L}^{(k)} \sum_{j=1}^{n_d^{(k)}} \E\left[\Big\| \nabla f_{\pi_j^{(k)}}(\rvx_{j}^{(k)}) - \nabla f_{\pi_j^{(k)}}(\rvz)\Big\|^2\right]
        + 5\eta^2 n \widehat{L}^{(k)} \sum_{j=n_d^{(k)}+1}^{n} \E\left[\Big\| \nabla f_{j-n_d^{(k)}}^{(k, pub)} - \nabla f_{j-n_d^{(k)}}^{(k,pub)}(\rvz)\Big\|^2\right]\\
    \nonumber
        &\quad - \frac{1}{n} \Big( 
        \sum_{i=1}^{n_d^{(k)}} \frac{ \E\left[\left\|\nabla f_{\pi_i^{(k)}}(\rvx_i) - \nabla f_{\pi_i^{(k)}}(\rvz) \right\|^2 \right] }{2 L_{\pi_i^{(k)}}}
        + \sum_{i=n_d^{(k)}+1}^{n} \frac{ \E\left[ \left\| \nabla f_{i-n_d}^{(k, pub)}(\rvx_i^{(k)}) - \nabla f_{i-n_d}^{(k, pub)}(\rvz) \right\|^2\right] }{2\widetilde{L}_{i-n_d}^{(k)}}
        \Big) 
\end{align}

By Lemma~\ref{lemma:noise_bias_proj_op},
\begin{align}
\label{eq:noise_bias_proj_op}
    &\E\left[\frac{1}{n}\sum_{i=1}^{n} \langle -\rho_i^{(k)}, \rvx_1^{(k+1)} - \rvz \rangle \right] 
    \leq (\sigma^{(k)})^2 nd \eta^2 \widehat{L}^{(k)*}
    + \frac{1}{2}d (\sigma^{(k)})^2\\
    \nonumber
    &\quad + \frac{5\eta^2}{2} \Bigg(
        n\sum_{j=1}^{n_d^{(k)}} \E\left[\Big\| \nabla f_{\pi_j^{(k)}}(\rvx_j^{(k)}) - \nabla f_{\pi_j^{(k)}}(\rvz)\Big\|^2\right]
        + n \sum_{j=n_d^{(k)}+1}^{n} \E\left[\Big\| \nabla f_{j-n_d^{(k)}}^{(k,pub)}(\rvx_j^{(k)})
        - \nabla f_{j-n_d^{(k)}}^{(k,pub)}(\rvz)
        \Big\|^2 \right]
        \\
    \nonumber
    &\quad + (C_n^{(k)})^2 + n L \E [B_F(\rvz, \rvx^*) ] + n^2 \sigma_{any}^2 + n d (\sigma^{(k)})^2
    \Bigg)
\end{align}

Plugging Eq.~\ref{eq:noise_bias_proj_op} back to Eq.~\ref{eq:proj_interm},
\begin{align}
\nonumber
    &\E\left[G(\rvx_1^{(k+1)}) - G(\rvz)\right]
    \leq \E\left[ H^{(k)}(\rvx_1^{(k+1)}) - H^{(k)}(\rvz)\right]
        + \frac{\E\left[\|\rvz - \rvx_1^{(k)}\|^2 \right]}{2n \eta}\\
    \nonumber
    &\quad - (\frac{1}{2n \eta} + \frac{\mu_\psi}{2}) \E\left[\|\rvz - \rvx_1^{(k+1)}\|^2\right]
        - \frac{1}{2n \eta}\E\left[\|\rvx_1^{(k+1)} - \rvx_1^{(k)}\|^2 \right]\\
    \nonumber
    &\quad
    + \widehat{L}^{(k)}\E\left[\| \rvx_1^{(k+1)} - \rvx_1^{(k)}\|^2\right]
    + 5\eta^2 \frac{1}{n}\sum_{i=n_d^{(k)}+1}^{n-1} \widehat{L}^{(k)*} (C_i^{(k)})^2
    + \frac{5}{2}\eta^2 (C_n^{(k)})^2\\
    \nonumber
    &\quad + 10 \eta^2 n^2 L (\widehat{L}^{(k)}+1) \E\left[B_F(\rvz, \rvx^*)\right]
    + 5\eta^2 n^2 (\widehat{L}^{(k)} +1) \sigma_{any}^2 + 6 \eta^2 nd (\sigma^{(k)})^2 (\widehat{L}^{(k)*}+1)
    + \frac{1}{2}d (\sigma^{(k)})^2 \\
    \nonumber
    &\quad + 5\eta^2 n (\widehat{L}^{(k)} +1 ) \sum_{j=1}^{n_d^{(k)}} \E\left[\Big\| \nabla f_{\pi_j^{(k)}}(\rvx_{j}^{(k)}) - \nabla f_{\pi_j^{(k)}}(\rvz)\Big\|^2\right]\\
    \nonumber
    &\quad + 5\eta^2 n (\widehat{L}^{(k)} + 1 )\sum_{j=n_d^{(k)}+1}^{n} \E\left[\Big\| \nabla f_{j-n_d^{(k)}}^{(k, pub)} - \nabla f_{j-n_d^{(k)}}^{(k,pub)}(\rvz)\Big\|^2\right]\\
    \nonumber
        &\quad - \frac{1}{n} \Big( 
        \sum_{i=1}^{n_d^{(k)}} \frac{ \E\left[\left\|\nabla f_{\pi_i^{(k)}}(\rvx_i) - \nabla f_{\pi_i^{(k)}}(\rvz) \right\|^2 \right]}{2 L_{\pi_i^{(k)}}}
        + \sum_{i=n_d^{(k)}+1}^{n} \frac{ \E\left[\left\| \nabla f_{i-n_d^{(k)}}^{(k, pub)}(\rvx_i^{(k)}) - \nabla f_{i-n_d^{(k)}}^{(k, pub)}(\rvz) \right\|^2 \right]}{2\widetilde{L}_{i-n_d^{(k)}}^{(k)}}
        \Big) 
\end{align}

If one sets the learning rate $\eta$ such that
\begin{align*}
    5\eta^2 n(\widehat{L}^{(k)} + 1) \leq \frac{1}{n}\cdot \frac{1}{2\widehat{L}^{(k)*}}, \quad
    \Rightarrow \eta \leq \frac{1}{n \sqrt{10 (\widehat{L}^{(k)} + 1) \widehat{L}^{(k)*}}}
\end{align*}
then it follows that
\begin{align}
\nonumber
    &\E\left[G(\rvx_1^{(k+1)}) - G(\rvz)\right]
    \leq \E\left[ H^{(k)}(\rvx_1^{(k+1)}) - H^{(k)}(\rvz)\right]
        + \frac{\E\left[\|\rvz - \rvx_1^{(k)}\|^2 \right]}{2n \eta}\\
    \nonumber
    &\quad - (\frac{1}{2n \eta} + \frac{\mu_\psi}{2}) \E\left[\|\rvz - \rvx_1^{(k+1)}\|^2\right]
        - \frac{1}{2n \eta}\E\left[\|\rvx_1^{(k+1)} - \rvx_1^{(k)}\|^2 \right]\\
    \nonumber
    &\quad
    + \widehat{L}^{(k)}\E\left[\| \rvx_1^{(k+1)} - \rvx_1^{(k)}\|^2\right]
    + 5\eta^2 \frac{1}{n}\sum_{i=n_d^{(k)}+1}^{n-1} \widehat{L}^{(k)*} (C_i^{(k)})^2
    + \frac{5}{2}\eta^2 (C_n^{(k)})^2\\
    \nonumber
    &\quad + 10 \eta^2 n^2 L (\widehat{L}^{(k)}+1) \E\left[B_F(\rvz, \rvx^*)\right]
    + 5\eta^2 n^2 (\widehat{L}^{(k)} +1) \sigma_{any}^2 + 6 \eta^2 nd (\sigma^{(k)})^2 (\widehat{L}^{(k)*}+1)
    + \frac{1}{2}d (\sigma^{(k)})^2
\end{align}

Finally, by Eq.~\ref{eq:dissim_ub},

\begin{align*}
    \E\left[H^{(k)}(\rvx_{1}^{(k+1)})\right] - \E\left[H^{(k)}(\rvz)\right]
    &\leq \frac{L_H^{(k)} + \beta}{2} \E\left[\|\rvx_{1}^{(k+1)} - \rvz\|^2\right] + \frac{1}{2n^2 \beta}(C_n^{(k)})^2
\end{align*}
and hence,
\begin{align}
\nonumber
    &\E\left[G(\rvx_1^{(k+1)}) - G(\rvz)\right]
    \leq \frac{1}{2n\eta}\Big(\E[\|\rvz - \rvx_1^{(k)}\|^2 ] - \E[\| \rvz - \rvx_1^{(k+1)}\|^2 ] \Big)
    + ( \frac{L_H^{(k)} + \beta}{2} - \frac{\mu_{\psi}}{2} ) \E [\| \rvz - \rvx_1^{(k+1)}\|^2 ] \\
\nonumber
    &\quad + (\widehat{L}^{(k)} - \frac{1}{2n \eta} ) \E[\| \rvx_1^{(k+1)} - \rvx_1^{(k)}\|^2]
    + \frac{1}{2n^2 \beta}(C_n^{(k)})^2
    + 5\eta^2 \frac{1}{n}\sum_{i=n_d^{(k)}+1}^{n-1} \widehat{L}^{(k)*} (C_i^{(k)})^2
    + \frac{5}{2}\eta^2 (C_n^{(k)})^2\\
    \nonumber
    &\quad + 10 \eta^2 n^2 L (\widehat{L}^{(k)} +1) \E\left[B_F(\rvz, \rvx^*)\right]
    + 5\eta^2 n^2 (\widehat{L}^{(k)} +1) \sigma_{any}^2 + 6 \eta^2 nd (\sigma^{(k)})^2 (\widehat{L}^{(k)*}+1)
    + \frac{1}{2}d (\sigma^{(k)})^2
\end{align}

Since $\eta \leq \frac{1}{n\sqrt{10 (\widehat{L}^{(k)} + 1)\widehat{L}^{(k)*}}}$, it follows that
$\widehat{L}^{(k)} \leq \sqrt{(\widehat{L}^{(k)} + 1) \widehat{L}^{(k)*}} \leq \frac{1}{n \eta \sqrt{10}} \leq \frac{1}{2n\eta}$.
Therefore, the term
$(\widehat{L}^{(k)} - \frac{1}{2n \eta}) \E[\|\rvx_1^{(k+1)} - \rvx_1^{(k)} \|^2] \leq 0$, which then yields the inequality in the lemma statement.
\end{proof}

The rest of the proof for the convergence across $K$ epochs directly follows the argument in section~\ref{subsec:appendix_k_epoch_convergence}. We provide the final convergence bound when $\psi(\rvx) = \gI\{\rvx \in \gB\}$ in Theorem~\ref{thm:convergence_proj} as follows.

\begin{theorem}[Convergence under projection]
\label{thm:convergence_proj}
    Under Assumptions~\ref{ass:convexity},~\ref{ass:dissim_partial_lipschitzness},~\ref{ass:appendix_refined_smoothness} and~\ref{ass:H_smoothness},
    suppose $\psi(\rvx) = \gI\{\rvx \in \gB\}$ for a convex set $\gB$.
    For $\beta > 0$, 
    if $\mu_{\psi} \geq L_H^{(k)} + \beta$, $\forall k\in [K]$, and 
     $\eta \leq \frac{1}{2n \sqrt{10 \bar{L}^* \max_{k\in [K]}\widehat{L}^{(k)*} (1+\log K)}}$, 
    where $\bar{L}^* = \max\{L, \max_{k\in[K]} \widehat{L}^{(k)}\}+1$,
    Algorithm~\ref{alg:generalized_shuffled_gradient_fm} guarantees

    \begin{align*}
        &\E\left[ G(\rvx_1^{(K+1)}) \right] - G(\rvx^*)
        \leq \underbrace{ \frac{\|\rvx_1^{(1)} - \rvx^{*}\|^2 }{\eta n K}
        }_{\text{Initialization}}
        + \underbrace{ 10 \eta^2 n^2 \sigma_{any}^2 (1+\log K)\max_{k\in[K]} \widehat{L}^{(k)} 
        }_{\text{Optimization Uncertainty}} + 2 M
    \end{align*}
    where
    \begin{align*}
        M &= \max_{s\in [K]} \Big(
        \underbrace{ \frac{1}{2n^2 \beta}\sum_{k=1}^{s} \frac{ (C_n^{(k)})^2 }{s+1-k}
        }_{\text{Non-vanishing Dissimilarity}}
        + \underbrace{5\eta^2 \sum_{k=1}^{s} \frac{ \widehat{L}^{(k)*} \frac{1}{n}\sum_{i=n_d^{(k)}+1}^{n-1}(C_i^{(k)})^2 + (C_n^{(k)})^2 } {s+1-k} }_{\text{Vanishing Dissimilarity}}\\
        \nonumber
        &\quad + \underbrace{ 6\eta^2 nd \sum_{k=1}^{s} \frac{ (\sigma^{(k)})^2 \widehat{L}^{(k)*} }{s+1-k} }_{\text{Injected Noise}} 
        + \underbrace{ \frac{1}{2}\sum_{k=1}^{s}\frac{d (\sigma^{(k)})^2}{s+1-k}
        }_{\text{Add. Error}}\Big)
    \end{align*}
    and the expectation is taken w.r.t. the injected noise $\{\rho_i^{(k)}\}$ and the order of samples $\pi^{(k)}$, $\forall i\in [n], k\in [K]$.
\end{theorem}

The convergence of $\DPSG$ follows directly from the above analysis as follows:
\begin{corollary}[Convergence of $\DPSG$ under projection]
\label{corollary:convergence_dpsg_proj}
    Under the conditions in Theorem~\ref{thm:convergence_proj}, with $\psi(\rvx) = \gI\{\rvx\in \gB\}$ for a convex set $\gB$, $n_d^{(k)}=n$, $\gP^{(k)} = \emptyset$, and $\sigma^{(k)} = \sigma$, for all $k\in [K]$,  Algorithm~\ref{alg:generalized_shuffled_gradient_fm} ($\DPSG$) guarantees
    \begin{align*}
        & \E [G(\rvx_1^{(K+1)})] - G(\rvx^*)\\
        &\lesssim \eta^2 n^2 \sigma_{any}^2 (1+\log K) L^*  + \frac{\|\rvx_1^{(1)} - \rvx^{*}\|^2}{\eta n K} + \eta^2 n d \sigma^2 L^{*} (1 + \log K)
        + d \sigma^2 (1 + \log K)
    \end{align*}
    and the expectation is taken w.r.t. the injected noise $\{\rho_i^{(k)}\}$ and the order of samples $\pi^{(k)}$, $\forall i\in [n], k\in [K]$.
\end{corollary}

It is unclear whether the additional error term $d\sigma^2(1+\log K)$ in the above convergence bound can be reduced when $\psi$ is the projection operator.
We leave this as an open question.

\section{Experiments}
\label{sec:appendix_experiments}

\textbf{Hardware.} All experiments were run on a 2021 MacBook Pro laptop, with an Apple M1 Pro chip.


\subsection{Tasks}
\label{subsec:appendix_tasks}

For every task, we describe the component function $f(\rvx; \rvq)$ on a given sample $\rvq \in \gD \cup \gP$, and the regularizer $\psi(\rvx)$. The true and the surrogate objective functions are constructed based on $f$ and $\psi$ accordingly. 
\begin{enumerate}[itemsep=0mm, leftmargin=4mm, topsep=0mm]
    \item \textbf{Mean Estimation}: $f(\rvx; \rvq) = \frac{1}{2}\| \rvx - \rvq\|^2$. $\psi(\rvx) = \gI\{\rvx \in \gB_{C}\}$, where $\gB_{C}$ is a ball of radius $C$ at the origin.
    \item \textbf{Ridge Regression}: 
    Let $\rvq = (\rva, y)$, where $\rva$ and $y$ represent the feature vector and the response, respectively.
    $f(\rvx; \rvq) = (\langle \rvx, \rva\rangle - y)^2$, $\psi(\rvx) = \frac{\lambda_{r}}{2}\|\rvx\|^2$ for $\lambda_{r} > 0$.
    \item \textbf{Lasso Logistic Regression}: 
    Let $\rvq = (\rva, y)$, for $y \in \{\pm 1\}$, represent the feature vector and label, respectively.
    $f(\rvx; \rvq) = -y \log (h(\rvx; \rva)) - (1-y) \log (h(\rvx; \rva))$, where $h(\rvx; \rva) = \frac{1}{1+\exp(-\langle \rvx, \rva \rangle)}$. $\psi(\rvx) = \lambda_{l} \|\rvx\|_1$ for $\lambda_{l} > 0$.
\end{enumerate}

\subsection{More about Datasets}
\label{subsec:appendix_datasets}

\begin{table}[h]
    \centering
\begin{adjustbox}{width=0.5\textwidth}
    \begin{tabular}{|c|c|c|c|}
    \hline
        Task & Dataset & $n$ & $d$ \\
    \hline
        Mean Estimation & \texttt{MNIST-69} & 1000 & 784\\
    \hline
        \multirow{2}{*}{Ridge Regression} & \texttt{CIFAR-10} & 1000 & 3072\\
        & \texttt{Crime} & 159 & 124\\
    \hline
        \multirow{2}{*}{\shortstack{Lasso \\Logistic Regression}} & \texttt{COMPAS} & 2013 & 11 \\
         & \texttt{CreditCard} & 200 & 21 \\
    \hline
    \end{tabular}
\end{adjustbox}
    \caption{A summary of datasets. }
    \label{tab:dataset_summary}
\end{table}

We construct the private ($\gD$) and public ($\gP$) sets of samples from each dataset for each task as follows:
\begin{enumerate}[itemsep=0mm, leftmargin=20pt]
        \item \textbf{Mean Estimation.}
        \begin{itemize}
            \item \texttt{MNIST-69}. 
            $n = 1000, d = 784$. We want to estimate the average pixel intensity of a given digit.
            $\gD$ consists of the first 1000 training samples of digit 6. $\gP$ consists of the first 1000 training samples of digit 9, with each sample rotated $180^{\circ}$ to mimic digit 6.
        \end{itemize}

    \item \textbf{Ridge Regression}:
    \begin{itemize}[itemsep=0mm]
        \item \texttt{CIFAR-10}. $n=1000$, $d = 3072$. The task is to predict the class of a given image. 
        $\gD$ contains 200 samples per class across 10 classes.
        $\gP$ simulates a real-world scenario where collecting data from certain classes is difficult, containing samples from only the first 4 classes (250 samples per class).
        \item \texttt{Crime}\footnote{\href{https://archive.ics.uci.edu/dataset/183/communities+and+crime}{Communities and Crime}}.
        $n = 159$, $d = 124$. The task is to predict per capita violent crimes in a region. Data with missing entries is removed and split into two halves. 
        $\gD$ consists of one half, while $\gP$ simulates corrupted data with a small random rotation: $\gP = \mX_0 \mR$, where $\mR = \sI_d + \gN(0, \sI_d)$ and $\mX_{0}$ represents the other half of the original dataset.
    \end{itemize}
    \item \textbf{Lasso Logistic Regression}:
    \begin{itemize}
        \item \texttt{COMPAS}
        \footnote{\href{https://raw.githubusercontent.com/propublica/compas-analysis/master/compas-scores-two-years.csv}{ProPublica Recidivism Dataset}}.
        $n= 2103, d = 11$. 
        The task is to predict whether a criminal defendant will reoffend within two years. The dataset, known for biases in predictions across ethnic groups, is split into African-American ($\gP$) and Caucasian ($\gD$) groups. This split reflects real-world disparities in data distributions.
         \item \texttt{CreditCard}
        \footnote{\href{https://archive.ics.uci.edu/dataset/350/default+of+credit+card+clients}{Default of Credit Card Clients}}.
        $n = 200, d = 21$. 
         The task is to predict whether a client defaults on their credit card payment. The dataset is split by education level: university-level ($\gP$) and below high school ($\gD$). 
         The private dataset ($\gD$) has a higher default rate, creating a balanced class distribution, while the public dataset ($\gP$) exhibits an extremely low default rate.
    \end{itemize}
\end{enumerate}

\subsection{Additional Results}
\label{subsec:appendix_more_results}

\subsubsection{Variants of $\DPSG$}
\label{subsec:appendix_var_dpsg}

In the main paper, we present results using Random Reshuffling (RR). Here, we show more results using the other two variants of shuffled gradient methods, Incremental Gradient (IG) and Shuffle Once (SO), on datasets \texttt{CreditCard} and \texttt{MNIST-69}.

Again, we replace ``ShuffleG'' in each algortihm's name with ``IG'' or ``SO''. This results in the following algorithms for comparison:
\begin{enumerate}
    \item IG-based: \textit{Interleaved-IG}, \textit{Priv-Pub-IG}, \textit{Pub-Priv-IG} and \textit{DP-IG}
    \item SO-based: \textit{Interleaved-SO}, \textit{Priv-Pub-SO}, \textit{Pub-Priv-SO} and \textit{DP-SO}
\end{enumerate}
We also include the baseline \textit{Public Only} which uses public samples ($\gP$) only. 

Here, we fix $p=0.5$ and the privacy parameters are $\eps\in \{5, 10\}$ and $\delta = 10^{-6}$.

\begin{figure}[H]
    \centering
    \includegraphics[width=0.24\linewidth]{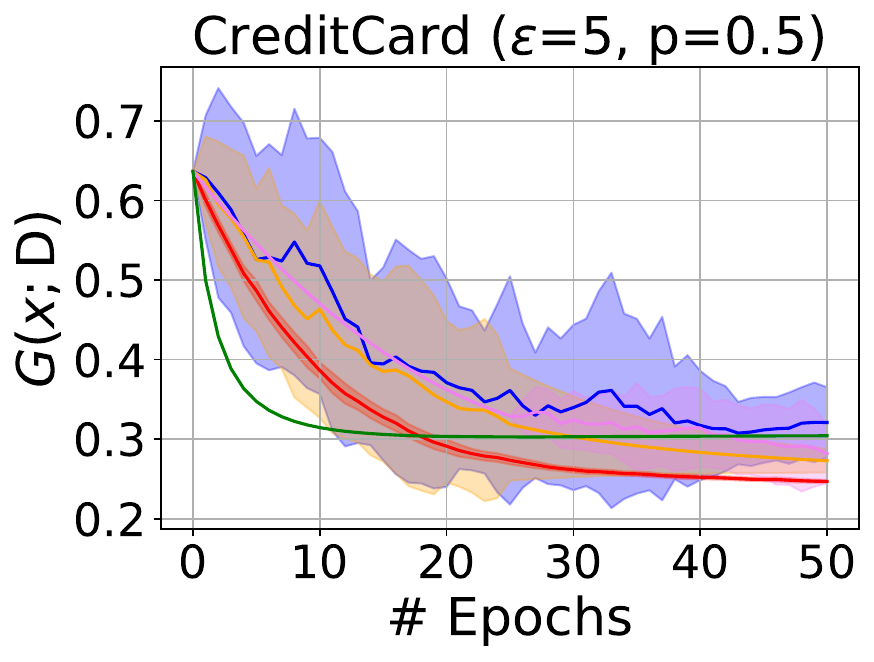}
    \includegraphics[width=0.24\linewidth]{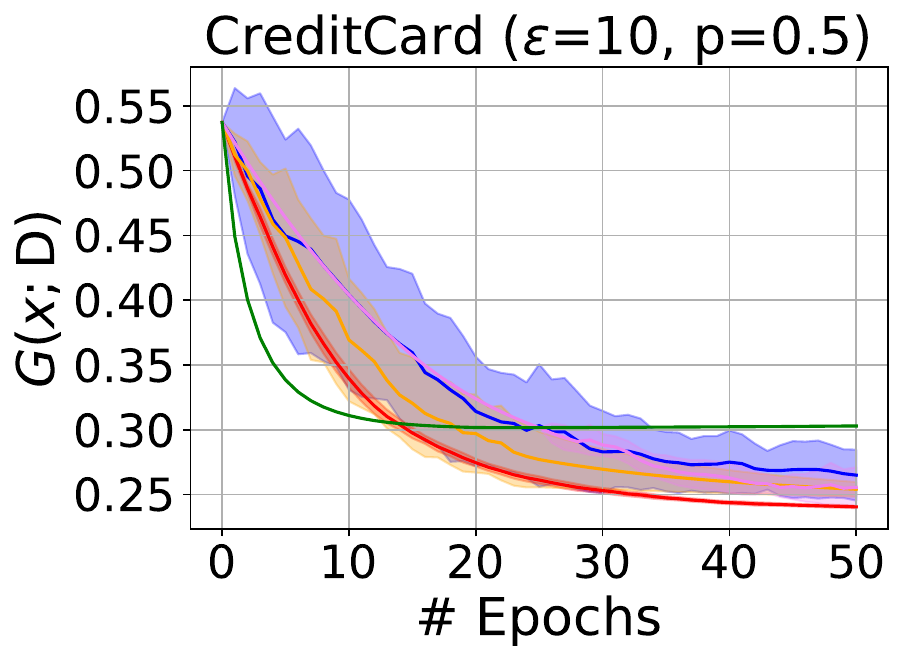}
    \includegraphics[width=0.24\linewidth]{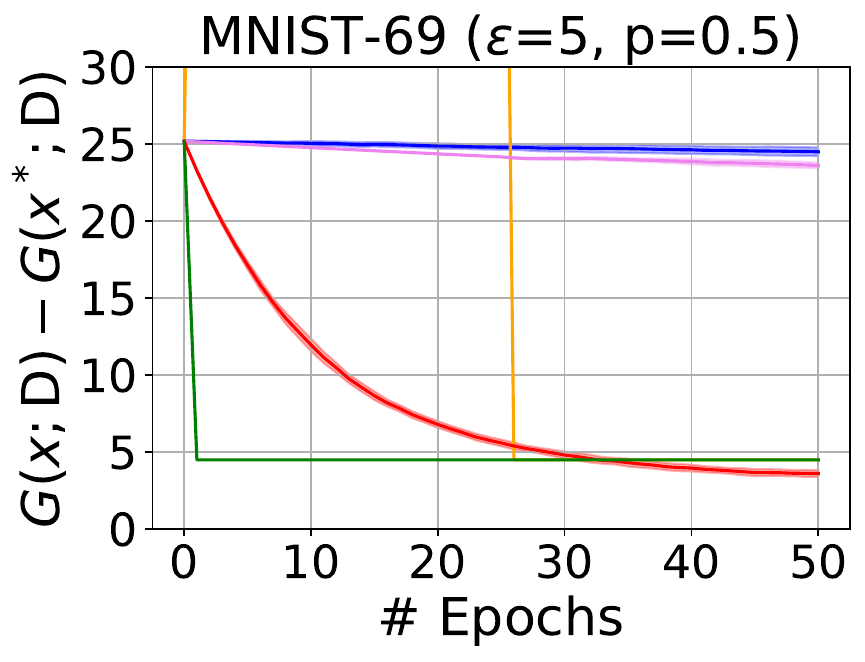}
    \includegraphics[width=0.24\linewidth]{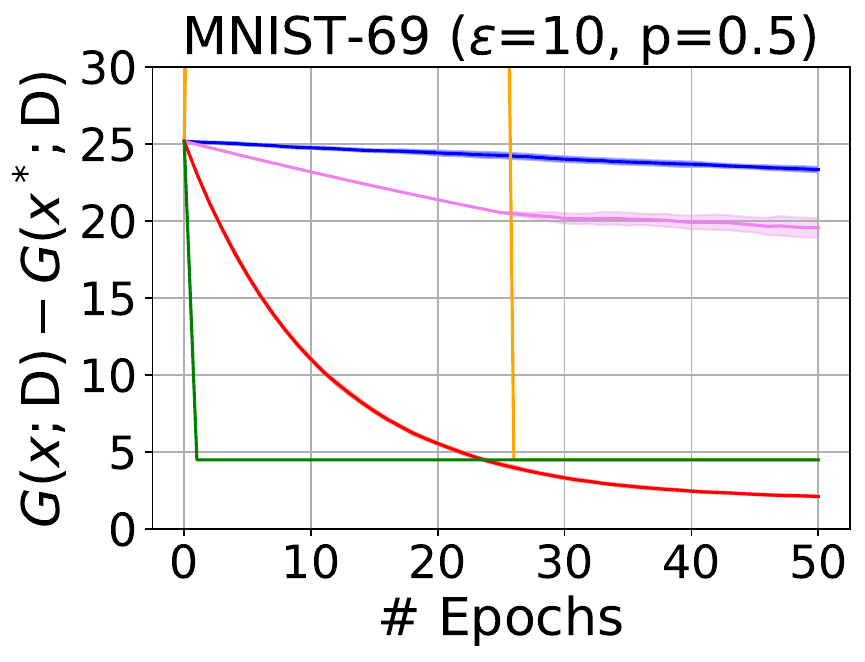}\\
    \includegraphics[width=0.48\linewidth]{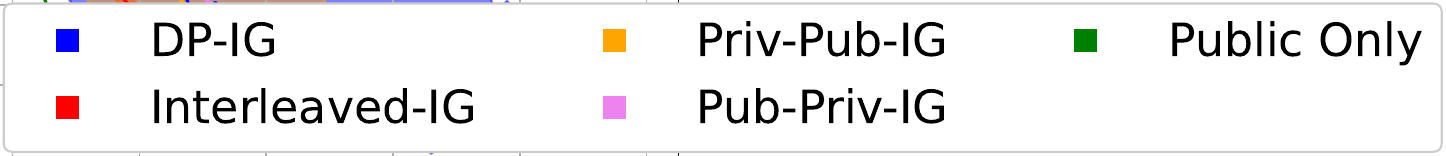}
    \caption{Results of comparing IG-based algorithms on two datasets.}
    \label{fig:appendix_IG}
\end{figure}

\begin{figure}[H]
    \centering
    \includegraphics[width=0.24\linewidth]{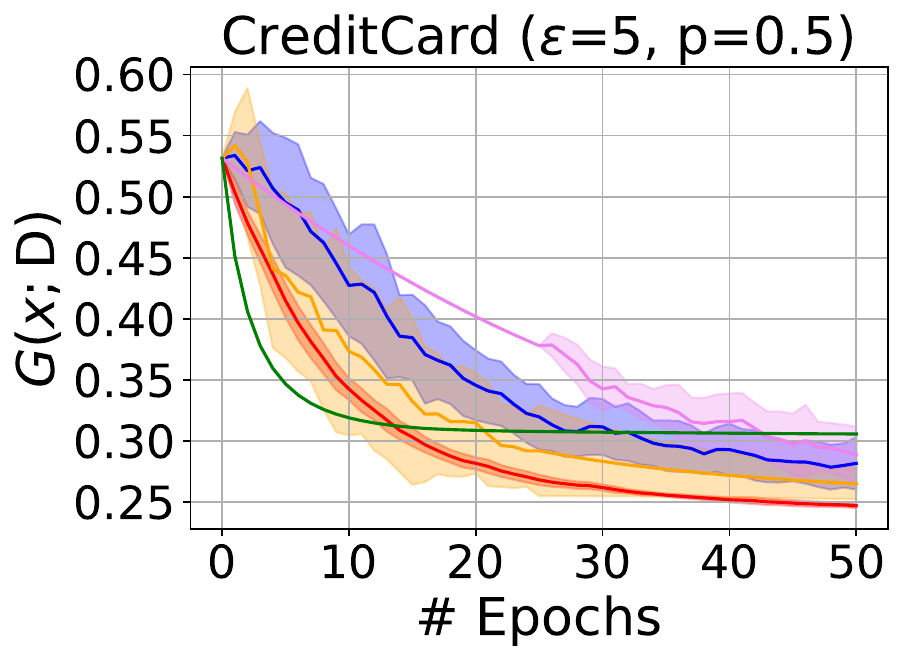}
    \includegraphics[width=0.24\linewidth]{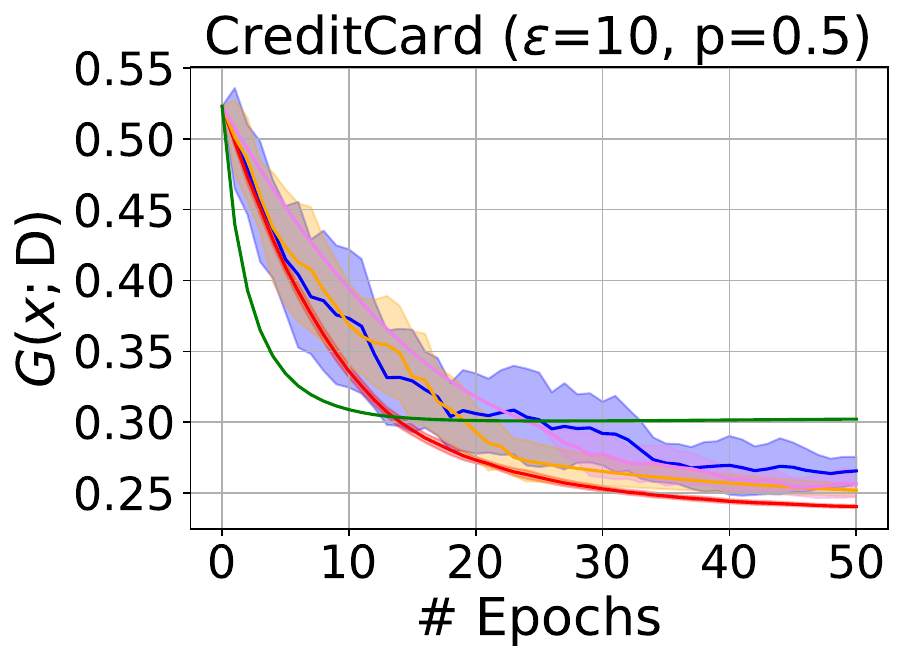}
    \includegraphics[width=0.24\linewidth]{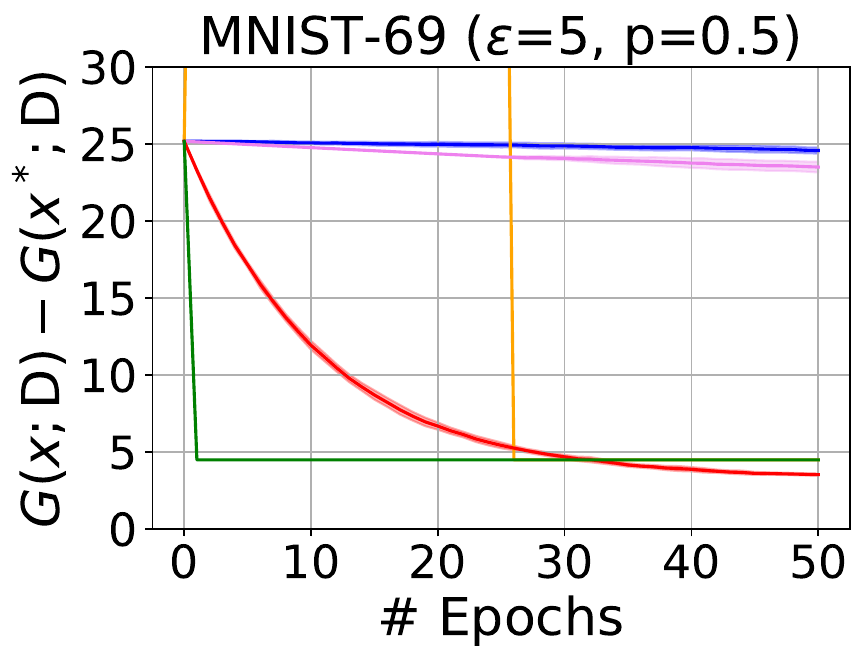}
    \includegraphics[width=0.24\linewidth]{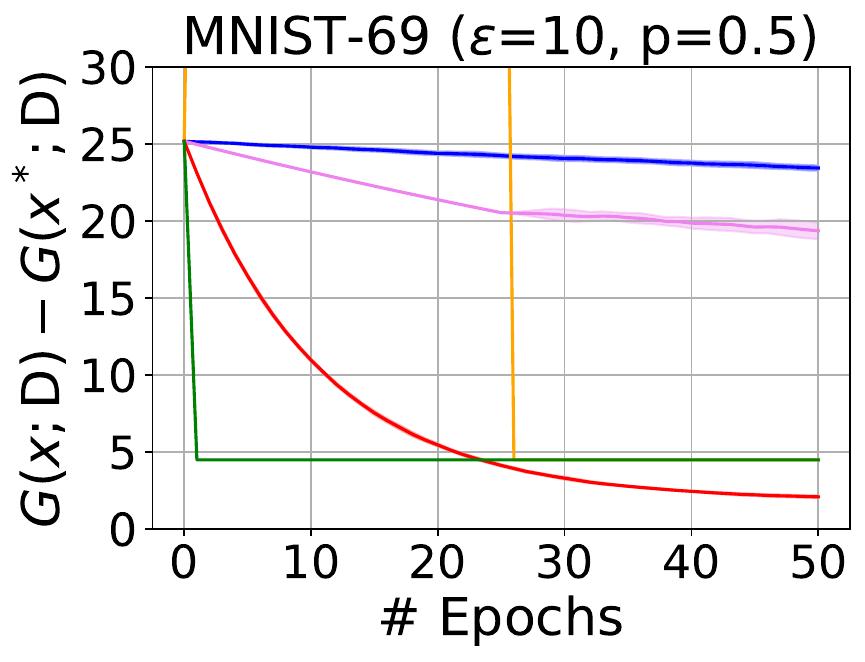}\\
    \includegraphics[width=0.48\linewidth]{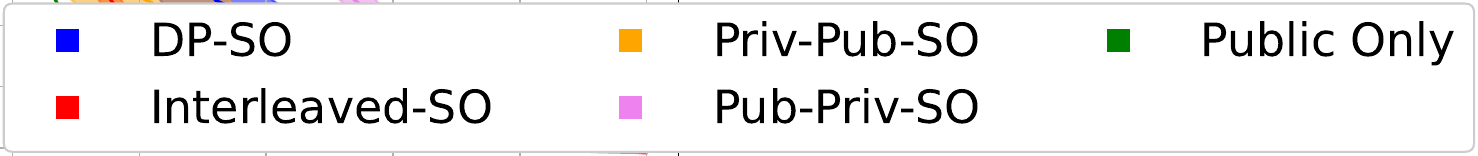}
    \caption{Results of comparing SO-based algorithms on two datasets.}
    \label{fig:appendix_SO}
\end{figure}

\vspace{-10pt}
\subsubsection{Varying $p$}

In this setting, we vary the fraction of private samples $p$ used in algorithms that leverage public data. Here, present results with $p\in \{0.25, 0.75\}$ on datasets \texttt{CreditCard} and \texttt{MNIST-69}.

We use RR in each algorithm. The privacy parameters are $\eps\in \{5, 10\}$ and $\delta=10^{-6}$.

\begin{figure}[H]
    \centering
    \includegraphics[width=0.24\linewidth]{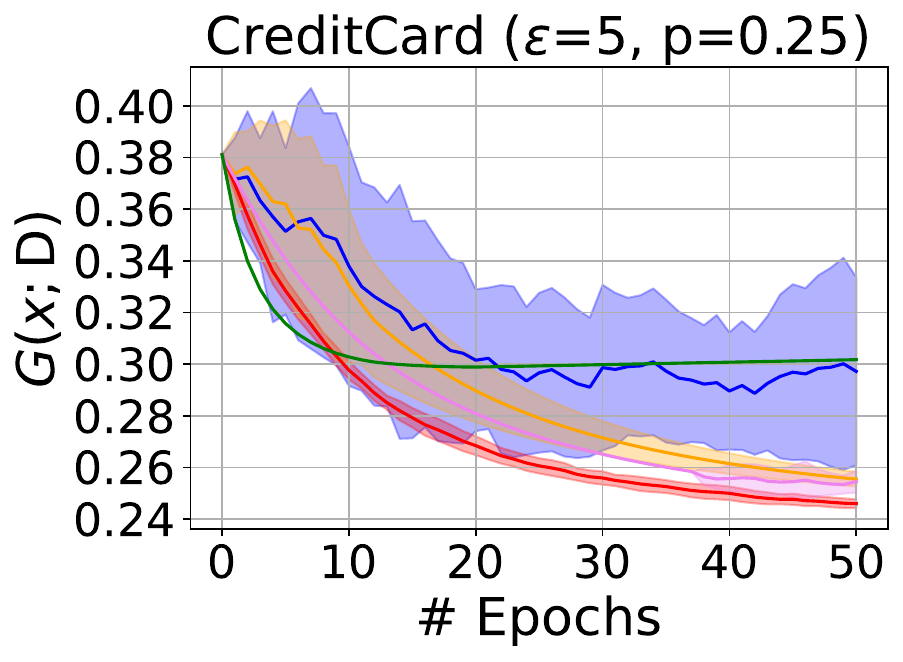}
    \includegraphics[width=0.24\linewidth]{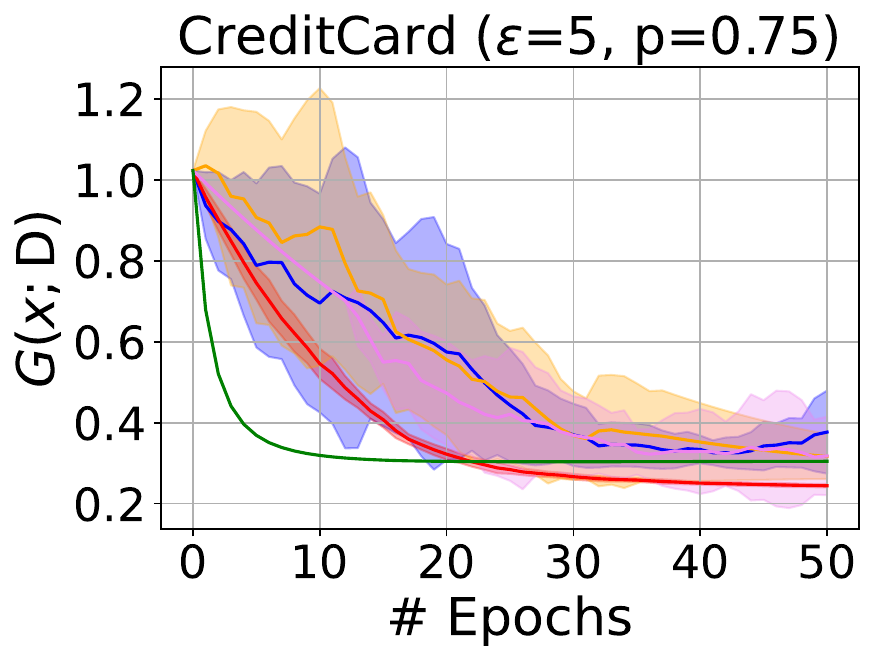}
    \includegraphics[width=0.24\linewidth]{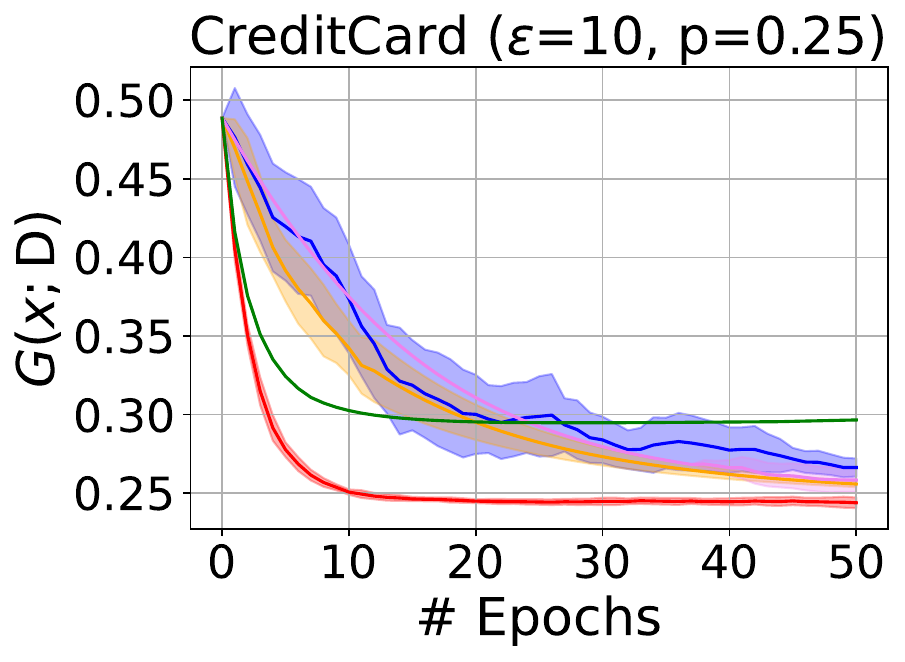}
    \includegraphics[width=0.24\linewidth]{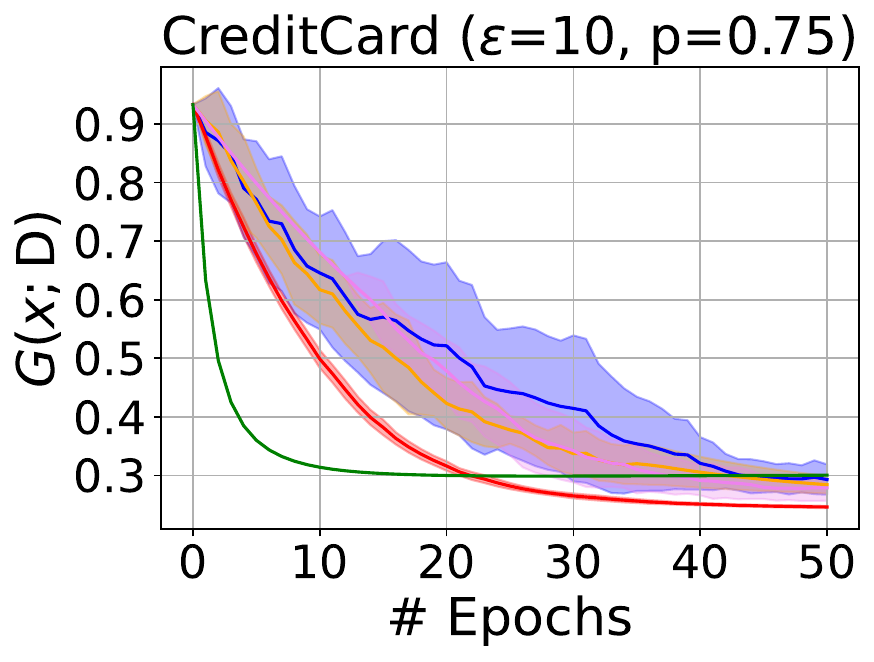}\\
    \includegraphics[width=0.48\linewidth]{res_plots/RR_legend.pdf}
    \caption{Results of using different fractions of private samples for $p\in \{0.25, 0.75\}$ on dataset \texttt{CreditCard}.}
    \label{fig:appendix_varying_p_credit_card}
\end{figure}

\begin{figure}[H]
    \centering
    \centering
    \includegraphics[width=0.24\linewidth]{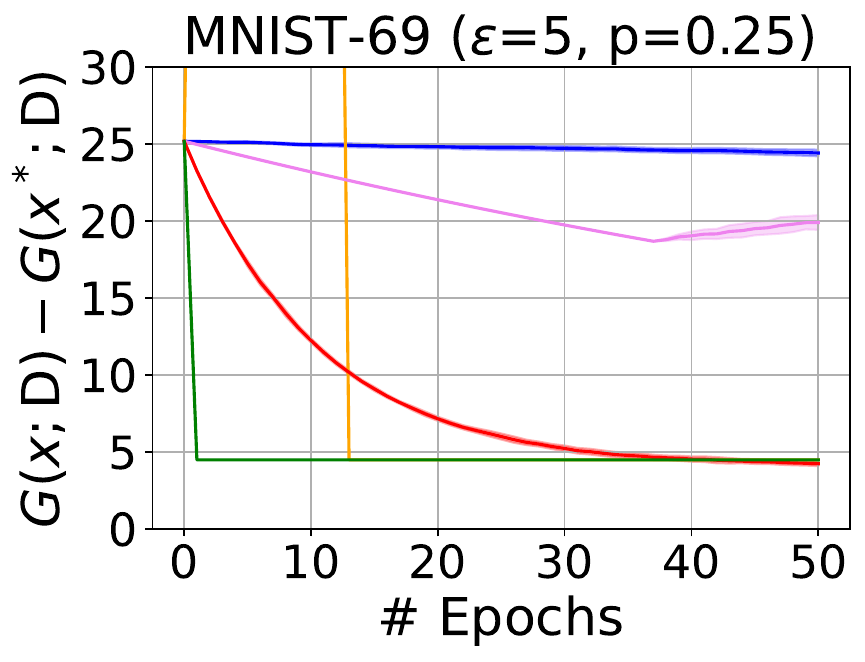}
    \includegraphics[width=0.24\linewidth]{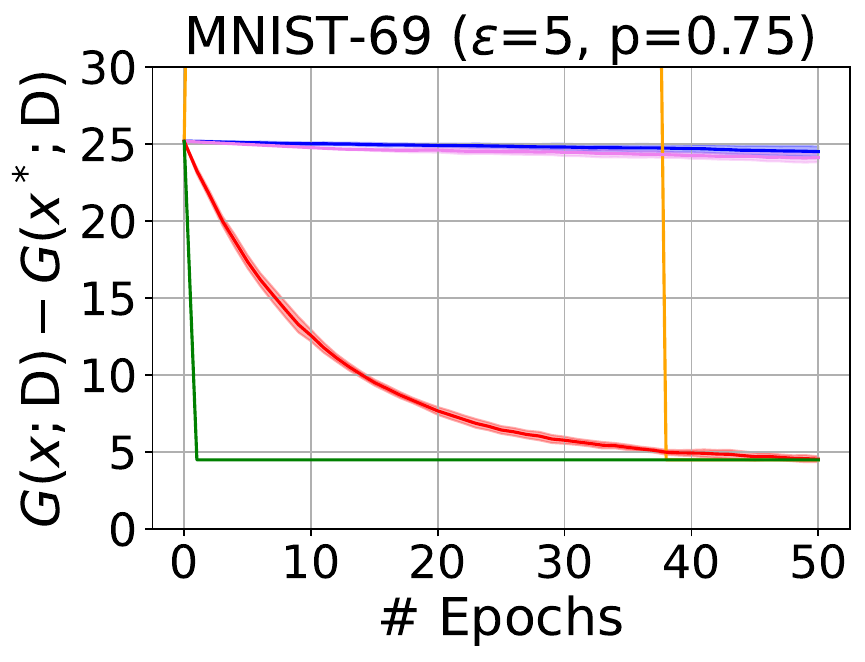}
    \includegraphics[width=0.24\linewidth]{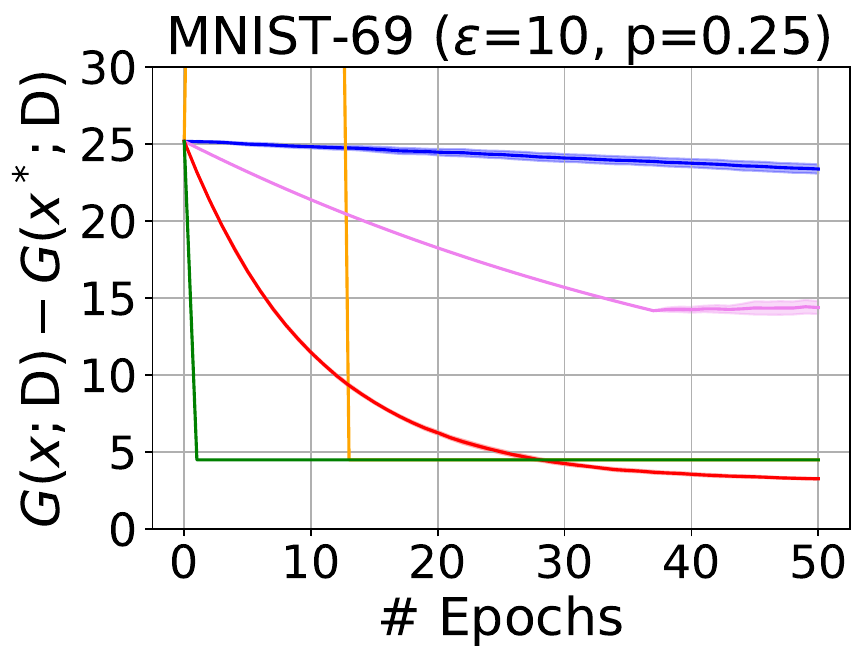}
    \includegraphics[width=0.24\linewidth]{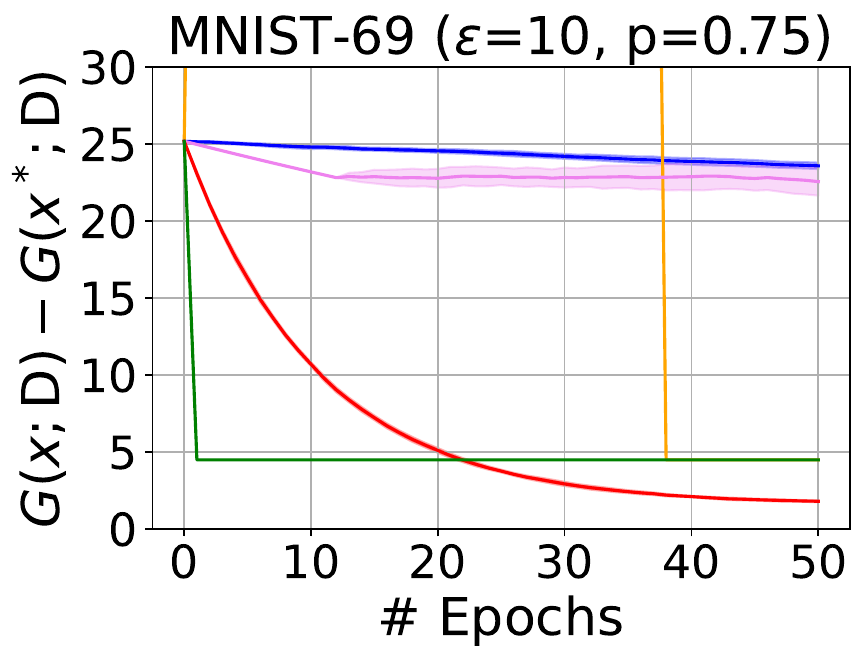}\\
    \includegraphics[width=0.48\linewidth]{res_plots/RR_legend.pdf}
    \caption{Results of using different fractions of private samples for $p\in \{0.25, 0.75\}$ on dataset \texttt{MNIST-69}.}
    \label{fig:appendix_varying_p_mnist69}
\end{figure}

\end{document}